%% file: main.tex
\documentclass[12pt, a4 paper]{article}
\usepackage[paper=A4,pagesize]{typearea}
\usepackage{graphicx}
\usepackage{afterpage}
\usepackage[utf8]{inputenc}
\usepackage{graphicx}
\usepackage{amssymb}
\usepackage{amsmath}
\usepackage{bm}
\usepackage{hyperref}
\usepackage{floatrow}
\usepackage{float, multirow}
\usepackage{xcolor}
\usepackage{setspace}
\usepackage{mathptmx} 
\usepackage{newtxtext,newtxmath}
\usepackage{lscape}
\usepackage{soul} 
\usepackage[title]{appendix}
\usepackage{setspace}
\usepackage{hanging}
\usepackage{tikz}
\usetikzlibrary{positioning}
\usetikzlibrary{shapes.geometric, arrows}
\usepackage{calc}
\usetikzlibrary{arrows.meta}
\usepackage{varwidth}
\usepackage[left=2.54cm, right=2.54cm, top=2.54cm, bottom=2.54cm]{geometry}
\usepackage[numbers, sort&compress, square]{natbib}
\usepackage{hyperref}
\hypersetup{colorlinks,linkcolor={blue},citecolor={blue},urlcolor={blue}}
\usepackage{tabulary}
\newcolumntype{L}[1]{>{\raggedright\let\newline\\\arraybackslash\hspace{0pt}}m{#1}}
\newcolumntype{C}[1]{>{\centering\let\newline\\\arraybackslash\hspace{0pt}}m{#1}}
\newcolumntype{R}[1]{>{\raggedleft\let\newline\\\arraybackslash\hspace{0pt}}m{#1}}
\usepackage{subcaption}
\makeatletter
\renewcommand{\fnum@figure}{Fig. \thefigure}									 
\makeatother
\usepackage{caption}
\usepackage{enumitem}
\newcommand\blfootnote[1]{%
	\begingroup
	\renewcommand\thefootnote{}\footnote{#1}%
	\addtocounter{footnote}{-1}%
	\endgroup
}

\usepackage{fancyhdr}
\usepackage{lineno}

\begin{document}
	\title{Physics-informed neural networks for predicting gas flow dynamics and unknown parameters in diesel engines}
	\author{Kamaljyoti Nath\textsuperscript{a, 1}, \and Xuhui Meng\textsuperscript{b, 1}, \and Daniel J Smith\textsuperscript{c}, \and George Em Karniadakis \textsuperscript{a,d,}\footnote{Corresponding author: G. E. Karniadakis (george\_karniadakis@brown.edu)}}
	\date{}
	
	\newcommand{\Addresses}{{%
			\textsuperscript{a} Division of Applied Mathematics, Brown University, United States of America \\
			\textsuperscript{b} Institute of Interdisciplinary Research for Mathematics and Applied Science, School of Mathematics and Statistics, Huazhong University of Science and Technology, Wuhan, China \\
			\textsuperscript{c} Cummins Inc., United States of America \\
			\textsuperscript{d} School of Engineering, Brown University, United States of America
	}} 
	
	\maketitle 
	\vspace*{-0.5cm}
	\begin{center}
		\Addresses
	\end{center}
	\blfootnote{\textsuperscript{1}These authors contributed equally (Kamaljyoti Nath, Xuhui Meng),\\
 \textit{E-mail addresses:} kamaljyoti\_nath@brown.edu (Kamaljyoti Nath), xuhui\_meng@hust.edu.cn (Xuhui Meng), daniel.j.smith@cummins.com (Daniel J Smith), george\_karniadakis@brown.edu (George Em Karniadakis) }
	\thispagestyle{fancy}
	\cfoot{}
	\rhead{}
	\renewcommand{\headrulewidth}{0pt}
	\rfoot{\today}
	\blfootnote{}
	\begin{abstract}
This paper presents a physics-informed neural network (PINN) approach for monitoring the health of diesel engines. The aim is to evaluate the engine dynamics, identify unknown parameters in a "mean value" model, and anticipate maintenance requirements. The PINN model is applied to diesel engines with a variable-geometry turbocharger and exhaust gas recirculation, using measurement data of selected state variables. The results demonstrate the ability of the PINN model to predict simultaneously both unknown parameters and dynamics accurately with both clean and noisy data, and the importance of the self-adaptive weight in the loss function for faster convergence. The input data for these simulations are derived from actual engine running conditions, while the outputs are simulated data, making this a practical case study of PINN's ability to predict real-world dynamical systems. The mean value model of the diesel engine incorporates empirical formulae to represent certain states, but these formulae may not be generalizable to other engines. To address this, the study considers the use of deep neural networks (DNNs) in addition to the PINN model. The DNNs are trained using laboratory test data and are used to model the engine-specific empirical formulae in the mean value model, allowing for a more flexible and adaptive representation of the engine's states. In other words, the mean value model uses both the PINN model and the DNNs to represent the engine's states, with the PINN providing a physics-based understanding of the engine's overall dynamics and the DNNs offering a more engine-specific and adaptive representation of the empirical formulae. By combining these two approaches, the study aims to offer a comprehensive and versatile approach to monitoring the health and performance of diesel engines.
\end{abstract}
\section{Introduction}
\label{Section:Introduction}
Powertrains of the future must meet increasingly stringent requirements for emissions, performance, reliability, onboard monitoring, and serviceability. Capable system models for estimating states and adapting to an individual system's behaviour are critical elements to meet control and health monitoring needs. Leveraging purely data-driven models to meet these requirements provides simplicity in modelling and captures dynamics difficult to formulate analytically. However, large data needs, poor physical interpretability, challenges with systems with long memory effects and sparse sensing, as well as inability to extrapolate beyond the training datasets present onerous burdens to practical implementation. Relying on purely theory-based models allows for directly interpretable results with higher confidence and fewer data for calibration but often causes a tradeoff of modelling relevant dynamics versus model complexity, challenges in systems with high uncertainties, poor modelling where dynamics are not well understood, and slow solution of higher-order models. Modelling solutions that leverage the strengths of theory-guided as well as data-driven models have the potential to reduce data needs, increase robustness, and effectively use theoretical and practical knowledge of the system.
\par To investigate model architectures, balancing the strengths of both theory-based models and data-driven models, this work explores the application of physics-informed neural networks (PINNs) to a diesel internal combustion engine model for the purposes of simultaneous parameter and state estimation. The physical portion is based on the mean value model of a diesel engine with a variable geometry turbocharger (VGT), and exhaust gas recirculation (EGR) proposed by \citeauthor{Wahlstrom} \cite{Wahlstrom}.
\par Physics-informed Neural Networks (PINNs) \citep{Raissi_2019} is a new method of training neural networks, which takes into account the physics of a problem while evaluating the parameters of the neural network. The method is suitable for both evaluation of the solution of PDF (forward problem) and the data-driven identification of parameters of PDF (inverse problem). It takes advantage of automatic differentiation \citep{Baydin_2018} in formulating a physical loss in the loss function along with data loss. Jagtap et al. \cite{Jagtap_2020_CPINN} proposed conservative PINNs (cPINNs) for conservation laws, which employs domain decomposition with a PINN formulation in each domain. Further, Jagtap and Karniadakis \cite{Jagtap_2020_XPINN} introduced domain decomposition for  general PDEs using the so-called extended PINN (XPINN). hp-VPINN is a variational formulation of PINN with domain decomposition proposed by Kharazmi et al. \cite{Kharazmi2021}. Meng et al. \cite{Meng_2020} proposed the Parareal PINN (PPINN) approach for long-time integration of time-dependent partial differential equations. The authors of \cite{Cho_2022_separable} proposed ``separable" PINN, which can reduce the computational time and increase accuracy for high dimensional PDEs. In PINN, there are multiple loss functions, and the total loss function is given by the weighted sum of individual losses. McClenny and Braga-Neto \cite{McClenny_2020} proposed a self-adaptive weight technique, which is capable of tuning the weights automatically. PINN and its variants were also  considered in various inverse problems like supersonic flows \cite{Jagtap_20221_supersonic_flows}, nano-optics and metamaterials \cite{Chen_2020_neno_optics}, unsaturated groundwater flow \cite{Depina_2022_unsaturated}. Detailed reviews of PINN can be found in \cite{Cai_2022, Karami_2022, Cola_2022_Scientific}.
\par Modelling of diesel engines using neural networks has been considered in the past. Biao et al. \cite{Biao_2009} considered Nonlinear Auto-Regressive Moving Average with eXogenous inputs (NARMAX) method for system identification of locomotive diesel engines. The model has three inputs to the network, i.e. the fuel injected, the load of the main generator, and the feedback rotation speed (from the output); the outputs are rotation speed and diesel power. The authors considered Levenberg-Marquardt (LM) algorithm to train the network. Finesso and Spessa \cite{Finesso_2014} developed a three-zone thermodynamic model to predict nitrogen oxide and in-cylinder temperature heat release rate for direct injection diesel engines under steady state and transient conditions. The model is zero-dimensional, and the equations can be solved analytically. Thus, it required a very short computational time. Tosun et al. \cite{Tosun_2016} predicted torque, carbon monoxide, and oxides of nitrogen using neural networks (3 independent networks) for diesel engines fueled with biodiesel-alcohol mixtures. The authors considered three fuel properties (density, cetane number, lower heating value) and engine speed as input parameters and the networks are optimized using the Levenberg-Marquardt method. The authors observed that neural network results are better than the least square method. Gonz\'{a}lez et al. \cite{Jorge_2022} integrated a data-driven model with a physics-based (equation-based) model for the gas exchange process of a diesel engine. The authors modelled the steady-state turbocharger using a neural network. Further, the authors integrated the data-driven model with an equation-based model. Recently, Kumar et al. \cite{Kumar_2023} considered DeepONet \cite{Lu_2021_deepOnet}  to predict the state variable of the same mean value engine model \cite{Wahlstrom} we considered in this study. The authors consider dynamic data to train the model. However, the model can predict only the state variable at the particular (trained) ambient temperature and pressure, as variations of ambient temperature and pressure are not considered in the training of DeepONet. The model also does not predict the parameters of the engine model. While the model was trained using dynamic data, the physics of the problem was not considered while training the network. The model (DeepONet) is capable of predicting dynamic responses.
\par In the present study, we formulate a PINN model for the data-driven identification of parameters and prediction of dynamics of system variables of a diesel engine. In PINN, the physics of the system is directly included in the form of physics loss along with data loss. While data-driven models require large amount over the entire operational range in training, PINN can be trained with a smaller amount of data as it is trained online. The dynamics characteristic of the state variables is automatically incorporated. PINN may be used for the solution of differential equations or for the identification of parameters and prediction of state variables. In the present study, we are specifically interested in estimating unknown parameters and states when we know a few state variables from field data. The dynamics of the state variables of the mean value engine \citep{Wahlstrom} are described by first-order differential equations. We will utilize these equations in the formulation of the physics-informed loss function. The unknown parameters are considered trainable and updated in the training process along with the neural network parameters.
\par The engine model also considers a few empirical formulae in its formulation. These equations are engine-specific, and the coefficients of these equations need to be evaluated from experimental data. These equations are static in nature, and thus may be trained with smaller data compared to dynamic equations. We know that deep neural networks (DNNs) are universal approximators of any continuous function, thus, DNNs may be considered more general approximators of these empirical formulae. One of the advantages of considering DNNs over empirical formulae is that we do not need to assume the type of non-linearity between the input out variables. The neural network learns the non-linearity if trained with sufficient data.
We approximate the empirical formulae using DNNs and train them using laboratory test data. Once these networks are trained using laboratory test data, these are considered in the PINNs model in places of the empirical formulae. During the training of the PINNs model, the parameters of these networks are remain constant.
\par The training data for the inverse problem and laboratory data are generated using the Simulink file \citep{Simulink_file} accompanied in \cite{Wahlstrom}. The input to Simulink is taken from actual field data. By doing this, we are trying to generate data as realistic as field data. Furthermore, we also consider noise to the field data generated. We observed that the proposed PINNs model can predict the dynamics of the states and unknown parameters. We summarize below a few of the salient features of the present study:
\begin{enumerate}
    \item We formulated PINNs-based parameter identification for real-world dynamical systems, in the present case, a diesel engine. This is significant as it started a new paradigm for future research for onboard systems for the health monitoring of engines.
    \item We showed how PINNs could be implemented in predicting important unknown parameters of diesel engines from field data. From these predicted parameters, one can infer the health and serviceability requirements of the engine.
    \item We showed the importance of self-adaptive weights (given the fast transient dynamics) in the accuracy and faster convergence of results for PINNs for the present study.
    \item The engine model generally considers empirical formulae to evaluate a few of its quantities. These empirical formulae are engine-specific and require lab test data for the evaluation of the coefficients. We have shown how neural networks can be considered to model the empirical formulae. We have shown how we can train these networks from lab-test data. This is important as it may provide a better relationship for the empirical formulae.
    \item The field data for the inverse problem are generated considering input recorded from actual engine running conditions. Further, we consider appropriate noise in the simulated data, mimicking near real-world field data.
\end{enumerate}
\par We organize the rest of the article as follows: in section \ref{Section:Problem setup}, we discuss the detailed problem statement and different cases considered for simulation studies. In section \ref{Section:Methodology}, first, we discuss PINNs for the inverse problems for the diesel engine and the surrogates for the empirical formula. We discuss a detailed flow chart for the inverse problem for the PINN engine model in section \ref{Subsection:Flowchart for PINN model for engine}. In section \ref{Section:Data acquisition}, we discuss the laboratory data required and their generation for the training of surrogates for the empirical formulae. We also discuss the field data generation for the inverse problem. We present the results and discussion in section \ref{Section:Results and discussions}. The conclusions of the present study are discussed in the section \ref{Section:Conclusions}.

\subsection{Problem setup}
\label{Section:Problem setup}
In this section, we first introduce the mean value model for the gas flow dynamics \cite{Wahlstrom} in the diesel engine, and then we will formulate the inverse problems that we are interested in.
\par
\begin{figure}[H]
    \centering
    \includegraphics[scale=1.0]{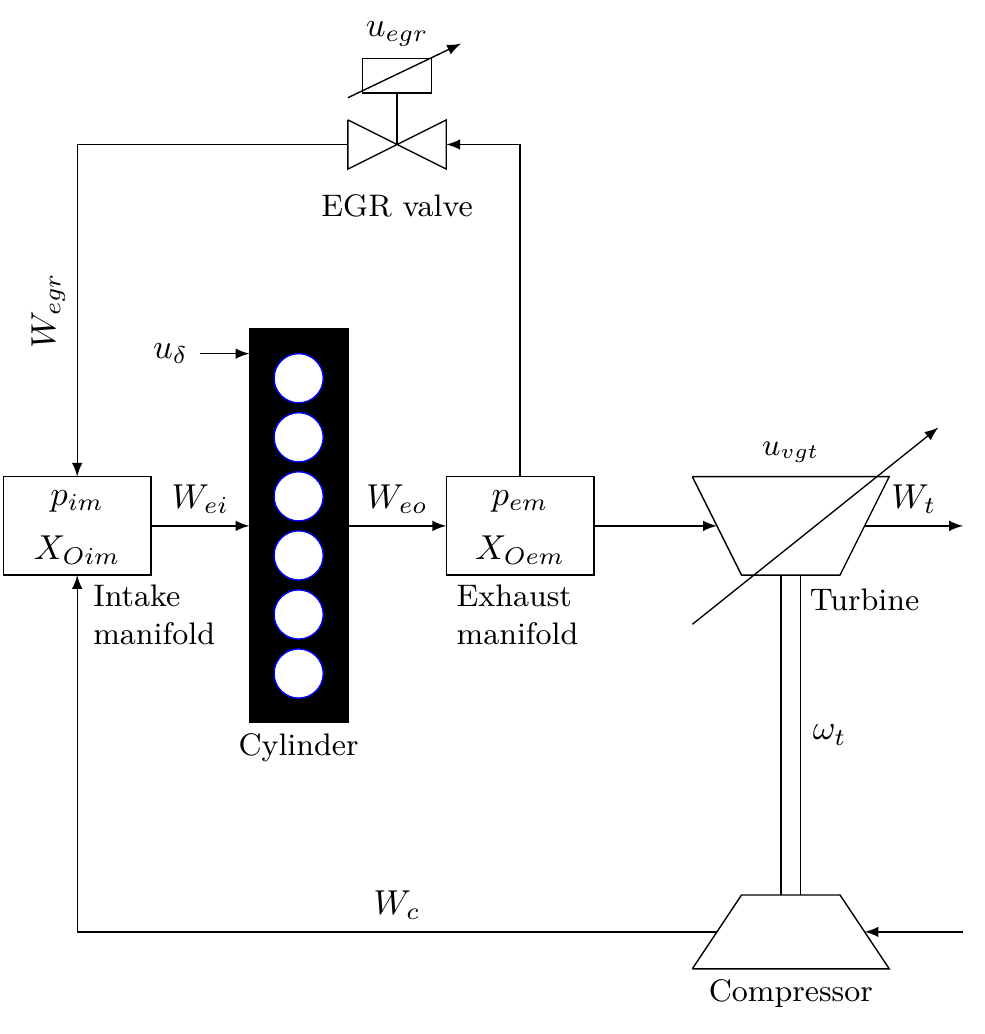}
    \caption{\textbf{Schematic diagram of the diesel engine:} A schematic diagram of the mean value diesel engine with a variable-geometry turbocharger (VGT) and exhaust gas recirculation (EGR) \citep{Wahlstrom}. The main components of the engine are the intake manifold, the exhaust manifold, the cylinder, the EGR valve system, the compressor, and the turbine. The control input vector is $\bm{u} = \{u_\delta, u_{egr}, u_{vgt}\}$, and engine speed is $n_e$. (Source: Figure is adopted from \cite{Wahlstrom})}
    \label{Fig:Schematic diagram of Engine}
\end{figure}

 As shown in Fig. \ref{Fig:Schematic diagram of Engine}, the engine model considered in the present study mainly comprises six parts: the intake and exhaust manifold, the cylinder, the exhaust gas recirculation (EGR) valve system, the compressor and the turbine. More details on each engine part can be seen in Appendix \ref{Appendix:Engine model}. We note that the engine considered here is the same as in \cite{Wahlstrom}.
\par To describe the gas flow dynamics in the engine illustrated in Fig. \ref{Fig:Schematic diagram of Engine}, e.g., the dynamics in the manifold pressures, turbocharger, EGR and VGT actuators, a mean value model of the diesel engine with variable geometric turbocharger and exhaust gas recirculation was proposed in  \cite{Wahlstrom}. We will also utilize the same model as the governing equations to describe the gas flow dynamics considered in the current study. Specifically, the model proposed in \cite{Wahlstrom} has eight states expressed as follows:
\begin{equation}
    \bm{x} = \{p_{im},\;\; p_{em},\;\; X_{Oim},\;\; X_{Oem},\;\; \omega_t,\;\; \tilde{u}_{egr1},\;\; \tilde{u}_{egr2},\;\; \tilde{u}_{vgt}\},
    \label{Eq:States}
\end{equation}
where $p_{im}$ and $p_{em}$ are the intake and exhaust manifold pressure, respectively, $X_{Oim}$ and $X_{Oem}$ are the oxygen mass fractions in the intake and exhaust manifold, respectively, $\omega_t$ is the turbo speed; $\tilde{u}_{vgt}$ represents the VGT actuator dynamics. A second-order system with an overshoot and a time delay is used to represent the dynamics of the EGR-valve actuator. The model is represented by subtraction of two first-order models, $\tilde{u}_{egr1}$ and  $\tilde{u}_{egr2}$, with
different gains and time constants. Further, the control inputs for the engine are $\bm{u} = \{u_\delta,\;\; u_{egr}, \;\; u_{vgt}\}$ and the engine speed is $n_e$, in which $u_\delta$ is the mass of injected fuel, $u_{egr}$ and $u_{vgt}$ are the EGR and VGT valve positions, respectively. Furthermore, the position of the valves, i.e., $u_{egr}$ and $u_{vgt}$, may vary from 0\% to 100\%, which indicates the complete close and opening of the valves, respectively. The mean value engine model is then expressed as
\begin{equation}
    \dot{\bm{x}} = f(\bm{x}, \bm{u}, n_e).
\end{equation}
\par In addition, the states describing the oxygen mass fraction of the intake and exhaust manifold, i.e., $X_{Oim}$ and $X_{Oem}$, are not considered in the present study as the rest of the states do not depend on these two states. Also, the parameters of the oxygen mass fractions are assumed to be constant and known. The governing equations for the remaining six states are as follows:
\begin{equation}
    \dfrac{d}{dt} p_{im} = \dfrac{R_a T_{im}}{V_{im}}(W_c+W_{egr} - W_{ei}),
    \label{Eq:p_im}
\end{equation}
\begin{equation}
    \dfrac{d}{dt} p_{em} = \dfrac{R_e T_{em}}{V_{em}}(W_{eo} - W_t - W_{egr}),
    \label{Eq:p_em}
\end{equation}
\begin{equation}
	\dfrac{d}{d t}\omega_t = \dfrac{P_t\eta_m - P_c}{J_t\omega_t},
	\label{Eq:omega_t}
\end{equation}
\begin{equation}
    \dfrac{d\Tilde{u}_{egr1}}{dt} = \dfrac{1}{\tau_{egr1}}\left[u_{egr}(t-\tau_{degr}) - \Tilde{u}_{egr1}\right],
    \label{Eq:u_egr_1}
\end{equation}
\begin{equation}
    \dfrac{d\Tilde{u}_{egr2}}{dt} = \dfrac{1}{\tau_{egr2}}\left[u_{egr}(t-\tau_{degr}) - \Tilde{u}_{egr2}\right],
    \label{Eq:u_egr_2}
\end{equation}
\begin{equation}
    \dfrac{d\Tilde{u}_{vgt}}{dt} = \dfrac{1}{\tau_{vgt}}\left[u_{vgt}(t-\tau_{dvgt}) - \Tilde{u}_{vgt}\right].
    \label{Eq:u_vgt}
\end{equation}
Two additional equations used for the computation of $T_{em}$ in Eq. \eqref{Eq:p_em} read as:
\begin{equation}
    T_1 = x_rT_e + (1-x_r)T_{im},
    \label{Eq:T_1}
\end{equation}
\begin{equation}
    x_r = \dfrac{\Pi_e^{1/\gamma_a}x_p^{-1/\gamma_a}}{r_c x_v},
    \label{Eq:x_r}
\end{equation}
where $T_1$ is the temperature when the inlet valve closes after the intake stroke and mixing, and $x_r$ is the residual gas fraction. A brief discussion on the governing equations of the engine model is presented in Appendix \ref{Appendix:Engine model}. Interested readers can also refer to \cite{Wahlstrom} for more details.
\par In the present study, we have field measurements on a certain number of variables, i.e., $p_{im}$, $p_{em}$, $\omega_t$, and $W_{egr}$ as well as the inputs, i.e., $\bm{u}$ and $n_e$, at discrete times. Further, some of the parameters in the system, e.g., $A_{egrmax}$, $\eta_{sc}$, $h_{tot}$ and $A_{vgtmax}$, which are difficult to measure directly, are unknown. $A_{egrmax}$ is the maximum effective area of the EGR valve, $\eta_{sc}$ is the compensation factor for non-ideal cycles, $h_{tot}$ is the total heat transfer coefficient of the exhaust pipes and $A_{vgtmax}$ is the maximum area in the turbine that the gas flows through. From the field prediction of these parameters, we can infer the health of the engine; a higher deviation from their design value may indicate a fault in the system. We are interested in (1) predicting the dynamics of all the variables in Eqs. \eqref{Eq:p_im}-\eqref{Eq:x_r}, and (2) identifying the unknown parameters in the system, given field measurements on $p_{im}$, $p_{em}$, $\omega_t$, and $W_{egr}$ as well as  Eqs. \eqref{Eq:p_im}-\eqref{Eq:x_r}. We refer to the above problem as the {\em inverse problem} in this study.  Specifically, the following cases are considered for a detailed study:
\begin{enumerate}[label=\bfseries Case \arabic*, leftmargin=*]
    \item Prediction of dynamics of the system and identification of 3 unknown parameters $A_{egrmax}$, $\eta_{sc}$ and $h_{tot}$ with clean data of $p_{im}$, $p_{em}$, $\omega_t$, and $W_{egr}$.
    \item Prediction of dynamics of the system and identification of 3 unknown parameters $A_{egrmax}$, $\eta_{sc}$ and $h_{tot}$ with noisy data of $p_{im}$, $p_{em}$, $\omega_t$, and $W_{egr}$.
    \item Prediction of dynamics of the system and identification of 4 unknown parameters $A_{egrmax}$, $\eta_{sc}$, $h_{tot}$ and $A_{vgtmax}$ with clean data of $p_{im}$, $p_{em}$, $\omega_t$, and $W_{egr}$.
    \item Prediction of dynamics of the system and identification of 4 unknown parameters $A_{egrmax}$, $\eta_{sc}, h_{tot}$ and $A_{vgtmax}$ with noisy data of $p_{im}$, $p_{em}$, $\omega_t$, and $W_{egr}$.
\end{enumerate}
    In the present study, we consider self-adaptive weights \cite{McClenny_2020} (discussed in section \ref{Section:Methodology} and Appendix \ref{Appendix:Neural network pinn}) in our loss function. We study the above four cases using self-adaptive weight. In order to understand the effect and importance of self-adaptive weights in the convergence and accuracy of results, we consider one more case without self-adaptive weight,
 \begin{enumerate}[label=\bfseries Case \arabic*, leftmargin=*, start=5]
     \item Prediction of dynamics of the system and identification of 4 unknown parameters $A_{egrmax}$, $\eta_{sc}$, $h_{tot}$ and $A_{vgtmax}$ with clean data of $p_{im}$, $p_{em}$, $\omega_t$, and $W_{egr}$ without self-adaptive weights.
 \end{enumerate}
\par The results of \textbf{Case 1} and \textbf{Case 2} are presented in Appendix \ref{Appendix:Additiona Figure}. First, we study the results of \textbf{Case 3} and \textbf{Case 5} to understand the accuracy and convergence of PINN method and the importance of self-adaptive weights. Then, we study the results of \textbf{Case 4}. The results for \textbf{Case 3}, \textbf{Case 4} and \textbf{Case 5} are discussed in section \ref{Section:Results and discussions}.

\section{Methodology}
\label{Section:Methodology}
We consider to employ the deep learning algorithm, particularly, the physics-informed neural networks (PINNs), to solve the inverse problem discussed in section \ref{Section:Problem setup}. To begin with, we first briefly review the basic principle of PINNs, and then we discuss how to employ PINNs for the present inverse problem.
\subsection{PINNs for inverse problems in the diesel engine}
\label{Subsecction:PINN general}
We first briefly review the PINNs \citep{Raissi_2019, Meng_2020, Jagtap_2020_CPINN, Jagtap_2020_XPINN, Kharazmi2021} for solving inverse problems, and then we introduce how to employ the PINNs to solve the specific problem that we are of interest for the diesel engine.
\begin{figure}[H]
    \centering
    \includegraphics[width=0.8\textwidth]{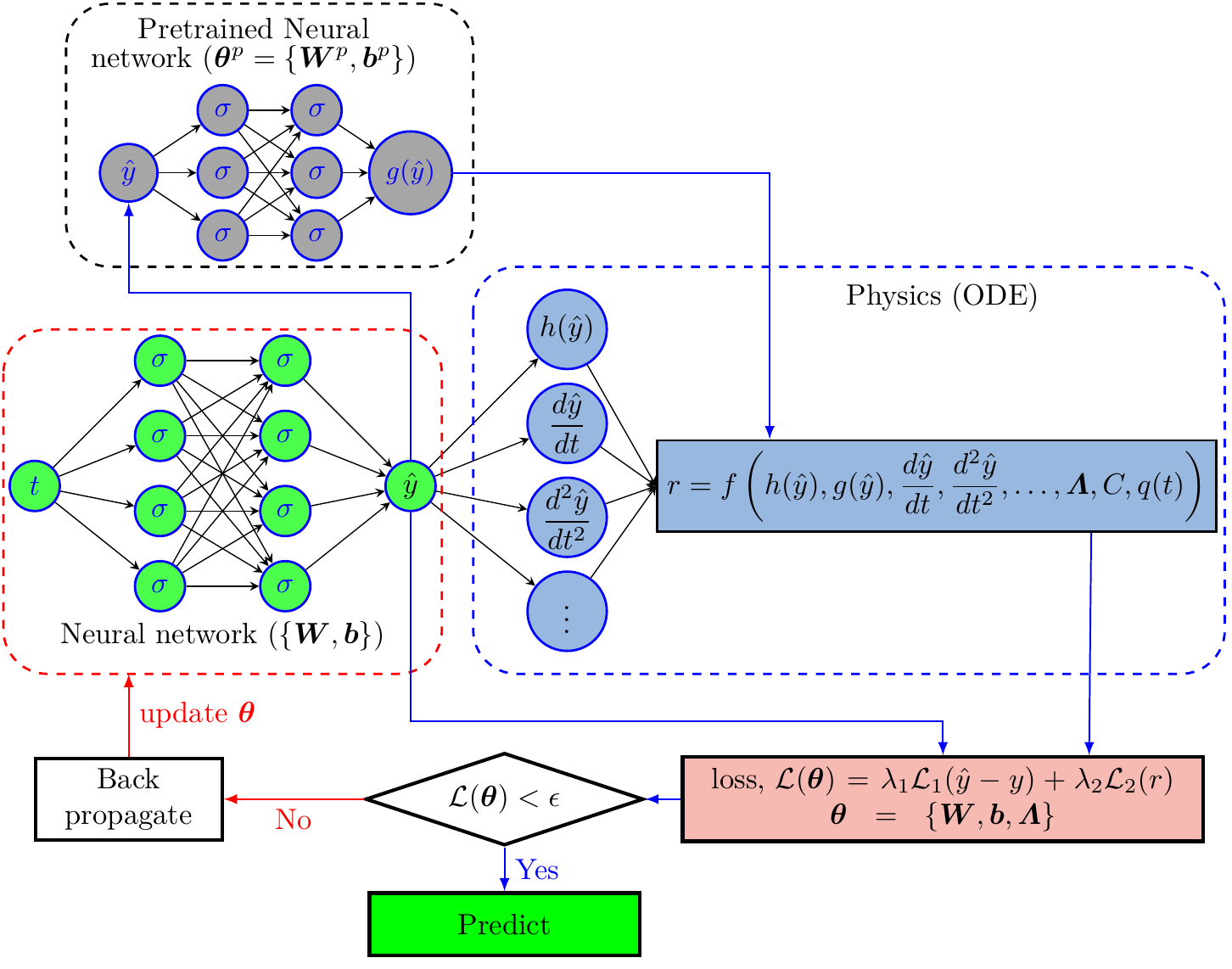}
    \caption{\textbf{Schematic of physics-informed neural networks (PINNs) for inverse problems:} The left part of the figure, enclosed in the red dashed line, shows a DNN whose input is time. The DNN is to approximate the solution ($y$) to a differential equation. The top left part of the figure enclosed in the black dashed line shows an DNN whose input is the output $y$ (maybe with other input, e.g. ambient condition). The output of this network is a function $g(y)$. This network is pre-trained with laboratory data of $y$ and $g(y)$. The right part of the figure, enclosed in the blue dashed line, denotes the physics loss/residue. The DNN (enclosed in a red dashed line) approximates the solution to any differential equation, and the equation is encoded using automatic differentiation. The total loss $\mathcal{L}(\bm{\theta})$ includes the loss of equation as well as the data. The $\lambda_1$ and $\lambda_2$ are two weights to the data loss and physics loss, which may be fixed or adaptive depending upon the problem and solution method. $\bm{\theta} = \{\bm{W}, \bm{b}, \bm{\varLambda}\}$ represents the parameters in DNN, $\bm{W}$ and $\bm{b}$ are the weights and biases of DNN, respectively and $\bm{\varLambda}$ are the unknown parameters of the ODE; $\sigma$ is the activation function, $q(t)$ is the right-hand side (RHS) of the differential equation (source term), $h$ is the function of the predicted variable, and $r$ is the residual for the equation. $\bm{\theta}^P = \{\bm{W}^P, \bm{b}^P\}$ represents the parameters in pre-trained neural network, $\bm{W}^P$ and $\bm{b}^P$ are the weights and biases of the pre-trained neural network.}
    \label{Fig:PINNs general}
\end{figure}
\par As illustrated in Fig. \ref{Fig:PINNs general}, the PINN is composed of two parts, i.e., a fully-connected neural network which is to approximate the solution to a particular differential equation and the physics-informed part in which the automatic differentiation \citep{Baydin_2018} is employed to encode the corresponding differential equation. Further, $\bm{\mathcal{\varLambda}}$ represents the unknowns in the equation, which can be either a constant or a field. In particular, $\bm{\mathcal{\varLambda}}$ are trainable variables as the unknowns are constant, but they could also be approximated by a DNN if the unknown is a field. The loss function for solving the inverse problems consists of two parts, i.e., the data loss and the equation loss, which reads as:
\begin{align}\label{eq:loss_inverse}
    \mathcal{L}(\bm{\theta}; \bm{\mathcal{\varLambda}}) = \underbrace{\frac{1}{M}\sum^M_{i=1} |\hat{y}(t_i; \bm{\theta}) - y(t_i)|^2}_{\mbox{data ~loss}} + \underbrace{\frac{1}{N}\sum^N_{i=1} |r(t_i; \bm{\theta}; \bm{\mathcal{\varLambda}})|^2}_{\mbox{equation loss}}
\end{align}
where $\bm{\theta}$ denotes the parameters in the DNN; $M$ and $N$ represent the number of measurements and the residual points, respectively; $\hat{y}(t_i; \bm{\theta})$ denotes the prediction of DNN at the time $t_i$; $y(t_i)$ is the measurement at $t_i$, and $r(t_i; \bm{\theta}; \bm{\mathcal{\varLambda}})$ represents the residual of the corresponding differential equation, which should be zero in the entire domain. By minimizing the loss in Eq. \eqref{eq:loss_inverse}, we can obtain the optimal parameters, i.e., $\bm{\theta}$, of the DNN as well as the unknowns, i.e., $\bm{\mathcal{\varLambda}}$, in the system. In the present study, we have a few empirical equations that we approximate using DNNs. These DNNs are trained first using data and considered in place of these empirical formulae. We fixed the parameters of these networks when we minimized the loss function for the PINN model. Furthermore, note that here we employ the system described by one equation as the example to demonstrate how to use PINNs for solving inverse problems. For the system with more than one equation, we can either utilize an DNN with multiple outputs or multiple DNNs as the surrogates for the solutions to differential equations. In addition, a similar idea can also be employed for systems with multiple unknown fields. The loss function can then be rewritten as
\begin{align}
    \mathcal{L}(\bm{\theta}; \bm{\mathcal{\varLambda}}) = \underbrace{ \sum^K_{k=1}\left[ \frac{1}{M_k}\sum^{M_k}_{i=1} |\hat{y}_k(t_i; \bm{\theta}) - y_k(t_i)|^2\right]}_{\mbox{data ~loss}} + \underbrace{\sum^L_{l=1} \left[\frac{1}{N_l}\sum^{N_l}_{i=1} |r_l(t_i; \bm{\theta}; \bm{\mathcal{\varLambda}})|^2\right]}_{\mbox{equation loss}}
    \label{eq:loss_inverse_multiple}
\end{align}
where $K$ and $L$ denote the number of variables that can be measured as well as the equations, respectively; $M_k$ and $N_l$ are the number of measurements for the $k^{\text{th}}$ variable and the number of residual points for the $l^{\text{th}}$ equation, respectively; and $\bm{\mathcal{\varLambda}}$ collects all the unknowns in the system. 
\begin{table}[H]
    \centering
    \caption{Neural network surrogates employed PINNs for solving the inverse problems. $\mathcal{N}_i(t; \bm{\theta}_i), i = 1, ..., 6$ denotes the surrogate for the $i^{th}$ DNN parameterized by $\bm{\theta}_i$ with the input $t$. In particular, $\mathcal{N}_1(t; \bm{\theta}_1)$ and $\mathcal{N}_4(t; \bm{\theta}_4)$ have two outputs, which are used to approximate $\{p_{im},p_{em}\}$ and $\{\tilde{u}_{egr1}, \tilde{u}_{egr2}\}$, respectively; the remaining DNNs have only one output.} 
    \label{Table:FNN for PINN} 
    \begin{tabular}{c|c|c|c|c|c|c}
    \hline
    Variables   & ${p}_{im}$, ${p}_{em}$ & ${x}_{r}$ & ${T}_{1}$ & ${\tilde{u}}_{egr1}$, ${\tilde{u}}_{egr2}$ & ${\omega}_{t}$ & ${\tilde{u}}_{vgt}$ \\ \hline
    Surrogate  & $\mathcal{N}_{1}(t; \bm{\theta}_1)$ &  $\mathcal{N}_{2}(t; \bm{\theta}_2)$ & $\mathcal{N}_{3}(t; \bm{\theta}_3)$ & $\mathcal{N}_{4}(t; \bm{\theta}_4)$ & $\mathcal{N}_{5}(t; \bm{\theta}_5)$ & $\mathcal{N}_{6}(t; \bm{\theta}_6)$\\ \hline
    Equation & \ref{Eq:p_im} and \ref{Eq:p_em} & \ref{Eq:x_r} & \ref{Eq:T_1} & \ref{Eq:u_egr_1} and \ref{Eq:u_egr_2} & \ref{Eq:omega_t} & \ref{Eq:u_vgt}\\
    \hline
\end{tabular}
\end{table}
\par For the inverse problem presented in section \ref{Section:Problem setup}, we are interested in (1) learning the dynamics of the six states (2) inferring the unknown parameters in the system, given measurements on $\{p_{im}, p_{em}, \omega_t, W_{egr}\}$ as well as Eqs. \eqref{Eq:p_im}-\eqref{Eq:x_r}, using PINNs. Specifically, we utilize six DNNs as the surrogates for the solutions to different equations, and the corresponding equations are encoded using the automatic differentiation, as illustrated in Table \ref{Table:FNN for PINN}. In addition, the loss for training the PINNs for Case 1 to Case 4 is expressed as follows:
\begin{equation}
    \begin{split}
    \mathcal{L}(\bm{\theta}, \bm{\mathcal{\varLambda}}, \bm{\lambda}_{p_{im}}, \bm{\lambda}_{p_{em}}, \bm{\lambda}_{\omega_t}, \bm{\lambda}_{W_{egr}}, \lambda_{T_1}) = & \mathcal{L}_{p_{im}} + \mathcal{L}_{p_{em}} + \mathcal{L}_{\omega_{t}} +  \mathcal{L}_{u_{egr1}} + \\ & \mathcal{L}_{u_{egr2}} + \mathcal{L}_{u_{vgt}} +  10\times \mathcal{L}_{x_{r}} + \lambda_{T_1}\times\mathcal{L}_{T_{1}} + \\
    & \mathcal{L}^{ini}_{p_{im}} + \mathcal{L}^{ini}_{p_{em}} + \mathcal{L}^{ini}_{\omega_{t}} +  \mathcal{L}^{ini}_{\tilde{u}_{egr1}} + \\ 
    & \mathcal{L}^{ini}_{\tilde{u}_{egr2}} + \mathcal{L}^{ini}_{\tilde{u}_{vgt}} + \mathcal{L}^{ini}_{x_{r}} + 100\times \mathcal{L}^{ini}_{T_{1}} + \\
    & \mathcal{L}^{data}_{p_{im}}(\bm{\lambda}_{p_{im}}) + \mathcal{L}^{data}_{p_{em}}(\bm{\lambda}_{p_{em}}) + \\ 
    & \mathcal{L}^{data}_{\omega_t}(\bm{\lambda}_{\omega_t}) + \mathcal{L}^{data}_{W_{egr}}(\bm{\lambda}_{W_{egr}}),
    \end{split}
\label{Eq:Loss total loss}
\end{equation}
where $\bm{\theta} = (\bm{\theta}_1, ..., \bm{\theta}_6)$ are the parameters of all NNs in PINNs, $\bm{\mathcal{\varLambda}}$ are the unknown parameters, which will be inferred from the given measurements, $\mathcal{L}_{\phi}, ~ \phi = (p_{im}, p_{em}, \omega_t, \tilde{u}_{egr1}, \tilde{u}_{egr2}, \tilde{u}_{vgt}, x_r, T_1)$ are the losses for the corresponding equations, and $\mathcal{L}^{data}_{\psi}, ~ \psi = (p_{im}, p_{em}, \omega_t, W_{egr})$ are the losses for the corresponding measurements, and $\mathcal{L}^{ini}_{\phi}, ~ \phi = (p_{im}, p_{em}, \omega_t, \tilde{u}_{egr1}, \tilde{u}_{egr2}, \tilde{u}_{vgt}, x_r, T_1)$ are the losses for the initial conditions, $\bm{\lambda}_{p_{im}}$, $\bm{\lambda}_{p_{em}}$, $\bm{\lambda}_{\omega_t}$, and $\bm{\lambda}_{W_{egr}}$ are the weights for different loss terms which are used to balance each term in the loss function. In particular, the self-adaptive weight technique proposed in \citep{McClenny_2020}, which is capable of tuning the weights automatically, is utilized here to obtain the optimal $\bm{\lambda}_{p_{im}}$, $\bm{\lambda}_{p_{em}}$, $\bm{\lambda}_{\omega_t}$, and $\bm{\lambda}_{W_{egr}}$. More details for self-adaptive weights in PINN can be found in Appendix \ref{Appendix:Neural network pinn}.
 \par In the Case 5 where we have not considered self-adaptive weights, so  the loss function is given as
 \begin{equation}
     \begin{split}
     \mathcal{L}(\bm{\theta}, \bm{\mathcal{\varLambda}}) = & \mathcal{L}_{p_{im}} + \mathcal{L}_{p_{em}} + \mathcal{L}_{\omega_{t}} +  \mathcal{L}_{u_{egr1}} + \\ & \mathcal{L}_{u_{egr2}} + \mathcal{L}_{u_{vgt}} +  10\times \mathcal{L}_{x_{r}} + 10^3\times\mathcal{L}_{T_{1}} + \\
     & \mathcal{L}^{ini}_{p_{im}} + \mathcal{L}^{ini}_{p_{em}} + \mathcal{L}^{ini}_{\omega_{t}} +  \mathcal{L}^{ini}_{\tilde{u}_{egr1}} + \\ 
     & \mathcal{L}^{ini}_{\tilde{u}_{egr2}} + \mathcal{L}^{ini}_{\tilde{u}_{vgt}} + \mathcal{L}^{ini}_{x_{r}} + 100\times \mathcal{L}^{ini}_{T_{1}} + \\
     & 10^3\times\mathcal{L}^{data}_{p_{im}} + 10^3\times\mathcal{L}^{data}_{p_{em}} + \\ 
     & 10^3\times\mathcal{L}^{data}_{\omega_t} + 10^3\times\mathcal{L}^{data}_{W_{egr}},
     \end{split}
 \label{Eq:Loss total loss Case 5}
 \end{equation}
\par As for training the PINNs in the present study, we first employ the first-order optimizer, i.e., Adam \citep{Kingma_2014adam}, to train the parameters in the NNs, unknowns in the systems as well as the self-adaptive weights for a certain number of steps. We then fix the self-adaptive weight and employed Adam to train the parameters in the NNs, and unknowns in the systems for another certain number of steps. We then switch to the second-order accuracy optimizer, i.e., LBFGS-B, to further optimize the parameters in the NNs and the unknowns in the systems. Note that the self-adaptive weights are optimized at the first training stage of Adam only, and they are fixed during the second training stage of Adam and LBFGS-B training with the values at the end of the first stage of Adam optimization.
\subsubsection{Neural network surrogates for empirical formulae}
\label{Subsection:Surrogate models for empirical formulae}
In the mean value engine model proposed in \cite{Wahlstrom}, empirical formulae, e.g., polynomial functions, are employed for the volumetric efficiency ($\eta_{vol}$), effective area ratio function for EGR valve ($f_{egr}$), turbine mechanical efficiency ($\eta_{tm}$), effective area ratio function for VGT ($f_{vgt}$), choking function (for VGT) ($f_{\Pi_t}$), compressor efficiency ($\eta_c$), and volumetric flow coefficient (for the compressor) ($\Phi_c$). Note that these empirical formulae are engine-specific and may not be appropriate for the diesel engines considered in the present study. Deep neural networks (DNNs),  which are known to be universal approximators of any continuous function, are thus utilized as more general surrogates for the empirical formulae here. Particularly, we employ six DNNs for the aforementioned variables, and the inputs for each DNN are presented in Table \ref{Table:Surrogate empirical formulae}. 
\begin{table}[H]
\centering
\caption{Neural network surrogates for empirical formulae
$\mathcal{N}_i^{(P)}(\bm{x}; \bm{\theta}_i^P)$, $i= 1,\dots, 6$ denotes the surrogate for the $i^{th}$ DNN parameterized by $\theta_i^P$ with the input $\bm{x}$. All the neural networks have one output each.}
\label{Table:Surrogate empirical formulae}
\footnotesize
\begin{tabular}{c|c|c|c|c|c|c} \hline
Variable & $\eta_{vol}$ & $f_{egr}$ & $F_{vgt, \Pi_t}$ & $\eta_{tm}$ & $\eta_{c}$ & $\Phi_c$  \\ \hline
Surrogate & $\mathcal{N}_1^{(P)}(\bm{x}; \bm{\theta}_1^P)$ & $\mathcal{N}_2^{(P)}(\bm{x}; \bm{\theta}_2^P)$ & $\mathcal{N}_3^{(P)}(\bm{x}; \bm{\theta}_3^P)$ & $\mathcal{N}_4^{(P)}(\bm{x}; \bm{\theta}_4^P)$ & $\mathcal{N}_5^{(P)}(\bm{x}; \bm{\theta}_5^P)$ & $\mathcal{N}_6^{(P)}(\bm{x}; \bm{\theta}_6^P)$ \\ \hline
Input ($\bm{x}$) & $\{p_{im}, n_e\}$ & $\tilde{u}_{egr}$ & $\{\Tilde{u}_{vgt}, \Pi_t\}$ & $\{ \omega_t, T_{em}, \Pi_t \}$ & $\{W_c, \Pi_c\}$ & $\{T_{amb}, \Pi_c, \omega_t\}$\\ \hline
Equation$^\dagger$ & \ref{Eq:Append:eta_vol} & \ref{Eq:Append:f_egr} & \ref{Eq:Appendix:F_VGT_PI_T_Cal} & \ref{Eq:Append:eta_tm} & \ref{Eq:Append:eta c} & \ref{Eq:Append:3}\\ \hline
\multicolumn{7}{l}{$^\dagger$ The empirical equations are discussed in Appendix} \\ \hline
\end{tabular}
\end{table}
\par We now discuss the training of the DNNs illustrated in Table \ref{Table:Surrogate empirical formulae}. In laboratory experiments, measurements on all variables are available. We can then train the neural network surrogates in Table \ref{Table:Surrogate empirical formulae} using the data collected in the laboratory. The loss function considered for the training of these networks is
\begin{equation}
    \mathcal{L}_i(\bm{\theta}_i^P) = \dfrac{1}{n_i}\sum_{j=1}^{n_i} \left[y_i^{(j)} - \hat{y}_i^{(j)}\right]^2 =  \dfrac{1}{n_i}\sum_{j=1}^{n_i}\left[y_i^{(j)} - \mathcal{N}_i^{(P)}(\bm{x}_i;\bm{\theta}_i^P)^{(j)}\right]^2,\;\;\;\;\; i=1,2,\dots,6
\end{equation}
where $i=1,2,\dots,6$ are the different neural networks for the approximation of the empirical formulae, $\bm{x}_i$ are the input corresponds to the $i$\textsuperscript{th} network, $\hat{y}_i$ and $y_i$ are the output of the $i$\textsuperscript{th} network and the corresponding labelled values respectively, $n_i$ is the number of labelled dataset corresponds to the $i$\textsuperscript{th} neural network. The laboratory data required for calculating labelled data for training each of these networks are shown in Table \ref{Table:Data lab test} (in \S \ref{Section:Data acquisition}). We discuss the calculation of labelled data from the laboratory data in Appendix \ref{Appendix:Data for pre-trained network}. We train these networks using the Adam optimizer. Upon the training of these DNNs, we will plug them in the PINNs to replace the empirical models, which are represented by the pretrained neural network with the output $g(y)$ in Fig. \ref{Fig:PINNs general}.

\subsection{Flowchart for PINN model for diesel engine}
\label{Subsection:Flowchart for PINN model for engine}
\begin{landscape}
\begin{figure}[H]
    \centering
	\includegraphics[scale=0.61]{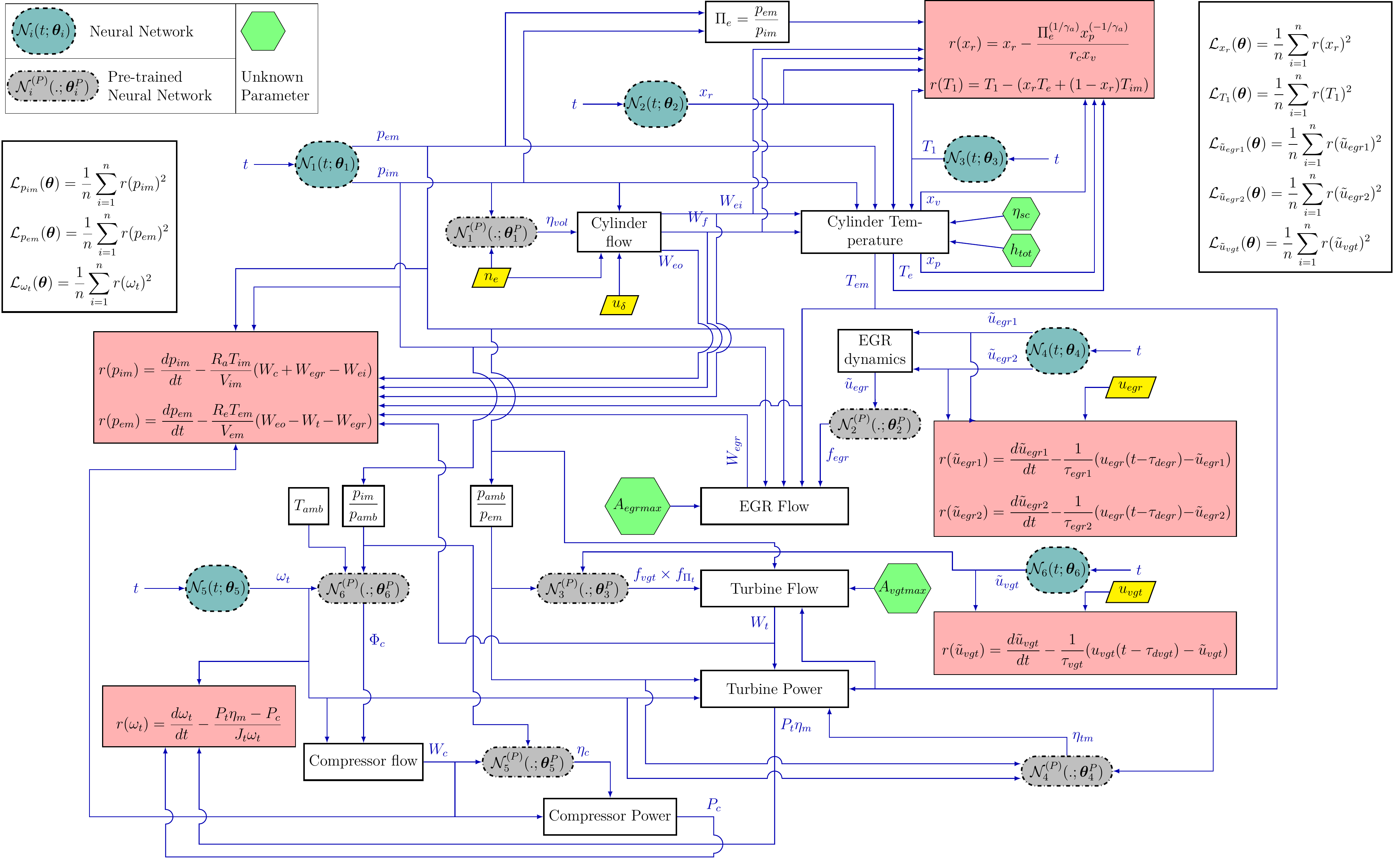}
	\caption{\textbf{Flow chart for the proposed PINNs model:} (Detailed caption is added on the other page).}
	\label{Fig:PINN diesel engine}
	\end{figure}
\end{landscape}

In section \ref{Section:Problem setup}, we discussed the problem setup, and in subsequent sections, we discussed the approximation of different variables using neural networks as well as  the basics of the PINN method and the implementation of PINN in the present problem. In Fig. \ref{Fig:PINNs general}, we have shown a schematic diagram along with a pre-trained network for a general ordinary differential equation. In this section, we show a complete flowchart for the calculation of physics loss functions for the engine problem. In Fig. \ref{Fig:PINN diesel engine}, we show the flow chart for calculating the physics-informed loss for the present problem. Note that we have not shown the data loss and the self-adaptive weights in the flow chart.
\begin{center}\fbox{\begin{minipage}{0.98\textwidth}
	\underline{Detailed Caption for Fig. \ref{Fig:PINN diesel engine}}\\
	Fig.\ref{Fig:PINN diesel engine}: \textbf{Flow chart for proposed PINNs model:} Flow chart for the proposed PINN model for the inverse problem for the engine for prediction of dynamics of the system variables and estimation of unknown parameters. The inputs are input control vector $\{u_\delta, u_{egr}, u_{vgt}\}$ and engine speed $n_e$.
	\vspace*{-2mm}
	\begin{itemize}[leftmargin=*]
	\setlength\itemsep{0cm}
	\item Six neural network $\mathcal{N}_i(t;\bm{\theta})$, $i=1,2,\dots,6$ indicated in dashed rectangular oval takes time $t$ as input and predict $p_{im}$, $p_{em}$, $x_r$ and $T_1$ $\tilde{u}_{egr1}$, $\tilde{u}_{egr2}$, $\omega_{t}$ and $\tilde{u}_{vgt}$ as shown in Table \ref{Table:FNN for PINN}.
	\item Four unknown parameters indicated in hexagon are $\eta_{sc}$, $h_{tot}$, $A_{egrmax}$ and $A_{vgtmax}$.
	\item Six pre-trained neural networks $\mathcal{N}_i^{(P)}(.;\bm{\theta})$, $i=1,2,\dots,6$ indicated in dashed-dotted rectangular oval takes appropriate input and predict the empirical formulae as shown in Table \ref{Table:Surrogate empirical formulae}. The parameters (weights and biases) of these pre-trained DNNs are kept fixed to predict the empirical formulae.
	\end{itemize}
	\vspace*{-2mm}
    There are eight main blocks calculating different variables. The equations for the calculation of each of the quantities are shown in Appendix \ref{Appendix:Engine model}.
	\vspace*{-2mm}
	\begin{itemize}[leftmargin=*]
	\setlength\itemsep{0cm}
	\item \textbf{Cylinder flow:} calculates $W_{ei}$, $W_f$ and $W_{eo}$ using Eqs. \ref{Eq:Append:W_ei}, \ref{Eq:Append:W_f} and \ref{Eq:Append:W_eo} respectively.
	\item \textbf{Cylinder Temperature:} calculates $x_v$, $x_p$, $T_e$ and $T_{em}$ using Eqs. \ref{Eq:Append:x_v}, \ref{Eq:Append:x_p}, \ref{Eq:Append:T_e} and \ref{Eq:Append:T_em} respectively. $h_tot$ and $\eta_{sc}$ are considered as learnable parameters in the calculation of $T_{em}$ and $T_e$ respectively.
	\item \textbf{EGR dynamics:} calculated $\Tilde{u}_{egr}$ using Eq. \ref{Eq:Append:u_egr}
	\item \textbf{EGR Flow:} calculates EGR mass flow $W_{egr}$ using Eq. \ref{Eq:Append:W egr}. $A_{egrmax}$ is considered as learnable parameter.
	\item \textbf{Compressor flow:} calculates compressor mass flow $W_c$ using Eq. \ref{Eq:Append:W_c}
	\item \textbf{Compressor Power:} calculates compressor power $P_c$ using Eq. \ref{Eq:Append:P_c}
	\item \textbf{Turbine Flow:} calculates turbine mass flow $W_t$ using Eq. \ref{Eq:Append:W_t}. $A_{vgtmax}$ is considered as trainable parameter.
	\item \textbf{Turbine Power:} calculates effective turbine power $P_t\eta_m$ using Eq. \ref{Eq:Append:P_t_eta_m}
	\end{itemize}
	\vspace*{-2mm}
	There are five blocks, which calculate the residual of the equation. The first block calculates the residual for state equations for $p_{im}$ and $p_{em}$; the second one calculates the residual for the equations of $x_r$ and $T_1$; the third block calculates the residuals for state equations for $\tilde{u}_{egr1}$ and $\tilde{u}_{egr2}$; the fourth block calculates the residual for the state equation for $\tilde{u}_{vgt}$, and the fifth block calculates the residual for the state equation for $\omega_t$. There are another two blocks, which calculate the physics loss. The first one calculates the physics loss corresponding to state variable $p_{im}$, $p_{em}$ and $\omega_t$. The second block calculates state physics loss corresponding to state variables $\tilde{u}_{egr1}$, $\tilde{u}_{egr2}$, $\tilde{u}_{vgt}$ and physics loss corresponding to $x_r$ and $T_1$. The data losses can be calculated from the variables calculated from the appropriate blocks.
	\end{minipage}}
\end{center}
\subsection{Data generation}
\label{Section:Data acquisition}
We now discuss the generation of data for training the NNs utilized in this study. Specifically, we  have mainly two different types of data here: (1) the data collected from the laboratory that are used to train the DNN surrogates to replace the empirical formulae used in \cite{Wahlstrom}; and (2) field data $p_{im}$, $p_{em}$, $\omega_t$ and $W_{egr}$.
\par In laboratory experiments, we have measurements on state variables, some of which can be employed for training the neural network surrogates for the empirical formulae. The laboratory data required to calculate the labelled data for each of the surrogates are shown in Table \ref{Table:Data lab test}. The calculation of labelled data from laboratory data is discussed in Appendix \ref{Appendix:Data for pre-trained network}. After training, we consider these pre-trained surrogates in field experiments in place of the empirical formulae. The parameters of these networks are kept constant in the PINNs model for the field experiment.
\par In field experiments, we have records for the four inputs, i.e., $\bm{u} = \{u_\delta,\;\; u_{egr}, \;\; u_{vgt}\}$ and $n_e$. In addition, we only have measurements on four variables in field experiments, i.e., the intake manifold pressure ($p_{im}$), exhaust manifold pressure ($p_{em}$), turbine speed ($\omega_t$) and EGR mass flow ($W_{egr}$).

In both the laboratory and field experiments, we have the records for the inputs (i.e., $\bm{u} = \{u_\delta,\;\; u_{egr}, \;\; u_{vgt}\}$ and $n_e$) from an actual engine running conditions. Considering that we only have a certain number of records for the variables in the running engine, which cannot be used to verify our PINN model since our objective is to use it to predict the whole gas flow dynamics in the engine. We, therefore, take the records for the real inputs (i.e., $\bm{u} = \{u_\delta,\;\; u_{egr}, \;\; u_{vgt}\}$ and $n_e$) and employ them as the inputs for the governing equations Eqs. \ref{Eq:p_im}-\ref{Eq:u_vgt}. We then solve these equations using Simulink to obtain the dynamics for all variables. We use the data from Simulink to mimic the real-world measurements,  which are employed as the training data for PINNs and the DNN for the pre-trained networks. The remaining data are used as the validation data to test the accuracy of PINN for reconstructing the gas dynamics in a running engine, given partial observations. Given that the real measurements are generally noisy, we add $3\%$, $3\%$, $1\%$ and $10\%$ Gaussian noise in $p_{im}$, $p_{em}$, $\omega_t$ and $W_{egr}$, respectively. These different signals have different noise values because they are different measurements with different noise characteristics.

\begin{table}[!htb]
\centering
\caption{List of empirical formulae represented using a pre-trained neural network and lab test data required for their training $^{\dagger}$.}
\label{Table:Data lab test}
\begin{tabular}[b]{L{5.cm}C{1.75cm}C{7.5cm}} \hline
    \multicolumn{1}{c}{Empirical quantities $^{\dagger\dagger}$\textsuperscript{,}$^{\dagger\dagger\dagger}$} & Symbol & Laboratory test data required $^{\dagger\dagger\dagger\dagger}$ \\ \hline
    Volumetric efficiency (\S \ref{Appendix:Cylinder}) & $\eta_{vol}$ & \begin{tabular}{L{6.75cm}} \textbullet\ Intake manifold pressure ($p_{im}$) \\
    \textbullet\ Engine speed ($n_e$) \\ 
    \textbullet\ Total mass flow from the intake manifold into the cylinders ($W_{ei}$)  \\
    \textbullet\ Intake manifold temperature ($T_{im}$)
\end{tabular}\\ \hline
Effective area ratio function for EGR (\S \ref{Appendix:EGR valve})  & $f_{egr}$ & 
\begin{tabular}{L{6.75cm}}
    \textbullet\ EGR position ($\tilde{u}_{egr}$) \\
    \textbullet\ EGR mass flow ($W_{egr}$)  \\
    \textbullet\ Exhaust manifold pressure ($p_{em}$) \\
    \textbullet\ Intake manifold pressure ($p_{im}$) \\
    \textbullet\ Exhaust manifold temperature ($T_{em}$) \\ 
\end{tabular}\\ \hline
Effective area ratio function for VGT ($f_{vgt}$) and chocking function ($f_{\Pi_t}$) (\S \ref{Appendix:Turbocharger}) & $f_{vgt}\times f_{\Pi_t}$ & 
\begin{tabular}[b]{L{6.75cm}}
    \textbullet\ VGT position ($\tilde{u}_{vgt}$) \\
    \textbullet\ Exhaust manifold pressure ($p_{em}$)\\
    \textbullet\ Ambient pressure ($p_{amb}$) \\
    \textbullet\ Turbine mass flow ($W_t$) \\
    \textbullet\ Exhaust manifold pressure ($p_{em}$) \\
    \textbullet\ Exhaust manifold temperature ($T_{em}$) \\
\end{tabular}\\ \hline
Turbine mechanical efficiency $^{\dagger\dagger\dagger\dagger\dagger}$ (\S \ref{Appendix:Turbocharger}) & $\eta_{tm}$ & 
\begin{tabular}{L{6.75cm}}
    \textbullet\ Turbine speed ($\omega_t$) \\
    \textbullet\ Exhaust manifold temperature ($T_{em}$) \\
    \textbullet\ Exhaust manifold pressure ($p_{em}$) \\
    \textbullet\ Ambient pressure ($p_{amb}$) \\
    \textbullet\ Compressor mass flow ($W_c$) \\
    \textbullet\ Compressor temperature ($T_c$) \\
    \textbullet\ Ambient temperature ($T_{amb}$) \\
    \textbullet\ Turbine mass flow ($W_t$) \\
\end{tabular}\\ \hline
Compressor efficiency (\S \ref{Appendix:Compressor}) & $\eta_c$ &
\begin{tabular}{L{6.75cm}}
    \textbullet\ Intake manifold pressure ($p_{im}$) \\
    \textbullet\ Compressor mass flow ($W_c$) \\
    \textbullet\ Temperature after the compressor ($T_c$) \\
    \textbullet\ Ambient temperature ($T_{amb}$) \\
    \textbullet\ Ambient pressure ($p_{amb}$) \\
\end{tabular}\\ \hline
Volumetric flow coefficient for compressor (\S \ref{Appendix:Compressor}) & $\Phi_c$ & 
\begin{tabular}[b]{L{6.75cm}}
    \textbullet\ Turbine speed ($\omega_t$) \\
    \textbullet\ Compressor mass flow ($W_{c}$) \\
    \textbullet\ Intake manifold pressure ($p_{im}$)\\
    \textbullet\ Ambient temperature ($T_{amb}$) \\
    \textbullet\ Ambient pressure ($p_{amb}$) \\
\end{tabular}\\ \hline
\multicolumn{3}{l}{$^{\dagger}$ It is assumed that the parameters/constant are known, however not the coefficients for} \\
& & \multicolumn{1}{r}{the empirical formulae.}\\
\multicolumn{3}{l}{$^{\dagger\dagger}$ The definition of the quantities are discussed in relevant sections in Appendix \ref{Appendix:Engine model}.} \\
\multicolumn{3}{l}{$^{\dagger\dagger\dagger}$ The calculations of the empirical quantify from the laboratory data are included } \\
& & \multicolumn{1}{r}{in Appendix \ref{Appendix:Data for pre-trained network}.} \\
\multicolumn{3}{l}{$^{\dagger\dagger\dagger\dagger}$ A brief discussion on instrumentation and test procedure is included in Appendix \ref{Appendix:Lab test data measure}.} \\
\multicolumn{3}{l}{$^{\dagger\dagger\dagger\dagger\dagger}$ For calculation of $\eta_{tm}$, dynamic data are required (discussed in Appendix \ref{Appendix:Data for pre-trained network} and} \\
& & \multicolumn{1}{r}{section \ref{Subsection:Traning of pretrained network}).}\\ \hline
\end{tabular}
\end{table}

In the present study, we consider two sets of input data in the training and testing of the surrogate neural networks for the empirical formulae. 
The first set of data (Set-I) is two (2) hours of data collected at a sampling rate of 1 sec. This control input vector $\{u_\delta, u_{egr}, u_{vgt}\}$ and $n_e$ are considered to generate simulated data with different ambient conditions, which are shown in Table \ref{Table:Pretrained data}. The second set of data (Set-II) is twenty-minute (20 minutes) data collected at a sampling rate of 0.2 sec. This control input vector $\{u_\delta, u_{egr}, u_{vgt}\}$ and $n_e$ are considered to generate simulated data with Case-V ambient conditions.
\par The labelled data for the training of surrogate neural network for $\eta_{vol}$, $F_{vgt,\Pi_t}$, $\eta_c$ and $\Phi_c$ are generated for Case-I to Case-IV with a $dt = 0.2$ sec. The testing data are generated for Case-V with the same $dt$. We observed from the engine model that the EGR valve actuator is independent of the other system of the engine and depends only on the EGR control signal ($u_{egr}$). Thus, for the training of surrogate neural network for $f_{egr}$ ($\mathcal{N}_2^{(P)}(:,\bm{\theta})$), we consider the training data set corresponding to Case-I only and the testing data set corresponding to Case-V. The labelled data for $\eta_{tm}$ are calculated from Eq. \eqref{Eq:Append:omega_t} (\S \ref{Appendix:Turbocharger}), which is a differential equation, thus requires a finer $dt$. The simulated data for the calculation of labelled $\eta_{tm}$ are generated with $dt = 0.025$ sec in all the Cases. We assume that the Set-I data for the input control vector includes a good operating range for training surrogate neural networks for the empirical formulae. The field data ($p_{im}$, $p_{em}$, $\omega_t$ and $W_{egr}$) for the inverse problem are considered from Case-V.
\begin{table}[H]
    \centering
	\caption{\textbf{Ambient conditions for training and testing of neural networks:} The different ambient conditions are considered for generating training and testing data. The input data Set-I is two hours of input control vector $\{u_\delta, u_{egr}, u_{vgt}\}$ and $n_e$ collected from an actual engine running condition. Similarly, Set-II is twenty-minute of input control vector $\{u_\delta, u_{egr}, u_{vgt}\}$ and $n_e$ collected from an actual engine running condition. Case-I to Case-IV are considered for the training of the surrogate neural network for the empirical formulae ($\mathcal{N}_i^{(P)}(:,\bm{\theta}_i)$), while Case-V is considered for testing of these networks. The data for the field data are also considered from Case-V.}
	\label{Table:Pretrained data}
	\begin{tabular}{C{1.5cm}|C{3.15cm}C{3.5cm}C{1.25cm}C{1.75cm}C{1.75cm}}\hline
	\multirow{2}{*}{Case} & $T_{amb}$ & $p_{amb}\times10^5$ (Pa) & \multirow{2}{*}{Input} & Sampling   & \multirow{2}{*}{Purpose} \\ 
	&  (kelvin) &  (Approx. elevation) & & $dt$ & \\ \hline
	Case-I & 233.15 ($-40^\circ$ C) & $0.7000$ (at 3000 m) & Set-I & 1 sec & Training\\
	Case-II & 233.15 ($-40^\circ$ C) & $1.0111$ (at 17.9 m) & Set-I & 1 sec & Training\\
	Case-III & 270.15 ($-3^\circ$ C) & $0.7000$ (at 3000 m) & Set-I & 1 sec & Training\\
	Case-IV & 313.15 ($40^\circ$ C) & $1.0111$ (at 17.9 m) & Set-I & 1 sec & Training\\ \hline
	Case-V & 298.15 ($25^\circ$ C) & $0.8000$ (at 1837 m) & Set-II & 0.2 sec & Testing \\ \hline
	\end{tabular}
\end{table}
\section{Results and discussions}
\label{Section:Results and discussions}
In this section, we demonstrate the applicability of proposed PINNs for solving the inverse problems discussed in section \ref{Section:Problem setup}. Case 1 and Case 2 have three unknown parameters, while Case 3 to Case 5 have four unknown parameters. The predicted values of the unknowns for all five cases are shown in Table \ref{Table:pinn_inverse parameter}. In this section, we will discuss the results of Case 3 to Case 5. The results of Case 1 and Case 2 are presented in Appendix \ref{Appendix:Additiona Figure}. 
\begin{table}[!htb]
\centering
\caption{\textbf{Predicted unknowns:} Predicted unknown parameters for different cases considered.}
\label{Table:pinn_inverse parameter}
\begin{tabular}{C{1.25cm}|C{2.15cm}C{1.25cm}C{1.25cm}C{2.25cm}|L{2.5cm}|L{2.75cm}}
\hline
\multicolumn{1}{c|}{} & $A_{egramx}$ & $\eta_{sc}$ & $h_{tot}$ & $A_{vgtmax}$ & Known variables & Predicted variables \\ \hline
\multicolumn{1}{c|}{True} & $4\times10^{-4}$ & $1.102$ & $96.28$ & $8.456\times10^{-4}$ & \\ \hline
Case 1 & $3.93\times 10^{-4}$ & $1.12$ & $110$ & NA & Clear data of $p_{im}$, $p_{em}$, $\omega_{t}$, $W_{egr}$ & \multirow{5}{2.75cm}{The neural networks predict: $p_{im}$, $p_{em}$, $\tilde{u}_{vgt}$, $\tilde{u}_{egr1}$, $\tilde{u}_{egr2}$, $T_1$, $x_r$. The pretrained neural networks predict: $\eta_{vol}$, $\eta_{tm}$, $\eta_{c}$, $\Phi_c$, $F_{vgt,\Pi_t}$, $f_{egr}$. Other variables are derived from these predicted quantities.} \\ \cline{1-6} 
Case 2 & $3.93\times 10^{-4}$ & $1.12$ & $109$ & NA & Noisy data of $p_{im}$, $p_{em}$, $\omega_{t}$, $W_{egr}$ & \\ \cline{1-6} 
Case 3 & $3.61\times 10^{-4}$ & $0.962$ & $113$ & $7.86\times 10^{-4}$ & Clear data of $p_{im}$, $p_{em}$, $\omega_{t}$, $W_{egr}$  & \\  \cline{1-6} 
Case 4 & $3.51\times 10^{-4}$ & $0.834$ & $134$ & $7.27\times10^{-4}$ & Noisy data of $p_{im}$, $p_{em}$, $\omega_{t}$, $W_{egr}$ & \\ \cline{1-6}  
Case 5 & $2.28\times 10^{-4}$ & $0.829$ & $140$ & $7.27\times 10^{-4}$ & Case 3 without self-adaptive weights & \\ \hline 
\multicolumn{7}{c}{Mask and scale considered} \\
\hline
Mask & Exponential & Softplus & Exponential & Exponential & \multicolumn{2}{L{5.8cm}}{For faster convergence and to have positive value} \\ \hline
Scale & $\times 10^{-4}$ & $\times 1$ & $\times 10$ & $\times 10^{-4}$ & \multicolumn{2}{L{5.8cm}}{Scale to obtain the parameters in physical domain}\\ \hline
\end{tabular}
\end{table}
\par First, we study the results of Case 3 and Case 5 to understand the applicability of PINN and the importance of self-adaptive weight in accuracy and convergence. Then, we study the results of Case 4, which is similar to Case 3; however, with added noise in the field data considered. We also discuss the results for the surrogate for the empirical formulae in Appendix \ref{Subsection:Traning of pretrained network}. Note that the results for all variables are presented in a normalized scale from zero to one using the following equation,
\begin{equation}
	x_{scale} = \dfrac{x - x_{min}}{x_{max} - x_{min}},
	\label{Eq:Scaling for plot}
\end{equation}
where $x$ and $x_{scale}$ are the data before and after scaling, respectively, $x_min$ is the minimum value of true data of $x$ within the time span considered, $x_{max}$ is the maximum value of true data of $x$ within the time span considered.
\par We are considering the input control vector $\{u_\delta, u_{egr}, u_{vgt}\}$ and engine speed $n_e$ from an actual field record. It is assumed that these data have inherent noise in their records. Detailed studies are carried out considering a 1-minute duration. The number of residual points considered in the physics-informed loss and data loss is 301 at equal $dt=0.2$ sec. The initial conditions considered for $\{p_{im},\;\; p_{em},\;\; x_r,\;\; T_1,\;\; \tilde{u}_{egr1},\;\; \tilde{u}_{egr2},\;\; \omega_t, \;\;\tilde{u}_{vgt}\}$ are $\{8.0239\times 10^4,\;\;  8.1220\times 10^4,\;\;  0.0505,\;\;  305.3786,\;\;  18.2518,\;\;  18.1813, \;\; 1.5827\times 10^3,\;\;  90.0317\}$ respectively. The measured field data are also considered for 1 min with equal $dt = 0.2$ sec. Thus, each of the measured field quantities has 301 records.
\par The details of the neural networks considered for the PINN problem are shown in Table \ref{Table:Network size}. We consider $\sigma(.)=\text{tanh}(.)$ activation function for hidden layers for all the neural networks. We would also like to emphasize that the scaling of output is one of the important considerations for faster and an accurate convergence of the neural network. Furthermore, output transformation is another important consideration. The outputs are physical quantity and always positive. The governing equations are valid only for positive quantities (e.g. in Eq. \ref{Eq:Append:W egr}, a negative $p_{im}$ will result in negative $W_{egr}$). The output transformation will ensure that the predicted quantities are always positive in each epoch. Similarly, as shown in Table \ref{Table:pinn_inverse parameter}, the mask for the unknown parameters will ensure a positive value. We also observed that the unknown parameters are of different scales. The scale considered for the unknown parameters will ensure the optimization of these parameters is on the same scale. The parameters of the neural network are optimized first using Adam optimized in Tensorflow-1 with $200\times 10^3$ epoch and further with LBFGS-B optimized. It is also important to note that we have considered self-adaptive weights in the proposed method; thus, we considered different optimizers for each set of self-adaptive weights. Further, self-adaptive weights are optimized only during the process of Adam optimization up to $100\times 10^3$ epoch. After $100\times 10^3$ epoch and during the process of optimization using LBFGS-B, the self-adaptive weights are considered constants with the values at $100\times 10^3$ epoch of Adam optimization. The sizes of self-adaptive weight are $301\times1$ for $\bm{\lambda}_{p_{im}}$, $\bm{\lambda}_{p_{em}}$, $\bm{\lambda}_{\omega_{t}}$ and $\bm{\lambda}_{W_{egr}}$. The size of self adaptive weight of $\bm{\lambda}_{T_{1}}$ is $1\times1$. Softplus masks are considered for all the self-adaptive weights.
\begin{table}[H]
    \centering
	\caption{\textbf{Details of neural network for PINNs:} Details of neural networks considered to approximate the state variables and $T_1$ and $x_r$. The input to the neural networks is time $t$ and the activation functions for the hidden layers are $\sigma(.)=\text{tanh}(.)$. The outputs for each network are shown in the ``Output" column. The ``Output transformation" column shows whether the output from the neural network is passed through any other function. The last column, "Scaling", shows the scaling factor to be multiplied by the final output to obtain the variable in physical space. The input to the networks is time 0 to 60 sec and scaled between $[-1,1]$. }
	\label{Table:Network size}
	\begin{tabular}{C{1.75cm}C{3.25cm}C{2cm}C{4cm}C{1.5cm}} \hline
	   Neural network & Network size & Output & Outputs transformation $^{\ddagger\ddagger}$ & Scaling \\ \hline
		$\mathcal{N}_1(t;\bm{\theta}_1)$ & $[1,\;10,\;10,\;10,\;2]$ & $p_{im}, p_{em}$ & $S_P(p_{im})+0.5, S_P(p_{em})$ & $\times 10^5$ \\[0.25cm]
		$\mathcal{N}_2(t;\bm{\theta}_2)$& $[1\;10,\;10,\;1]$ & $x_r$ & $S_P(x_r)$ & $\times0.03$ \\[0.25cm]
		$\mathcal{N}_3(t;\bm{\theta}_3)$ & $[1,\;15,\;15,\;15,\;1]$ & $T_1$ & $S_P(T_1) + 230/300$ & $\times300$\\[0.25cm]
		$\mathcal{N}_4(t;\bm{\theta}_4)$ & $[1,\;10,\;10,\;10,\;2]$ & $\tilde{u}_{egr1},\tilde{u}_{egr2}$ & $S(\tilde{u}_{egr1},\tilde{u}_{egr2})$ & $\times100$ \\[0.25cm]
		$\mathcal{N}_5(t;\bm{\theta}_5)$ & $[1,\;10,\;10,\;1]$ & $\omega_t$ & $S_P(\omega_t)$ & $\times5\times10^3$ \\[0.25cm]
		$\mathcal{N}_6(t;\bm{\theta}_6)$ & $[1,\;10,\;10,\;1]$ & $\tilde{u}_{vgt}$ & $S(\tilde{u}_{vgt})$ & $\times100$ \\[0.25cm] \hline
		\multicolumn{5}{l}{$^{\ddagger\ddagger}\;\;$ $S_p(.)\longrightarrow$ softplus function. $\;\;\;\;\;\;\; S(.)\longrightarrow$ sigmoid function} \\ \hline
	\end{tabular}
\end{table}

\subsection{PINN for the inverse problem with four unknown parameters}
\subsubsection*{Results for Case 3 and Case 5}
We first consider Case 3 and Case 5 in which we have four unknown parameters $A_{egrmax}$, $\eta_{sc}$, $h_{tot}$ and $A_{vgtmax}$. The dynamics of $p_{im}$, $p_{em}$, $\omega_t$ and $W_{egr}$ can be obtained from the corresponding sensor measurements. We then employ the PINN to predict the dynamics for the variables and infer the four unknowns in the system. The difference between the two cases is that in Case 3, we have considered self-adaptive weights, while in Case 5, we have not considered self-adaptive weights. We consider these two cases to study the applicability of PINN and the importance of self-adaptive weights in the present problem.
\par
\begin{figure}[H]
    \centering
    \includegraphics{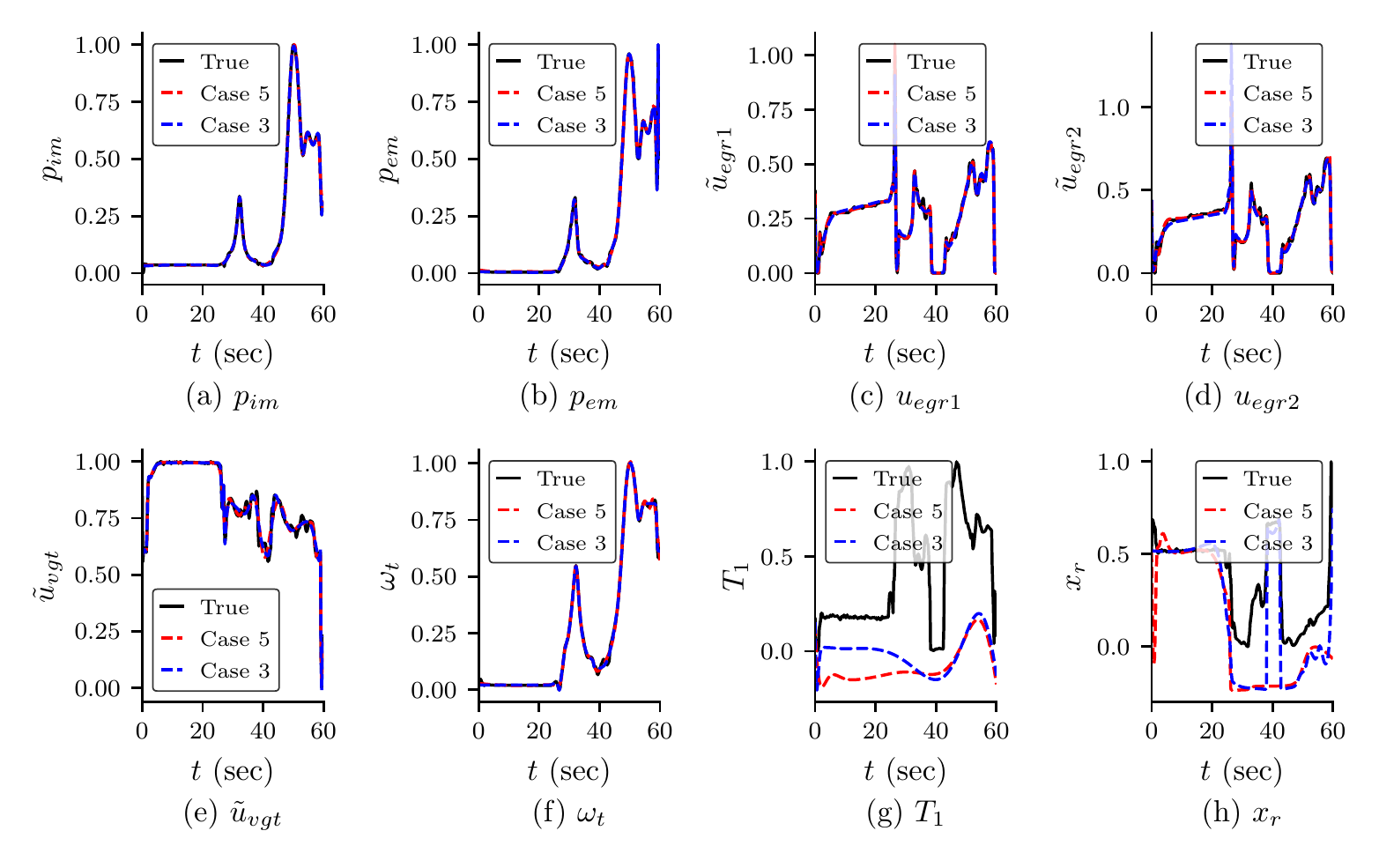}
    \caption{\textbf{Predicted states and $T_1$ and $x_r$ for Case 3 and Case 5:} Predicted dynamics of the state variables of the engine and $T_r$ and $x_1$ for Case 3 (PINN with self-adaptive weights) and Case 5 (standard PINN without self-adaptive weights). The variables are scaled using Eq. \ref{Eq:Scaling for plot}. It can be observed that the predicted dynamics of the states are in good agreement with the true values. However, $T_1$ and $x_r$ do not match with the true value. We study the dependent variables of these two variables, and are shown in Fig. \ref{Figure:Case_3_5_Variable}.}
    \label{Figure:Case_3_5_states}
\end{figure}

\begin{figure}[H]
    \centering
    \includegraphics{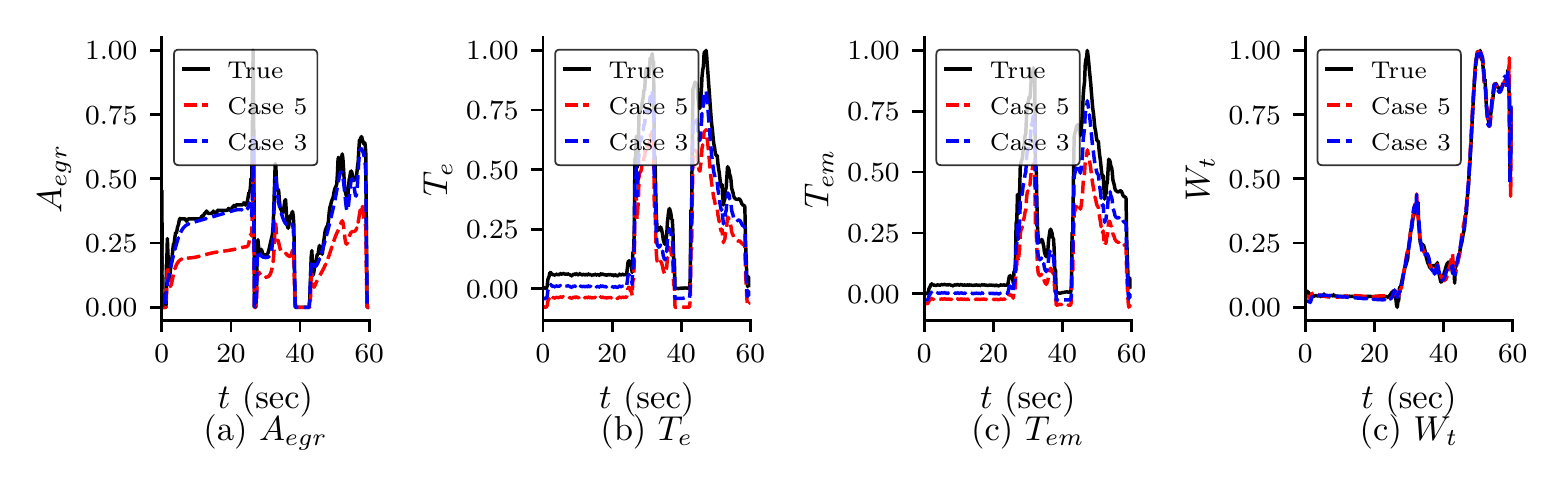}
    \caption{\textbf{Predicted dynamics of dependent variables for Case 3 and Case 5:} Predicted dynamics of $A_{egr}$, $T_e$, $T_{em}$ and $W_t$ for Case 3 and Case 5. These variables depend on the unknown parameters $A_{egrmax}$, $\eta_{sc}$, $h_{tot}$ and $A_{vgtmax}$ respectively. We also note that $T_e$ depends on $T_1$ and $x_r$.}
    \label{Figure:Case_3_5_Variable}
\end{figure}

 The predicted output from the neural networks, i.e., the states and $T_1$ and $x_r$ are shown in Fig. \ref{Figure:Case_3_5_states}. The predicted values of the unknown parameters are shown in Table \ref{Table:pinn_inverse parameter}. We observe that the predicted states are in good approximation with the true value in both cases. However, the predicted $T_1$ and $x_r$ are not in good agreement with the true value. We study the effect of $T_1$ and $x_r$ on the other variables by comparing the predicted dynamics of $T_{e}$ and $T_{em}$ (ref Eqs. \eqref{Eq:Append:T_e} and \eqref{Eq:Append:T_em}). We also note that $T_e$ depends on the unknown $\eta_{sc}$ and $T_{em}$ depends on unknowns $T_e$ and $h_{tot}$. The predicted dynamics of $T_e$ and $T_{em}$ are shown in Fig. \ref{Figure:Case_3_5_Variable}.b and \ref{Figure:Case_3_5_Variable}.c, respectively. We observe that both $T_e$ and $T_{em}$ show somewhat good agreement even $T_1$ and $x_r$ do not match with the true value. The accuracy is more in Case 3 compared to Case 5. We believe that the difference in the true value and the predicted value is due to the error in the predicted value of unknown parameters. We also study the dependent variables $A_{egr}$ and $W_t$ of unknown $A_{egrmax}$ and $A_{vgtmax}$, and are shown in Fig. \ref{Figure:Case_3_5_Variable}.a and \ref{Figure:Case_3_5_Variable}.d, respectively. We observe that in Case 3, the predicted dynamics for both variables show good agreement with true value. However, in Case 5, the $A_{egr}$ does not show good agreement with true value. This is because the predicted value of $A_{egrmax}$ has more error than Case 3.
\par
\begin{figure}[H]
    \centering
    \includegraphics[width=1\textwidth]{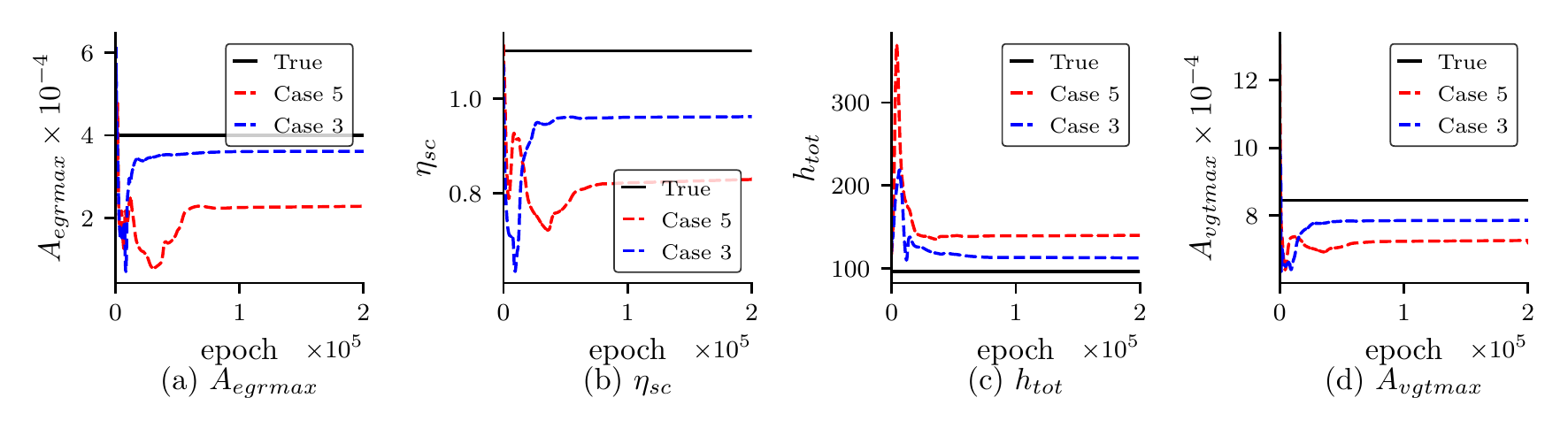}
    \caption{\textbf{Convergence of the unknown parameters for Case 3 and Case 5:} Convergence of the unknown parameters with epoch for Case 3 (PINN with self-adaptive weights) and Case 5 (standard PINN without self-adaptive weights). It is observed that Case 3 converges faster and also shows better accuracy.}
    \label{Figure:Case_3_5_convergence}
\end{figure}

\begin{figure}[H]
    \centering
    \includegraphics{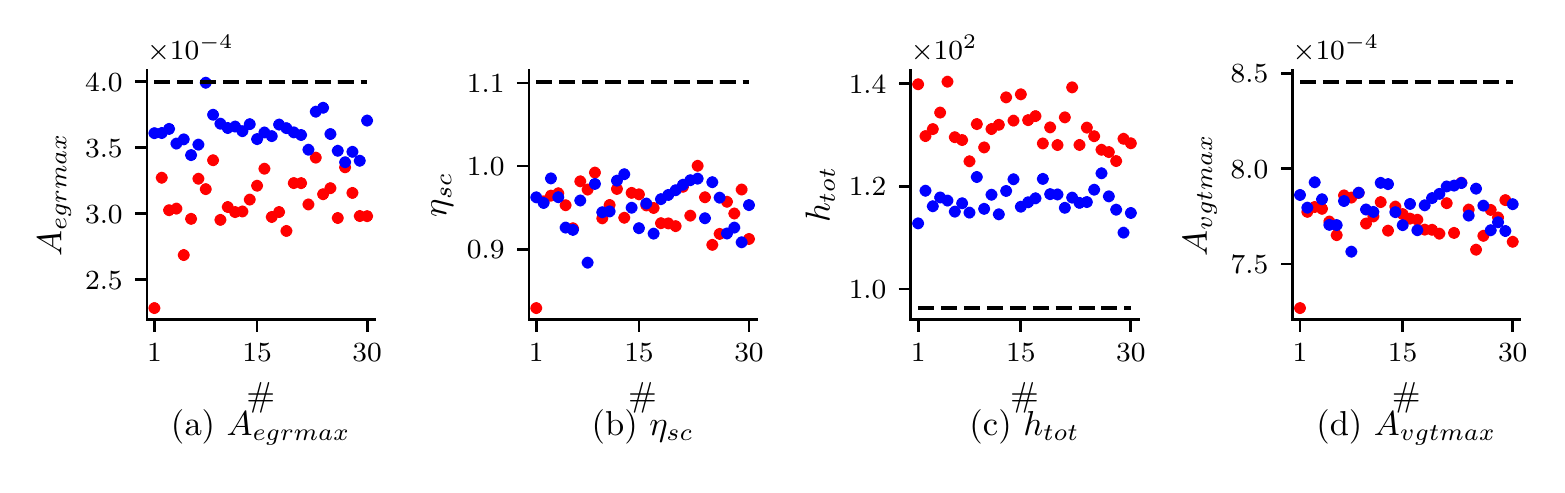}
    \RawCaption{\caption*{Black dashed line $\rightarrow$ true value, Blue dots $\rightarrow$ Case 3, Red dots $\rightarrow$ Case 5}}
    \caption{\textbf{Predicted unknown parameters for Case 3 and Case 5:} Predicted unknown parameters for Case 3 (PINN with self-adaptive weights) and Case 5 (standard PINN) when prediction is made multiple times with different initialisation of the parameters of PINN, self-adaptive weights, and the unknown parameters.}
    \label{Figure:Case_3_5_deepensumble}
\end{figure}

\begin{figure}[H]
    \centering
    \includegraphics{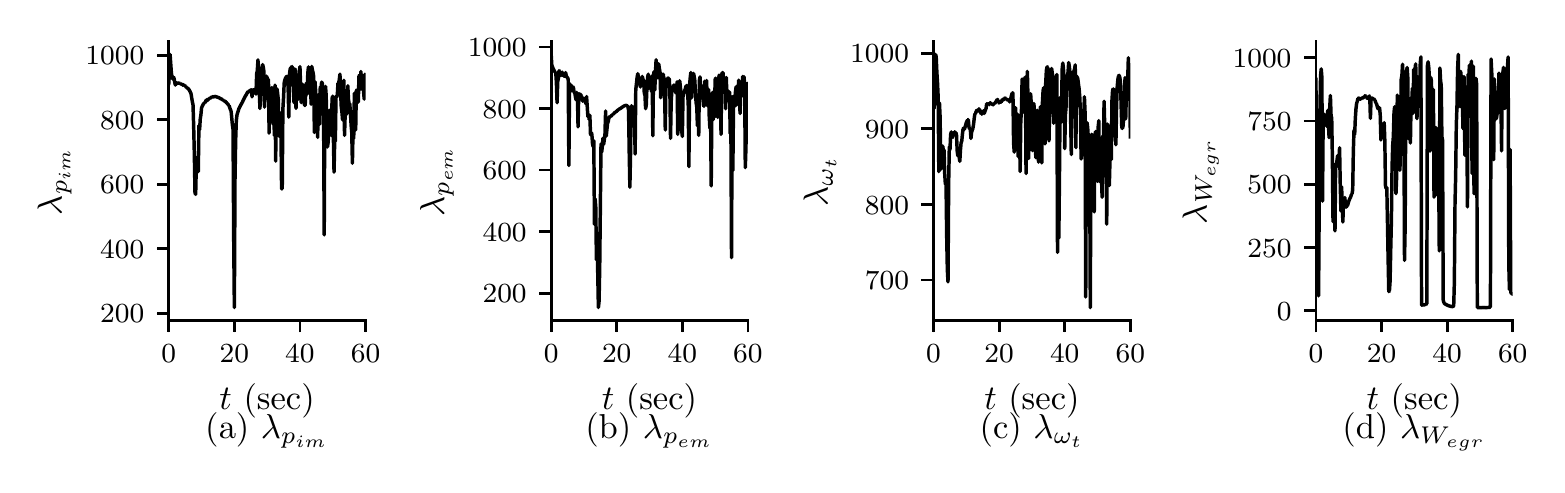}
    \caption{\textbf{Self-adaptive weights for Case 3:} Self-adaptive wights for $p_{im}$, $p_{em}$, $\omega_{t}$ and $W_{egr}$ after $100\times 10^3$ epoch of Adam optimization. The values of the self-adaptive weights after $100\times 10^3$ Adam optimization and LBFGS-B optimization are constant with the values of self-adaptive weight at $100\times 10^3$ epoch.}
    \label{Figure:Case_3 self adaptive weight}
\end{figure}

 In order to study the importance of self-adaptive weights, we study the convergence of the unknown parameters for both cases with self-adaptive weight (Case 3) and without self-adaptive weight (Case 5). The convergences of the unknown parameters with epoch for both cases are shown in Fig. \ref{Figure:Case_3_5_convergence}. In Case 3 (with self-adaptive weights), we can observe that the unknown parameters converge faster and are more accurate. Furthermore, we also study the effect of different initialization of network parameters for PINN and self-adaptive weights. We run the PINN model for Case 3 and Case 5 with different initialization of parameters of PINN (DNN and unknown parameters) and self-adaptive weight keeping other hyperparameters (number of epoch considered, learning rate scheduler etc.) the same. The results for both cases are shown in Fig. \ref{Figure:Case_3_5_deepensumble}. It is observed that for unknowns, $\eta_{sc}$ and $A_{vgtmax}$ for both cases show similar accuracy. However, for unknowns, $A_{egrmax}$ and $h_{tot}$, Case 3, which is with self-adaptive weights, shows better accuracy than Case 5 (without self-adaptive weights) for all the runs. In Fig. \ref{Figure:Case_3 self adaptive weight}, we show the self-adaptive weights for $p_{im}$, $p_{em}$, $\omega_{t}$ and $W_{egr}$ after $100\times 10^3$ epoch (constant value after $100\times 10^3$ epoch). Thus, we conclude that self-adaptive weights are important for better accuracy and convergence for the present problem.

\subsubsection*{Results for Case 4: Four unknowns with noisy measurement data}
In the previous section, we have shown the effectiveness of PINN and the importance of self-adaptive weights. In this section, we test the robustness of the proposed PINN formulation for predicting the gas flow dynamics of the diesel engine given noisy data. In particular, we are considering Case 4 (the same Case 3 but noisy measure data), in which we have four unknown parameters $A_{egrmax}$, $\eta_{sc}$, $h_{tot}$ and $A_{vgtmax}$, with noise measurement of $p_{im}$, $p_{em}$, $\omega_{t}$, $W_{egr}$.
\par
\begin{figure}[H]
    \centering
    \includegraphics[width=1\textwidth]{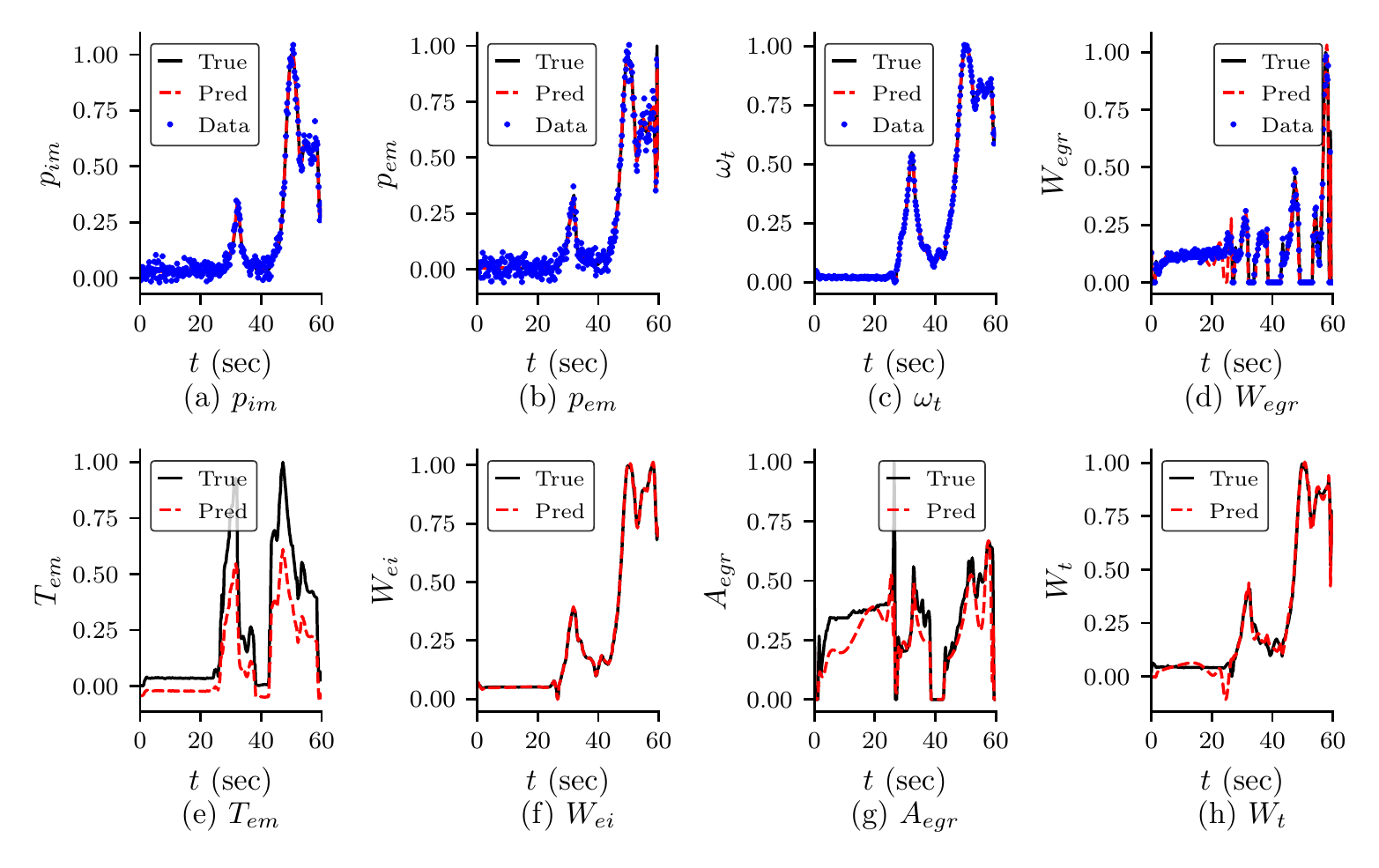}
    \caption{\textbf{Predicted dynamics for variables for Case 4:} Predicted dynamics of (a)-(d) variables whose noisy field measurements are known. (e)-(h) dynamics of other important variables, which are also dependent on the unknown parameters. These results are for Case 4 with 4 unknown parameters and noisy field measurements. }
    \label{Figure:Case 4 dynamics}
\end{figure}

\begin{figure}[H]
    \centering
    \includegraphics[width=1\textwidth]{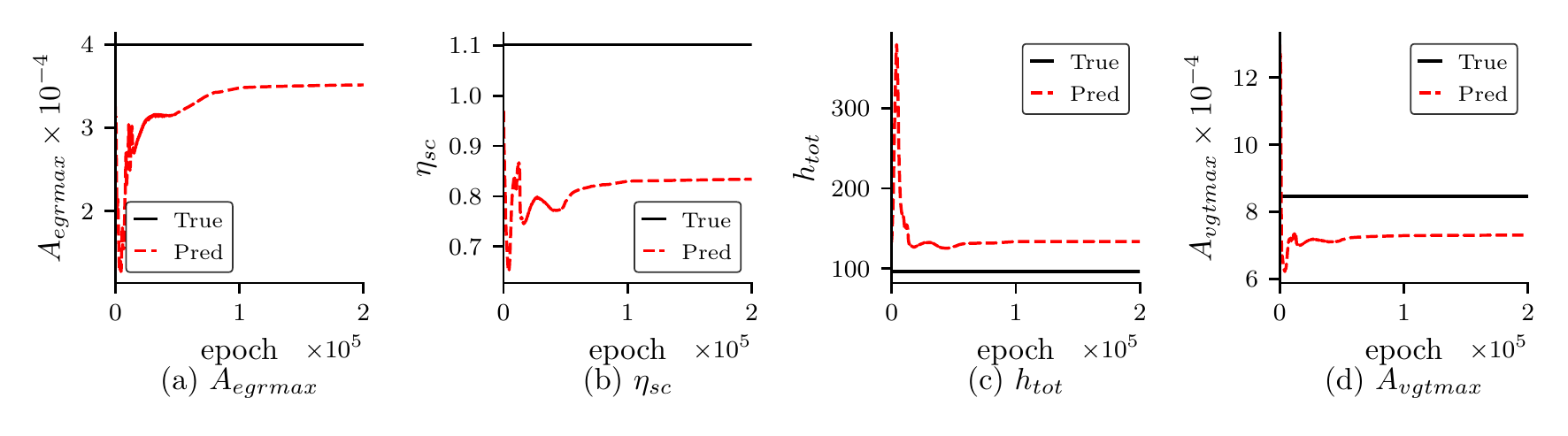}
    \caption{\textbf{Convergence of the unknown parameters for Case 4:} Convergence of the unknown parameters with epoch for Case 4 (PINN with self-adaptive weights and noisy field data)}
    \label{Figure:Case 4 parameter convergence}
\end{figure}

 We contaminate the training data $p_{im}$, $p_{em}$, $\omega_{t}$, $W_{egr}$ considered in Case 3 with Gaussian noise and consider these as synthetic field measurements. We present the predicted dynamics of the known data in Fig. \ref{Figure:Case 4 dynamics}(a)-(d) and unknown parameters in Table \ref{Table:pinn_inverse parameter}. We observe that the dynamics of the predicted $p_{im}$, $p_{em}$, $\omega_{t}$, $W_{egr}$ matches with the reference solution. However, in the case of $W_{egr}$, there is a small discrepancy in the predicted values near 20-25 sec, which we can attribute to over-fitting caused by the noisy training data. We study the dynamics of $T_{em}$, $W_{ei}$ $A_{egr}$ and $W_t$ on which we do not have any measured data, and these are shown in Fig. \ref{Figure:Case 4 dynamics}(e)-(h). We observe that $W_{ei}$ matches with the reference results. Most of the dynamics of $A_{egr}$ and $W_t$ match with the reference solution. The mismatch in these two variables may also be attributed to over-fitting caused by noisy data. The profile of $T_{em}$ matches with the reference solution, however, it is not an exact match with the reference solution. This is because of the error in the predicted value of unknown parameter $\eta_{sc}$ and $h_{tot}$. We also note that in the present study, we do not have any temperature measurements of field data. Thus, we expect an error in the predicted temperature measurement. We also study the convergence of the unknown parameters with epoch and shown in Fig. \ref{Figure:Case 4 parameter convergence}. We note that, in this case, we consider the same hyperparameters in the optimization process. In some cases, we see over-feeting due to noisy data. This may be controlled by changing the hyperparameters, specially the learning rate for the self-adaptive weights.
\par
\begin{figure}[H]
    \centering
    \includegraphics[width=1\textwidth]{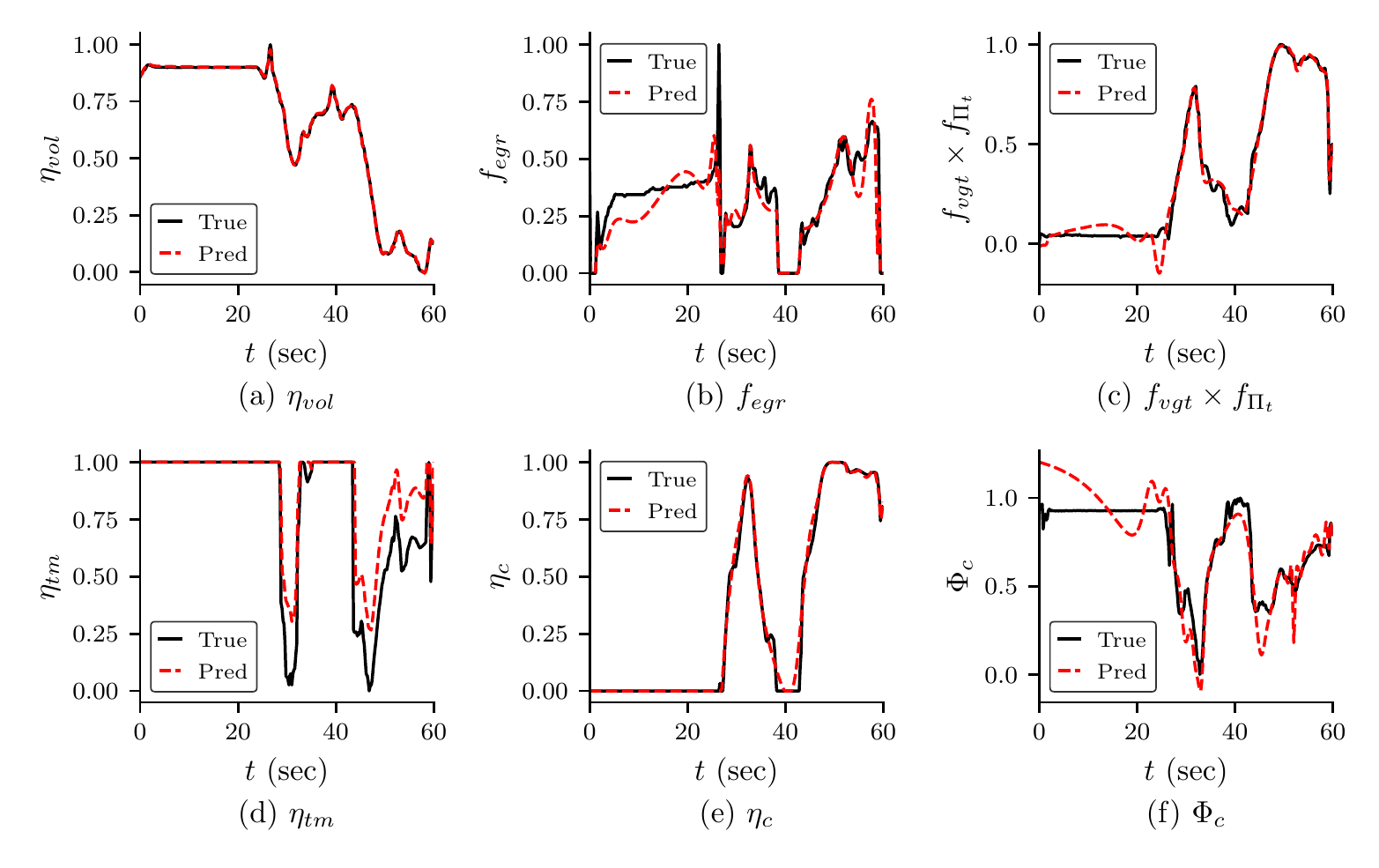}
    \caption{\textbf{Empirical formulae for Case 4:} The prediction of empirical formulae for Case 4 (PINN with self-adaptive weights and noisy field data).}
    \label{Figure:Case 4 empirical formulae}
\end{figure}

 We also study the prediction of empirical formulae in this case and shown in Fig. \ref{Figure:Case 4 empirical formulae}. We observe that $\eta_{vol}$ and $\eta_c$ match with the reference solution. These two quantity gives the volumetric efficiency of the cylinder and the efficiency of the compressor. The other four quantities ($f_{egr}$, $f_{vgt}\times f_{\Pi_t}$, $\eta_{tm}$, $\Phi_c$), also match most of its points. The discrepancy can be attributed to the noisy measurement of field data.

\section{Summary}
\label{Section:Conclusions}
In this study, we proposed a PINNs-based method for estimating unknown parameters and predicting the dynamics of variables of a mean value diesel engine with VGT and EGR, given the measurement of a few of its variables. Specifically, we know field data of intake manifold pressure ($p_{im}$), exhaust manifold pressure ($p_{em}$), turbine speed ($\omega_{t}$) and EGR flow ($W_{egr}$). We predicted the dynamics of the system variables and unknown parameters ($A_{egrmax}$, $\eta_{sc}$, $h_{tot}$ and $A_{vgtmax}$). The input data for the study are considered from actual engine running conditions and show good accuracy in predicted results. We also studied the importance of self-adaptive weight in the accuracy and convergence of results. Furthermore, we also showed how we could approximate empirical formulas for different quantities using neural networks and train them. We believe the proposed method could be considered for an online monitoring system of diesel engines. The field-measured data are collected using individual sensors. Thus, in the event of sensor failure or erroneous data, the method may give erroneous results. The method also does not consider a failure of engine components, e.g. leakage in the EGR valve. We considered the engine model proposed in \cite{Wahlstrom}. Future research may include modelling of failure of engine components. Since the proposed PINN consider online training, with change in input data, field measured data or ambient condition, the PINN networks are required to train again. The accuracy of the results also depends on the size of neural networks and the optimization strategy (e.g. optimizer, learning rate scheduler) considered. For example, a large neural network or higher value in learning rate may result in overfitting of predicted results. The activation function also plays an important role in the accuracy and computational cost \cite{Jagtap_2023_important}. Further study may include a neural architecture search for optimal network sizes considering different operational ranges. Future studies may include a more robust and efficient PINN method for the problem that can be used with edge systems, including proper transfer learning strategies to reduce the computation cost. As there is noise in the measured data, the future study in this regard may also be towards uncertainty quantification of the predicted dynamics and unknown parameters.

\section*{Input data and data generations}
The input data Set-I and Set-II are collected from actual engine running conditions. These data are considered to generate simulated data with different ambient conditions using the Simulink file \cite{Simulink_file} accompany \cite{Wahlstrom}.
\section*{Author Contributions Statement}
\textbf{Kamaljyoti Nath:} Conceptualization, Formal analysis, Investigation, Methodology, Software, Validation, Visualization, Writing – original draft, Writing – review \& editing. \textbf{Xuhui Meng:} Conceptualization, Formal analysis, Investigation, Methodology, Software, Validation, Visualization, Writing – original draft, Writing – review \& editing. \textbf{Daniel J Smith:} Conceptualization, Data curation, Project administration, Supervision, Writing – original draft, Writing – review \& editing. \textbf{George Em Karniadakis:} Conceptualization, Funding acquisition, Project administration, Resources, Supervision, Writing – original draft, Writing – review \& editing. All authors reviewed the manuscript.
\section*{Acknowledgement}
KN, XM and GEK would like to acknowledge the support by Cummins Inc. USA.
\bibliographystyle{agsm}
\bibliography{Reference}

\input{Appendix.tex}

\end{document}

%% file: Appendix.tex
\begin{appendices}
\renewcommand{\thesection}{\Alph{section}}
\renewcommand{\thesubsection}{\Alph{section}.\arabic{subsection}}
\renewcommand{\thesubsubsection}{\Alph{section}.\arabic{subsection}.\arabic{subsection}}
\setcounter{equation}{0}
\renewcommand{\theequation}{\thesection.\arabic{equation}}
\setcounter{table}{0}
\renewcommand{\thetable}{\Alph{section}\arabic{table}}
\setcounter{figure}{0}
\renewcommand{\thefigure}{\Alph{section}\arabic{figure}}
\clearpage
\begin{center}
    \Large \textbf{Appendices} 
\end{center}
\begin{enumerate}[label = \textbf{Appendix \Alph* :}, leftmargin=*]
    \item Note on neural network and training of PINN
    \item Engine model
    \item Brief discussion on lab test data
    \item Calculation of labelled data for training of the neural network for the empirical formulae
    \item Detailed loss function for the PINNs model for the engine
    \item Additional Tables
    \item Additional figures (Results for Case 1 and Case 2)
    \item Neural network surrogates for empirical formulae
\end{enumerate}
\section{Note on neural network and training of PINN}
\label{Appendix:Neural network pinn}
In this section, we present more details on neural networks, PINN and optimization for inverse problems. As shown in Fig. \ref{Fig:PINNs general}, the neural network (FFN/DNN) takes time $t$ as input and approximates the unknown variable $y$. For a DNN with $n-1$ hidden layers, the equation for the neural network can be written as,
\begin{subequations}
	\begin{align}
		\bm{y}_0 & = t \hspace{6.35cm} \text{Input}\\
		\bm{y}_i & = \sigma\left(\bm{W}_i\bm{y}_{i-1} + \bm{b}_i\right)\;\;\;\;\;\; i\;\forall \;1\le i\le n - 1 \hspace{0.75cm} \text{Hidden layers}\\
		\bm{\hat{y}} = \bm{y}_n & = \bm{W}_n\bm{y}_{n-1} + \bm{b}_{n-1} \hspace{4.15cm} \text{Output layer}
	\end{align}
\end{subequations}
where, $\bm{W}$ and $\bm{b}$ are the weights matrices and bias vectors of the network, $\sigma(.)$ is an activation function, which is considered as hyperbolic tangent function in the present study. The output $\hat{y}$ is a function of input $t$ parameterized by the weights $\bm{W}$ and biases $\bm{b}$. We can tune the parameters of the network to predict a large number of snapshots $y$ by minimizing a loss function using an appropriate optimization technique.
\par In the case of a data-driven model of neural networks, the loss function is generally considered as the mean square error (MSE) between the predicted ($\hat{y}$) and the exact value ($y$). On the other hand, in the case of PINN, the neural network output is made to satisfy the differential equation and the initial/boundary conditions. The derivatives of the equation are generally evaluated using automatic differentiation. As discussed in section \ref{Subsecction:PINN general}, the loss function is a weighted sum of physics loss which is MSE of residual and boundary/initial loss, which is MSE between predicted and exact boundary/initial value. The optimal parameters (weights and biases) of the network are obtained using an optimization method such as Adam or L-BFGS-B. In the case of an inverse problem using PINN where the objective is to predict unknown parameters ($\Lambda$) along with the variable ($y$). The unknown parameters are also optimized along with the network parameters. Thus, the trainable parameters are weights, biases and the unknown parameters ($\bm{\theta} = \{\bm{W}, \bm{b}, \bm{\varLambda}\}$). Also, an additional loss function is added, which is data loss between the predicted and the known value of $y$. Furthermore, in the present study we have considered self adaptive wights \cite{McClenny_2020} for the loss function. The loss function is maximized with respect to the self self adaptive weights ($\lambda$). Thus, the optimization process may be written as,
\begin{equation}
	\underset{\bm{\theta}}{\text{min}} \;\;\underset{\bm{\lambda}}{\text{max}}\;\; \mathcal{L}(\bm{\theta},\bm{\lambda})
\end{equation}
Consider the updates of a gradient descent/ascent approach to this problem
\begin{subequations}
	\begin{align}
		\bm{\theta} & = \bm{\theta} - lr_\theta \nabla_\theta \mathcal{L}(\bm{\theta},\bm{\lambda}) \\
		\bm{\lambda} & = \bm{\lambda} + lr_\lambda \nabla_\lambda \mathcal{L}(\bm{\theta},\bm{\lambda}) 
	\end{align}
\end{subequations}
where, $lr_\theta$ and $lr_\lambda$ are the learning rate associated with $\bm{\theta}$ and $\bm{\lambda}$.
\par In the present study, the parameters are optimized using Adam and L-BFGS-B optimizer in Tensorflow-1 (with single precision floating point). Further, self-adaptive weights are optimized only during the process of Adam optimization up to fixed epoch as discussed in section \ref{Section:Results and discussions}
\setcounter{equation}{0}
\renewcommand{\theequation}{\thesection.\arabic{equation}}
\section{Engine model}
\label{Appendix:Engine model}
As discussed in section \ref{Section:Problem setup}, we consider a mean value engine model proposed by \citeauthor{Wahlstrom} \cite{Wahlstrom} in our present study. The engine has eight states, and we have considered six in the present study. These are,
\begin{equation}
    \bm{x} = \{p_{im},\;\; p_{em},\;\; \omega_t,\;\; \tilde{u}_{egr1},\;\; \tilde{u}_{egr2},\;\; \tilde{u}_{vgt}\}
    \label{Eq:Appendix:States}
\end{equation}
where $p_{im}$ and $p_{em}$ are the intake and exhaust manifold pressure, respectively, $\omega_t$ is the turbo speed. $\tilde{u}_{egr1}$ and  $\tilde{u}_{egr2}$ are the two states for the EGR actuator dynamics, and $\tilde{u}_{vgt}$ represents the VGT actuator dynamics. The control inputs for the engine are $\bm{u} = \{u_\delta,\;\; u_{egr}, \;\; u_{vgt}\}$ and the engine speed is $n_e$. Where $u_\delta$ is the mass of injected fuel, $u_{egr}$ and $u_{vgt}$ are the EGR and VGT valve positions, respectively. The mean value engine model is then expressed as
\begin{equation}
    \dot{\bm{x}} = f(\bm{x}, \bm{u}, n_e).
\end{equation}
The engine model consists of 6 parts intake and exhaust manifold, the cylinder, the turbine, EGR valve system, and the compressor system. A schematic diagram of the engine is shown in Fig. \ref{Fig:Schematic diagram of Engine}. In this section, we briefly discuss the equation required for the present study and these are taken from \cite{Wahlstrom}. For detail of the engine model, the interested reader may refer to \citeauthor{Wahlstrom} \cite{Wahlstrom}.
\subsection{Manifold pressures}
\label{Appendix:Manifold pressure}
The pressure at the intake manifold ($p_{im}$) is modelled using a first-order differential equation as,
\begin{equation}
	\dfrac{d}{dt} p_{im} = \dfrac{R_a T_{im}}{V_{im}}\left(W_c+W_{egr} - W_{ei}\right)
    \label{Eq:Append:p_im}
\end{equation}
where $T_{im}$ and $V_{im}$ are the temperature and volume of the intake manifold, respectively, and both are assumed to be constant, $W_c$, $W_{egr}$ and $W_{ei}$ are the compressor mass flow, EGR mass flow and total mass flow, respectively. The ideal gas constant and specific heat capacity of the air are $R_a$ and $\gamma_a$, respectively.
\par Similarly, the exhaust manifold pressure $p_{em}$ is modelled as,
\begin{equation}
	\dfrac{d}{dt} p_{em} = \dfrac{R_e T_{em}}{V_{em}}\left(W_{eo} - W_t - W_{egr}\right)
	\label{Eq:Append:p_em}
\end{equation}
where $R_e$ is the ideal gas constant of the exhaust gas with specific heat capacity $\gamma_e$, $T_{em}$ and $V_{em}$ are the exhaust manifold temperature, and volume, $W_{eo}$, $W_t$ are the mass flow out from the cylinder and turbine mass flow, respectively.
\subsection{Cylinder}
\label{Appendix:Cylinder}
The total mass flow from the intake manifold to the cylinder $W_{ei}$, and the total mass flow out of the cylinder $W_{eo}$ are modelled as,
\begin{align}
	W_{ei} & =  \dfrac{\eta_{vol}p_{im}n_eV_d}{120R_aT_{im}}
	\label{Eq:Append:W_ei} \\
	W_{eo} & = W_f + W_{ei}
	\label{Eq:Append:W_eo}
\end{align}
where $W_f$ is the fuel mass flow into the cylinder is given by,
\begin{equation}
	W_f = \dfrac{10^{-6}}{120}u_\delta n_e n_{cyl},
	\label{Eq:Append:W_f}
\end{equation}
$V_d$, $n_e$ and $n_{cyl}$ are the displaced volume, engine speed and the number of cylinders, respectively. The volumetric efficiency, $\eta_{vol}$ of the cylinder may be modelled as
\begin{equation}
	\eta_{vol} = c_{vol1} \sqrt{p_{im}} +c_{vol2} \sqrt{n_e} + c_{vol3}
       \label{Eq:Append:eta_vol}
\end{equation}
where $c_{vol1}$, $c_{vol2}$ and $c_{vol3}$ are constant.
\par The temperature at cylinder out based upon ideal-gas Seliger cycle (or limited pressure cycle) and given as,
\begin{equation}
	T_e = \eta_{sc}\Pi_e^{1-1/\gamma_a}r_c^{1-\gamma_a}x_p^{1/\gamma_a -1}\left[q_{in}\left(\dfrac{1-x_{cv}}{c_{pa}} + \dfrac{x_{cv}}{c_{Va}} \right) + T_1r_c^{\gamma_a-1}\right]
	\label{Eq:Append:T_e}
\end{equation}
where the pressure ratio ($\Pi_e$) over the cylinder is ratio of pressure at exhaust ($p_{em}$) and intake ($p_{im}$),
\begin{equation}
	\Pi_e = \dfrac{p_{em}}{p_{im}},
	\label{Eq:Append:Pi_e}
\end{equation}
the fuel consumed during constant-volume combustion is $x_{cv}$ and fuel consumed during constant pressure combustion is $1-x_{cv}$, $\eta_{sc}$ and $r_c$ are compensation factor for non-ideal cycles and compression ratio. The temperature, $T_1$ when the inlet valve closes and after the intake stroke and mixing is given by,
\begin{equation}
	T_1 = x_rT_e + (1-x_r)T_{im}
	\label{Eq:Append:T_1}
\end{equation}
The residual gas fraction ($x_r$) is model as
\begin{equation}
	x_r = \dfrac{\Pi_e^{1/\gamma_a}x_p^{-1/\gamma_a}}{r_c x_v}
	\label{Eq:Append:x_r}
\end{equation}
The pressure ratio ($x_p$) and the volume ratio ($x_v$) in the Seliger cycle between point 3 (after combustion) and point 2 (before combustion) are modelled as,
\begin{align}
	x_p = & \dfrac{p_3}{p_2} = 1 + \dfrac{q_{in}x_{cv}}{c_{Va}T_1r_c^{\gamma_a-1}}
	\label{Eq:Append:x_p} \\
	x_v = & \dfrac{v_3}{v_2} = 1 + \dfrac{q_{in}(1-x_{cv})}{c_{pa}\left[(q_{in}x_{cv}/c_{Va})+T_1r_c^{\gamma_a - 1}\right]}
	\label{Eq:Append:x_v}
\end{align}
where the specific energy constant of the charge is modelled as
\begin{equation}
	q_{in} = \dfrac{W_fq_{HV}}{W_{ei}+Wf}(1-x_r)
\end{equation}
The temperature at cylinder out modelled in Eq. \eqref{Eq:Append:T_e} is the temperature at the cylinder exit. However, it is not the same as the temperature as the exhaust manifold. This is due to the heat loss in the exhaust pipes between the cylinder and the exhaust manifold. The exhaust manifold temperature ($T_{em}$) is given as
\begin{equation}
	T_{em} = T_{amb} + (T_e - T_{amb})\exp\left(\dfrac{-h_{tot}\pi d_{pipe}l_{pipe}n_{pipe}}{W_{eo}c_{pe}}\right)
	\label{Eq:Append:T_em}
\end{equation}
where $T_{amb}$ is the ambient temperature, $d_{pipe}$, $l_{pipe}$ and $n_{pipe}$ are the pipe diameter, pipe length and the number of pipes, respectively.
\subsection{EGR valve}
\label{Appendix:EGR valve}
The actuator dynamics of the EGR-valve are modelled as,
\begin{align}
	\dfrac{d}{dt}\Tilde{u}_{egr1} & = \dfrac{1}{\tau_{egr1}}\left[u_{egr}(t-\tau_{degr}) - \Tilde{u}_{egr1}\right]
	\label{Eq:Append:u_egr_1}\\
	\dfrac{d}{dt}\Tilde{u}_{egr2} & = \dfrac{1}{\tau_{egr2}}\left[u_{egr}(t-\tau_{degr}) - \Tilde{u}_{egr2}\right]
	\label{Eq:Append:u_egr_2} \\
	\Tilde{u}_{egr} & = K_{egr}\Tilde{u}_{egr1} - (K_{egr} - 1)\Tilde{u}_{egr2}
	\label{Eq:Append:u_egr}
\end{align}
where $\tau_{egr1}$, $\tau_{egr2}$ are time constants, $\tau_{degr}$ is the time delay and $K_{egr}$ is a constant that affect the overshoot.
\par We model the mass flow through the EGR valve through the restriction ($p_{em}<p_{im}$) as,
\begin{equation}
	W_{egr} = \dfrac{A_{egr}p_{im}\Psi_{egr}}{\sqrt{T_{em}R_e}}
			\label{Eq:Append:W egr}
\end{equation}
where $\Psi_{egr}$ is a parabolic function
\begin{equation}
			\Psi_{egr} = 1 - \left(\dfrac{1 - \Pi_{egr}}{1-\Pi_{egropt}} -1 \right)^2
			\label{Eq:Append:Psi egr}
\end{equation}
The effective area is modelled as,
\begin{equation}
			A_{egr} = A_{egrmax}f_{egr}(\Tilde{u}_{egr})
			\label{Eq:Append:A egr}
\end{equation}
When the sonic conditions are reached (flow is choked) in the throat and when no backflow can occur ($1<p_{im}/p_{em}$), the pressure ratio $\Pi_{egr}$ over the valve is limited and modelled as,
\begin{equation}
		\Pi_{egr} = \begin{cases}
				\Pi_{egropt} \;\;\;\;\;\; & \text{if}\;\; \dfrac{p_{im}}{p_{em}}< \Pi_{egropt} \\
				\dfrac{p_{im}}{p_{em}} \;\;\;\;\;\; & \text{if}\;\; \Pi_{egropt} \le \dfrac{p_{im}}{p_{em}} \le 1 \\
				1 \;\;\;\;\;\; & \text{if} \;\; 1<\dfrac{p_{im}}{p_{em}}
			\end{cases}
			\label{Eq:Append:Pi_egr}
\end{equation}
$A_{egrmax}$, $\Pi_{egropt}$ are constant and $f_{egr}(\Tilde{u}_{egr})$ is modelled as a polynomial function,
\begin{equation}
			f_{egr}(\Tilde{u}_{egr}) = \begin{cases}
				c_{egr1}\Tilde{u}_{egr}^2 + c_{egr2}\Tilde{u}_{egr} + c_{egr3} \;\;\; & \text{if}\;\; \Tilde{u}_{egr}\le -\dfrac{c_{egr2}}{2c_{egr1}} \\
				c_{egr3} - \dfrac{c_{egr2}^2}{4c_{egr1}} \;\;\; & \text{if}\;\; \Tilde{u}_{egr}> -\dfrac{c_{egr2}}{2 c_{egr1}}
			\end{cases}
			\label{Eq:Append:f_egr}
\end{equation}
where $c_{egr1}$, $c_{egr2}$ and $c_{egr3}$ are constant. 
\subsection{Turbocharger}
\label{Appendix:Turbocharger}
\par The turbo speed, $\omega_t$ is modelled as a first-order differential model as,
\begin{equation}
		\dfrac{d}{d t}\omega_t = \dfrac{P_t\eta_m - P_c}{J_t\omega_t}
		\label{Eq:Append:omega_t}
\end{equation}
where $J_t$ is the inertia, $P_t$ and $P_c$ are the power delivered by the turbine and power required to drive the compressor, respectively, $\eta_m$ is the mechanical efficiency of the turbocharger. 
\par The VGT actuator system is modelled as a first-order system
\begin{equation}
    \dfrac{d\Tilde{u}_{vgt}}{dt} = \dfrac{1}{\tau_{vgt}}\left[u_{vgt}(t-\tau_{dvgt}) - \Tilde{u}_{vgt}\right]
	\label{Eq:Append:u_vgt}
\end{equation}
where $\tau_{vgt}$ and $\tau_{dvgt}$ are the time constant and time delay respectively.
\par The turbine mass flow ($W_t$) is calculated using 
\begin{equation}
		W_t = \dfrac{A_{vgtmax}p_{em}f_{\Pi_t}(\Pi_t)f_{vgt}(\Tilde{u}_{vgt}))}{\sqrt{T_{em}R_e}}
		\label{Eq:Append:W_t}
\end{equation}
\par $A_{vgtmax}$ is the maximum area in the turbine that the gas flow through.
\begin{equation}
	f_{\Pi_t}(\Pi_t) = \sqrt{1-\Pi_t^{K_t}}
	\label{Eq:Append:f_Pi_t}
\end{equation}
where $K_t$ a constant and $\Pi_t = p_{es}/{p_{em}}$. $p_{es}>p_{amb}$ if there is a restriction like an after-treatment system. However, in the model we consider, there is no restriction after the turbine, thus
\begin{eqnarray}
			\Pi_t =\dfrac{p_{amb}}{p_{em}}
\end{eqnarray}
\par Further, with the increase in VGT control signal ($u_{vgt}$), the effective area increases and thus also increases the flow. The effective area of the VGT $f_{vgt}(\Tilde{u}_{vgt})$ is modelled as an ellipse
\begin{equation}
\left[\dfrac{f_{vgt}(\Tilde{u}_{vgt}) - c_{f2}}{c_{f1}}\right]^2 + \left[\dfrac{\Tilde{u}_{vgt} - c_{vgt2}}{c_{vgt1}}\right]^2= 1
\end{equation}
which is 
\begin{equation}
	f_{vgt}(\Tilde{u}_{vgt}) = c_{f2} + c_{f1}\sqrt{\text{max}\left(0, 1 - \left(\dfrac{\Tilde{u}_{vgt} - c_{vgt2}}{c_{vgt1}}\right)^2\right)}
	\label{Eq:Append:f vgt}
\end{equation}
\par The power delivered by the turbine, $P_t$ and the mechanical efficiency of the turbocharger $\eta_m$ are modelled as,
\begin{equation}
	P_t\eta_m = \eta_{tm}W_tc_{pe}T_{em}\left(1 - \Pi_t^{1-1/\gamma_e}\right)
	\label{Eq:Append:P_t_eta_m}
\end{equation}
\begin{equation}
	\eta_{tm} = \eta_{tm,max} - c_m(BSR - BSR_{opt})^2
          \label{Eq:Append:eta_tm}
\end{equation}
where, the blade speed ratio (BSR) is defined as the ratio of the turbine blade tip speed to the speed which a gas reaches when expanded entropically at the given pressure ratio $\Pi_t$. \begin{equation}
	BSR = \dfrac{R_t\omega_t}{\sqrt{2c_{pe}T_{em}(1-\Pi_t^{1-1/\gamma_e})}}
\end{equation}
where $R_t$ is the turbine blade radius, and 
\begin{equation}
	c_m = c_{m1}[max(0, \omega_t - c_{m2}]^{c_{m3}}
\end{equation}
\subsection{Compressor}
\label{Appendix:Compressor}
The compressor model consists of two models: the compressor efficiency model and the compressor mass flow model. The compressor efficiency is defined as the ratio of the power from the isentropic process ($P_{c,s}$) to the compressor power ($P_c$)
\begin{equation}
	\eta_c = \dfrac{P_{c,s}}{P_c} = \dfrac{T_{amb}(\Pi_c^{1-1/\gamma_a}-1)}{T_c - T_{amb}}
	\label{Eq:Append:eta c 1}
\end{equation}
where $T_c$ is the temperature after the compressor, and the pressure ratio is given by,
\begin{equation}
	\Pi_c = \dfrac{p_{im}}{p_{amb}}
\end{equation}
The power from the isentropic process is given as,
\begin{equation}
	P_{c,s} = W_c c_{pa}T_{amb}(\Pi_c^{1-1/\gamma_a}-1)
	\label{Eq:Append:P_c,s}
\end{equation}
where $W_c$ is the compressor mass flow and $c_{pa}$ is a constant. Thus, the compressor power can be modelled from Eq. \ref{Eq:Append:eta c 1} and Eq. \ref{Eq:Append:P_c,s} as
\begin{equation}
	P_c = \dfrac{P_{c,s}}{\eta_c} = \dfrac{W_cc_{pa}T_{amb}}{\eta_c} (\Pi_c^{1-1/\gamma_a - 1})
	\label{Eq:Append:P_c}
\end{equation}
$\eta_c$ is modelled as an ellipses, which depends on the pressure ratio ($\Pi_c$) and compressor mass flow ($W_c$),
\begin{equation}
	\eta_c = \eta_{cmax} - \bm{\mathcal{X}}^T\bm{Q}_c\bm{\mathcal{X}}
	\label{Eq:Append:eta c}
\end{equation}
where $\bm{\mathcal{X}}$ is a vector and given as,
\begin{equation}
	\bm{\mathcal{X}} = \begin{bmatrix}
		W_c - W_{copt} \\
		\pi_c - \pi_{copt}
	\end{bmatrix}
\end{equation}
where $W_{copt}$ and $\pi_{copt}$ are the optimum values of $W_c$ and $\pi_c$ respectively. $\pi_c$ is a non linear transformation of $\Pi_c$ as 
\begin{equation}
	\pi_c = (\Pi_c -1)^{c_\pi}
\end{equation}
and $\bm{Q}_c$ is a semi-definite matrix
\begin{equation}
	\bm{Q}_c = \begin{bmatrix}
		a_1 & a_3 \\
		a_3 & a_2
	\end{bmatrix}
\end{equation}
\par The model for compressor mass flow, $W_c$ is modelled using two non-dimensional variables: energy transfer coefficient ($\Psi_c$) and volumetric flow coefficient ($\Phi_c$). The energy transfer coefficient is defined as,
\begin{equation}
	\Psi_c = \dfrac{2c_{pa}T_{amb}(\Pi_c^{1-1/\gamma_a}-1)}{R_c^2\omega_t^2}
	\label{Eq:Append:Psi c}
\end{equation}
where $R_c$ is the compressor blade ratio. The volumetric flow coefficient is defined as,
\begin{equation}
	\Phi_c = \dfrac{W_c/\rho_{amb}}{\pi R_c^3\omega_t} = \dfrac{R_a T_{amb}}{p_{amb}\pi R_c^3\omega_t} W_c
	\label{Eq:Append:Phi_c}
\end{equation}
The energy transfer coefficient ($\Psi_c$) and volumetric flow coefficient ($\Phi_c$) can be described by a part of an ellipse,
\begin{equation}
	c_{\Psi1}(\omega_t)(\Psi_c - c_{\Psi2} )^2 + c_{\Phi1}(\omega_t)(\Phi_c -c_{\Phi2})^2 = 1
	\label{Eq:Append:elliplse c_psi}
\end{equation}
where $c_{\Psi1}$ and $c_{\Phi1}$ are function of turbine speed ($\omega_t$) and modelled as a second order polynomial as,
\begin{equation}
	c_{\Psi1}(\omega_t) = c_{\omega_{\Psi1}}\omega_t^2 + c_{\omega_{\Psi2}}\omega_t + c_{\omega_{\Psi3}}
	\label{Eq:Append:1}
\end{equation}
\begin{equation}
	c_{\Phi1}(\omega_t) = c_{\omega_{\Phi1}}\omega_t^2 + c_{\omega_{\Phi2}}\omega_t + c_{\omega_{\Phi3}}
	\label{Eq:Append:2}
\end{equation}
Solving Eq. \ref{Eq:Append:elliplse c_psi} for $\Phi_c$ and Eq. \ref{Eq:Append:Phi_c} for $W_c$, the compressor mass flow is given as,
\begin{equation}
	W_c = \dfrac{p_{amb} \pi R_c^3\omega_t}{R_aT_{amb}} \Phi_c
	\label{Eq:Append:W_c}
\end{equation}
\begin{equation}
	\Phi_c = \sqrt{\text{max}\left(0, \dfrac{1-c_{\psi1}(\Psi_c - c_{\Psi 2})^2}{c_{\Phi 1}} \right)} + c_{\Phi 2}
	\label{Eq:Append:3}
\end{equation}
where $c_{\omega_{\Psi1}}$, $c_{\omega_{\Psi2}}$, $c_{\omega_{\Psi3}}$, $c_{\omega_{\Phi1}}$, $c_{\omega_{\Phi2}}$, $c_{\omega_{\Phi3}}$, $c_{\Phi 2}$ and $c_{\Psi 2}$ are constant.
\setcounter{equation}{0}
\renewcommand{\theequation}{\thesection.\arabic{equation}}
\section{Brief discussion on lab test data}
\label{Appendix:Lab test data measure}
The engine model considered in this study uses empirical formulae. These equations are engine specific and may not be appropriate for the present study. As discussed in section \ref{Subsection:Surrogate models for empirical formulae}, we consider surrogate neural networks and uses lab test data to train these model. In this section, we briefly discuss lab test data.
\par Practical limitations exist when instrumenting engines for testing. Some physical phenomena are easily measurable, while others are not. When conducting modelling efforts, one must consider the necessary measurements for model tuning to ensure the experimental setup is adequate. The data collection capabilities can also impact loss function weights based on data trustworthiness, as well as noise values applied in the analysis. There are a few signals that pose particular challenges in cost-effective and simple measurement in part due to the high temperature, pressure, dynamics and flow constituents in some areas.
\par Often as areas closer to the cylinder are considered, measurements become increasingly difficult. For example, exhaust port flow, $W_{eo}$ is difficult to measure directly, as the gas is very hot and reactive. In-cylinder measurements are limited by high pressure and temperatures, requiring specialized equipment. Even measuring charge flows directly can be challenging. Because of these limitations, care must be taken in the experimental methods and analysis design to ensure enough data is gathered to be able to observe and identify the system. Sometimes steady state characterizations are used to obtain a static characterization. Consider volumetric efficiency as an example: because measuring flow directly into or out of the cylinder is difficult, a fresh air flow measurement combined with an EGR flow measurement can be used to estimate charge flow to enable the calculation of volumetric efficiency. However, any intake, EGR, or Exhaust leak impacts this measurement, as does the tolerance stack up of both measurements.

\setcounter{equation}{0}
\renewcommand{\theequation}{\thesection.\arabic{equation}}
\section{Calculation of labelled data for training of the neural network for the empirical formulae}
\label{Appendix:Data for pre-trained network}
We approximate the empirical formula using surrogate neural networks and are discussed in section \ref{Subsection:Surrogate models for empirical formulae}. The lab test data required for calculating each of these quantities are shown in Table \ref{Table:Data lab test}. In this section, we discuss the calculation of labelled data from lab-test data. The functional approximation of the empirical formulae is independent of time; thus, static data may be considered for the calculation of labelled data. However, in the case of calculation of labelled for $\eta_{tm}$, the differential equation Eq. \eqref{Eq:Append:omega_t} is considered. Thus, we consider dynamic data for this calculation.
\par We approximate the volumetric efficiency using a surrogate neural network ($\mathcal{N}_1^{(P)}(:,\bm{\theta)}_1^P$). The inputs to the network are intake manifold pressure ($p_{im}$) and engine speed ($n_e$) and trained using labelled data of $\eta_{vol}$. The labelled $\eta_{vol}$ are calculated from measurement of $W_{ei}$ using Eq. \eqref{Eq:Append:W_ei},
\begin{equation}
	\eta_{vol} = \dfrac{120R_aT_{im}W_{ei}}{p_{im}n_eV_d}
\end{equation}
\par The effective area ratio function for EGR valve is approximated using a surrogate neural network ($\mathcal{N}_2^{(P)}(:,\bm{\theta)}_2^P$). The input to the network is $\tilde{u}_{egr}$ and trained using labelled data of $f_{egr}$. The labelled $f_{egr}$ are calculated from the measurement of $W_{egr}$ using Eqs. \eqref{Eq:Append:W egr} and \eqref{Eq:Append:A egr},
\begin{align}
	A_{egr} & = \dfrac{\sqrt{T_{em}R_e}}{p_{im}\Psi_{egr}}W_{egr} \\
	f_{egr} & = \dfrac{A_{egr}}{A_{egrmax}}
\end{align}
\par It is important to node that the value of $\Psi_{egr}$ varies from 0 to 1 (Eq. \eqref{Eq:Append:Psi egr}), thus in the calculation of $A_{egr}$, a situation may occurs where division by 0. This situation occurs when $p_{em}<p_{im}$ (Eq. \eqref{Eq:Append:Pi_egr}). In order to avoid this, labelled data are calculated only for $\Psi_{egr}>10^{-15}$.
\par The neural network approximating ($\mathcal{N}_3^{(P)}(:,\bm{\theta)}_3^P$) for $F_{vgt,\Pi} = f_{vgt}\times f_{\Pi_t}$ is trained using labelled data which are calculated from the measurement of turbine mass flow ($W_t$) using Eq. \eqref{Eq:Append:W_t},
\begin{equation}
	F_{vgt, \Pi_t}(\Tilde{u}_{vgt}, \Pi_t) = f_{vgt}(\tilde{u}_{vgt})\times f_{\Pi_t}(\Pi_t) = \dfrac{W_t\sqrt{T_{em}R_e}}{A_{vgtmax}p_{em}}
    \label{Eq:Appendix:F_VGT_PI_T_Cal}
\end{equation}
\par The training for the neural network ($\mathcal{N}_4^{(P)}(:,\bm{\theta)}_4^P$) for the surrogate model of turbine mechanical efficiency ($\eta_{tm}$) is done using labelled $\eta_{tm}$ which is calculated from the measurement of $\omega_t$ using Eqs. \eqref{Eq:Append:omega_t} and \eqref{Eq:Append:P_t_eta_m}
\begin{equation}
    P_t\eta_m = P_c + J_t\omega_t \dfrac{d\omega_t}{dt}
    \label{Eq:Append:eta_tm_1}
\end{equation}
The compressor power ($P_c$) is calculated as,
\begin{equation}
	P_c = W_cc_{pa}(T_c - T_{amb})
	\label{Eq:Append:P_c 1}
\end{equation}
In the present study, we consider a five-point method to approximate the derivative present in Eq. \eqref{Eq:Append:eta_tm_1}.
\begin{equation}
	f^{(1)}(x) \approx \dfrac{-f(x+2h)+8f(x+h)-8f(x-h)+f(x-2h)}{12h} 
\end{equation}
Once $P_t\eta_m$ calculated from Eq. \eqref{Eq:Append:eta_tm_1}, the labelled  $\eta_{tm}$ are calculated using Eq. \eqref{Eq:Append:P_t_eta_m}
\begin{equation}
	\eta_{tm} = \dfrac{P_t\eta_m}{W_tc_{pe}T_{em}\left(1-\Pi_t^{1-1/\gamma_e}\right)}
\end{equation}
The values of $\eta_{tm}$ are restricted to maximum value $\eta_{tm,max}$ 
\begin{equation}
		\eta_{tm} = \text{min}(\eta_{tm,max}, \eta_{tm}),\;\;\;\;\;\; \eta_{tm,max} = 0.8180
\end{equation}
\par The labelled data for the training of neural network ($\mathcal{N}_5^{(P)}(:,\bm{\theta)}_5^P$) for compressor efficiency ($\eta_{c}$) is calculated using Eqs. \eqref{Eq:Append:P_c} and \eqref{Eq:Append:eta c 1}
\begin{subequations}
\begin{align}
    \eta_c = & \dfrac{P_{c,s}}{P_c} \\
	= & \dfrac{W_cc_{pa}T_{amb}\left(\Pi_c^{1-1/\gamma_a}-1\right)}{W_cc_{pa}\left(T_c - T_{amb}\right)} \\
	= & \dfrac{T_{amb}\left(\Pi_c^{1-1/\gamma_a}-1\right)}{T_c - T_{amb}}
\end{align}
\end{subequations}
To avoid any division by 0, the value of $T_c - T_{amb}$ less than $10^{-6}$ are considered as $10^{-6}$. Further, the value of $\eta_c$ is clipped between 0.2 and $\eta_{cmax}$
\begin{subequations}
\begin{align}
	\eta_c = & \;\text{max}(0.2, \eta_c)\\
	\eta_c = & \;\text{min}(\eta_{cmax}, \eta_c), \;\;\;\;\;\;\; \eta_{cmax}=0.7364
\end{align}
\end{subequations}
\par The training for the neural network ($\mathcal{N}_6^{(P)}(:,\bm{\theta)}_6^P$) for surrogate model of volumetric flow coefficient ($\Phi_c$) is done using labelled data which are calculated from the measurement of compressor mass flow ($W_c$) Eq. \eqref{Eq:Append:Phi_c}
\begin{eqnarray}
	\Phi_c = \dfrac{R_a T_{amb}}{p_{amb}\pi R_c^3\omega_t} W_c
\end{eqnarray}
\setcounter{equation}{0}
\renewcommand{\theequation}{\thesection.\arabic{equation}}
\section{Detail loss function for the PINNs model for the engine}
\par We consider the following loss function for Case 1 to Case 4, which have self-adaptive weights in the loss function, 
\begin{equation}
    \begin{split}
    \mathcal{L}(\bm{\theta}, \bm{\mathcal{\varLambda}}, \bm{\lambda}_{p_{im}}, \bm{\lambda}_{p_{em}}, \bm{\lambda}_{\omega_t}, \bm{\lambda}_{W_{egr}}, \lambda_{T_1}) = & \mathcal{L}_{p_{im}} + \mathcal{L}_{p_{em}} + \mathcal{L}_{\omega_{t}} +  \mathcal{L}_{u_{egr1}} + \\ & \mathcal{L}_{u_{egr2}} + \mathcal{L}_{u_{vgt}} +  10\times \mathcal{L}_{x_{r}} + \lambda_{T_1}\times\mathcal{L}_{T_{1}} + \\
    & \mathcal{L}^{ini}_{p_{im}} + \mathcal{L}^{ini}_{p_{em}} + \mathcal{L}^{ini}_{\omega_{t}} +  \mathcal{L}^{ini}_{\tilde{u}_{egr1}} + \\ 
    & \mathcal{L}^{ini}_{\tilde{u}_{egr2}} + \mathcal{L}^{ini}_{\tilde{u}_{vgt}} + \mathcal{L}^{ini}_{x_{r}} + 100\times \mathcal{L}^{ini}_{T_{1}} + \\
    & \mathcal{L}^{data}_{p_{im}}(\bm{\lambda}_{p_{im}}) + \mathcal{L}^{data}_{p_{em}}(\bm{\lambda}_{p_{em}}) + \\ 
    & \mathcal{L}^{data}_{\omega_t}(\bm{\lambda}_{\omega_t}) + \mathcal{L}^{data}_{W_{egr}}(\bm{\lambda}_{W_{egr}}),
    \end{split}
\label{Eq:Appendix:Loss total loss}
\end{equation}
In the Case 5 where we have not considered self-adaptive weights, the loss function is given as
\begin{equation}
    \begin{split}
    \mathcal{L}(\bm{\theta}, \bm{\mathcal{\varLambda}}) = & \mathcal{L}_{p_{im}} + \mathcal{L}_{p_{em}} + \mathcal{L}_{\omega_{t}} +  \mathcal{L}_{u_{egr1}} + \\ & \mathcal{L}_{u_{egr2}} + \mathcal{L}_{u_{vgt}} +  10\times \mathcal{L}_{x_{r}} + 10^3\times\mathcal{L}_{T_{1}} + \\
    & \mathcal{L}^{ini}_{p_{im}} + \mathcal{L}^{ini}_{p_{em}} + \mathcal{L}^{ini}_{\omega_{t}} +  \mathcal{L}^{ini}_{\tilde{u}_{egr1}} + \\ 
    & \mathcal{L}^{ini}_{\tilde{u}_{egr2}} + \mathcal{L}^{ini}_{\tilde{u}_{vgt}} + \mathcal{L}^{ini}_{x_{r}} + 100\times \mathcal{L}^{ini}_{T_{1}} + \\
    & 10^3\times\mathcal{L}^{data}_{p_{im}} + 10^3\times\mathcal{L}^{data}_{p_{em}} + \\ 
    & 10^3\times\mathcal{L}^{data}_{\omega_t} + 10^3\times\mathcal{L}^{data}_{W_{egr}},
    \end{split}
\label{Eq:Appendix:Loss total loss Case 5}
\end{equation}
where $\bm{\theta} = (\bm{\theta}_1, ..., \bm{\theta}_6)$ are the hyperparameters of all NNs in PINNs, which include both weights and biases, $\bm{\mathcal{\varLambda}}$ are the unknown parameters of the equations which need to be found out. $\bm{\lambda}_{p_{im}}$, $\bm{\lambda}_{p_{em}}$, $\bm{\lambda}_{\omega_t}$ and $\bm{\lambda}_{W_{egr}}$ are the self-adaptive weight \citep{McClenny_2020} for the data loss in $p_{im}$, $p_{em}$, $\omega_t$ and $W_{egr}$ respectively. $\lambda_{T_1}$ is the self-adaptive weight for physics loss in $T_1$. $\mathcal{L}_{p_{im}}(\bm{\theta})$, $\mathcal{L}_{p_{em}}(\bm{\theta})$, $\mathcal{L}_{\omega_{t}}$, $\mathcal{L}_{\tilde{u}_{egr1}}$, $\mathcal{L}_{\tilde{u}_{egr2}}$, and $\mathcal{L}_{\tilde{u}_{vgt}}$ are the physics loss corresponding to the differential equations of the states of the diesel engine $p_{im}$, $p_{em}$, $\omega_t$ $\tilde{u}_{egr1}$, $\tilde{u}_{egr2}$ and $\tilde{u}_{vgt}$ respectively. $\mathcal{L}_{x_{r}}$ and $\mathcal{L}_{T_{1}}$ are the physics loss correspond to $x_r$ and $T_1$ respectively.
\begin{subequations}
	\begin{align}
		\mathcal{L}_{p_{im}} = & \dfrac{1}{n}\sum_{i=1}^{n}r(p_{im})^2 = \dfrac{1}{n}\sum_{i=1}^{n}\left(\dfrac{dp_{im}}{dt} - \dfrac{R_aT_{im}}{V_{im}}\left(W_c+W_{egr}-W_{ei}\right)\right)^2 \\
		\mathcal{L}_{p_{em}} = &  \dfrac{1}{n}\sum_{i=1}^{n}r(p_{em})^2 = \dfrac{1}{n}\sum_{i=1}^{n}\left(\dfrac{dp_{em}}{dt} - \dfrac{R_eT_{em}}{V_{em}}\left(W_{eo} - W_t-W_{egr}\right) \right)^2 \\
		\mathcal{L}_{\omega_{t}} = &  \dfrac{1}{n}\sum_{i=1}^{n}r(\omega_{t})^2 = \dfrac{1}{n}\sum_{i=1}^{n}\left(\dfrac{d\omega_t}{dt} - \dfrac{P_t\eta_{m}-P_c}{J_t\omega_t} \right)^2\\
		\mathcal{L}_{\tilde{u}_{egr1}} = &  \dfrac{1}{n}\sum_{i=1}^{n}r(\tilde{u}_{egr1})^2 = \dfrac{1}{n}\sum_{i=1}^{n}\left(\dfrac{d\tilde{u}_{egr1}}{dt} - \dfrac{1}{\tau_{egr1}}\left[u_{egr}(t-\tau_{degr}) - \tilde{u}_{egr1}\right]\right)^2\\
		\mathcal{L}_{\tilde{u}_{egr2}} = &  \dfrac{1}{n}\sum_{i=1}^{n}r(\tilde{u}_{egr2})^2 = \dfrac{1}{n}\sum_{i=1}^{n}\left(\dfrac{d\tilde{u}_{egr2}}{dt} - \dfrac{1}{\tau_{egr2}}\left[u_{egr}(t-\tau_{degr}) - \tilde{u}_{egr2}\right] \right)^2\\
		\mathcal{L}_{\tilde{u}_{vgt}} = &  \dfrac{1}{n}\sum_{i=1}^{n}r(\tilde{u}_{vgt})^2 = \dfrac{1}{n}\sum_{i=1}^{n}\left(\dfrac{d\tilde{u}_{vgt}}{dt} - \dfrac{1}{\tau_{vgt}}\left[u_{vgt}(t-\tau_{dvgt}) - \tilde{u}_{vgt}\right] \right)^2\\
		\mathcal{L}_{x_{r}} = & \dfrac{1}{n}\sum_{i=1}^{n}r(x_{r})^2 = \dfrac{1}{n}\sum_{i=1}^{n}\left(x_r - \dfrac{\Pi_e^{1/\gamma_a}x_p^{-1/\gamma_a}}{r_cx_v}\right)^2\\
		\mathcal{L}_{T_{1}} = & \dfrac{1}{n}\sum_{i=1}^{n}r(T_{1})^2 = \dfrac{1}{n}\sum_{i=1}^{n}\left(T_1 - \left(x_rT_e + (1- x_r)T_{im}\right)\right)^2
	\end{align}
\end{subequations}
where $n$ and $r(.)$ are the number of residual points and residual, respectively. 
\par In the Case 1 to Case 4, $\mathcal{L}^{data}_{p_{im}}(\bm{\lambda}_{p_{im}}$,  $\mathcal{L}^{data}_{p_{em}}(\bm{\lambda}_{p_{em}})$, $\mathcal{L}^{data}_{\omega_t}(\bm{\lambda}_{\omega_t})$, and  $\mathcal{L}^{data}_{W_{egr}}(\bm{\lambda}_{W_{egr}})$ are the data loss in $p_{im}$, $p_{em}$, $\omega_t$ and $W_{egr}$ respectively and defined as,
\begin{subequations}
	\begin{align}
		\mathcal{L}^{data}_{p_{im}}(\bm{\lambda}_{p_{im}}) = & \dfrac{1}{n}\sum_{j=1}^n\left[\left(p_{im_{data}}^{(j)} - \hat{p}_{im_{\mathcal{N}_1}}^{(j)}\right)\lambda_{p_{im}}^{(j)}\right]^2\\ \mathcal{L}^{data}_{p_{em}}(\bm{\lambda}_{p_{em}}) = & \dfrac{1}{n}\sum_{j=1}^n\left[\left(p_{em_{data}}^{(j)} - \hat{p}_{em_{\mathcal{N}_1}}^{(j)}\right)\lambda_{p_{em}}^{(j)}\right]^2\\ \mathcal{L}^{data}_{\omega_t}(\bm{\lambda}_{\omega_t}) = & \dfrac{1}{n}\sum_{j=1}^n\left[\left(\omega_{t_{data}}^{(j)} - \hat{\omega}_{t_{\mathcal{N}_5}}^{(j)}\right)\lambda_{\omega_{t}}^{(j)}\right]^2\\ \mathcal{L}^{data}_{W_{egr}}(\bm{\lambda}_{W_{egr}}) = & \dfrac{1}{n}\sum_{j=1}^n\left[\left(W_{egr_{data}}^{(j)} - \widehat{W}_{egr_{NN}}^{(j)}\right)\lambda_{W_{egr}}^{(j)}\right]^2
	\end{align}
\end{subequations}
The same in the Case 5 is given as,
\begin{subequations}
	\begin{align}
		\mathcal{L}^{data}_{p_{im}} = & \dfrac{1}{n}\sum_{j=1}^n\left[p_{im_{data}}^{(j)} - \hat{p}_{im_{\mathcal{N}_1}}^{(j)}\right]^2\\ \mathcal{L}^{data}_{p_{em}} = & \dfrac{1}{n}\sum_{j=1}^n\left[p_{em_{data}}^{(j)} - \hat{p}_{em_{\mathcal{N}_1}}^{(j)}\right]^2\\ \mathcal{L}^{data}_{\omega_t} = & \dfrac{1}{n}\sum_{j=1}^n\left[\omega_{t_{data}}^{(j)} - \hat{\omega}_{t_{\mathcal{N}_5}}^{(j)}\right]^2\\ \mathcal{L}^{data}_{W_{egr}} = & \dfrac{1}{n}\sum_{j=1}^n\left[W_{egr_{data}}^{(j)} - \widehat{W}_{egr_{NN}}^{(j)}\right]^2
	\end{align}
\end{subequations}
where $p_{im_{data}}$, $p_{em_{data}}$, $\omega_{t_{data}}$ and $W_{egr_{data}}$ are the measured data of $p_{im}$, $p_{em}$, $\omega_{t}$ and $W_{egr}$ respectively. $\hat{p}_{im_{\mathcal{N}_1}}$, $\hat{p}_{em_{\mathcal{N}_1}}$ and $\hat{\omega}_{t_{\mathcal{N}_5}}$ are the predicted values in $p_{im}$, $p_{em}$ and $\omega_{t}$ respectively from $\mathcal{N}_1(t;\bm{\theta}_1)$, $\mathcal{N}_1(t;\bm{\theta}_1)$ and $\mathcal{N}_5(t;\bm{\theta}_5)$ respectively. Similarly, $\widehat{W}_{egr_{NN}}$ is predicted value in $W_{egr}$ from NNs output. $n$ is the number of measured data points. 
\par $\mathcal{L}^{ini}_{p_{im}}$, $\mathcal{L}^{ini}_{p_{em}}$, $\mathcal{L}^{ini}_{\omega_{t}}$, $\mathcal{L}^{ini}_{\tilde{u}_{egr1}}$, $\mathcal{L}^{ini}_{\tilde{u}_{egr2}}$, $\mathcal{L}^{ini}_{\tilde{u}_{vgt}}$, $\mathcal{L}^{ini}_{x_{r}}$ and $\mathcal{L}^{ini}_{T_{1}}$ are the losses in initial conditions in $p_{im}$, $p_{em}$, $\omega_t$ $\tilde{u}_{egr1}$, $\tilde{u}_{egr2}$, $\tilde{u}_{vgt}$, $x_r$ and $T_1$ respectively.
\begin{subequations}
	\begin{align}
		\mathcal{L}^{ini}_{p_{im}} & = \dfrac{1}{1}\sum_{j=1}^1\left(p_{im_{0}}^{(j)} - \hat{p}_{im_{0}}^{(j)}\right)^2\\
		\mathcal{L}^{ini}_{p_{em}} & = \dfrac{1}{1}\sum_{j=1}^1\left(p_{em_{0}}^{(j)} - \hat{p}_{em_{0}}^{(j)}\right)^2\\
		\mathcal{L}^{ini}_{\omega_{t}} & = \dfrac{1}{1}\sum_{j=1}^1\left(\omega_{t_{0}}^{(j)} - \hat{\omega}_{t_{0}}^{(j)}\right)^2\\
		\mathcal{L}^{ini}_{\tilde{u}_{egr1}} & = \dfrac{1}{1}\sum_{j=1}^1\left(\tilde{u}_{egr1_{0}}^{(j)} - \hat{\tilde{u}}_{egr1_{0}}^{(j)}\right)^2 \\ 
		\mathcal{L}^{ini}_{\tilde{u}_{egr2}} & = \dfrac{1}{1}\sum_{j=1}^1\left(\tilde{u}_{egr2_{0}}^{(j)} - \hat{\tilde{u}}_{egr2_{0}}^{(j)}\right)^2\\ \mathcal{L}^{ini}_{\tilde{u}_{vgt}} & = \dfrac{1}{1}\sum_{j=1}^1\left(\tilde{u}_{vgt_{0}}^{(j)} - \hat{\tilde{u}}_{vgt_{0}}^{(j)}\right)^2\\
		\mathcal{L}^{ini}_{x_{r}} & = \dfrac{1}{1}\sum_{j=1}^1\left(x_{r_{0}}^{(j)} - \hat{x}_{r_{0}}^{(j)}\right)^2\\
		\mathcal{L}^{ini}_{T_{1}} & = \dfrac{1}{1}\sum_{j=1}^1\left(T_{1_{0}}^{(j)} - \hat{T}_{1_{0}}^{(j)}\right)^2
	\end{align}
\end{subequations}
where $p_{im_{0}}$, $p_{em_{0}}$, $\omega_{t_{0}}$, $\tilde{u}_{egr1_{0}}$, $\tilde{u}_{egr2_{0}}$, $\tilde{u}_{vgt_{0}}$, $x_{r_{0}}$ and  $T_{1_{0}}$ are the initial conditions and $\hat{p}_{im_{0}}$, $\hat{p}_{em_{0}}$, $\hat{\omega}_{t_{0}}$, $\hat{\tilde{u}}_{egr1_{0}}$, $\hat{\tilde{u}}_{egr2_{0}}$, $\hat{\tilde{u}}_{vgt_{0}}$, $\hat{x}_{r_{0}}$ and $\hat{T}_{1_{0}}$ are corresponding output from neural network at time $t=0$ for $p_{im}$, $p_{em}$, $\omega_{t}$, $\tilde{u}_{egr1}$, $\tilde{u}_{egr2}$, $\tilde{u}_{vgt}$, $x_{r}$ and $T_{1}$ respectively.
\setcounter{equation}{0}
\renewcommand{\theequation}{\thesection.\arabic{equation}}
\section{Additional Tables}
\label{Appendix:Additional Table}
\begin{table}[H]
\centering
\caption{Values of the constants considered in the present study.}
\label{Table:Constant}
\begin{tabular}{C{0.5cm}L{9cm}C{1.4cm}C{3.5cm}} \hline
    & \multicolumn{1}{c}{Description} & Symbol & Value \\ \hline
1 & Ideal gas constant of air & $R_a$ &  287 \\ 
2 & Intake manifold temperature & $T_{im}$ & 300.6186 \\ 
3 & Intake manifold volume & $V_{im}$ & 0.0220\\ 
4 & Ideal gas constant of exhaust gas & $R_e$ & 286\\ 
5 & Exhaust manifold volume & $V_{em}$ & 0.0200\\ \hline
6 & Displaced volume of the cylinder & $V_d$ & 0.0127\\ 
7 & Number of cylinder & $n_{cyl}$ & 6\\ 
8 & Specific heat capacity ratio of air & $\gamma_a$ & 1.3964\\ 
9 & Specific heat capacity at constant pressure of air & $c_{pa}$  & 1011\\ 
10 & Specific heat capacity at constant volume air & $c_{va}$  & 724\\ \hline
11 & Compression ratio & $r_c$  & 17\\ 
12 & Fuel consumed during constant-volume combustion & $x_{cv}$ & $2.3371\times10^{-14}$ \\ 
13 & Heating value of fuel & $q_{HV}$ & 42900000 \\ 
14 & Diameter of exhaust pipe & $d_{pipe}$ & 0.1 \\ 
15 & Length of exhaust pipe & $l_{pipe}$ & 1\\ \hline
16 & Number of exhaust pipe & $n_{pipe}$ & 2\\ 
17 & Specific heat capacity at constant pressure of exhaust gas & $c_{pe}$ &  1332\\ 
18 & Time constant 1 for EGR & $\tau_{egr1}$ & 0.05\\ 
19 & Time constant 2 for EGR & $\tau_{egr2}$ & 0.13 \\ 
20 & Time delay constant for EGR & $\tau_{degr}$ & 0.065\\ \hline
21 & Constant for EGR overshoot & $K_{egr}$ & 1.8\\ 
22 & Optimal value of pressure ratio of EGR & $\Pi_{egropt}$ & 0.6500 \\ 
23 & Inertial of turbocharger & $J_t$ & $2.0\times10^{-4}$ \\ 
24 & Time constant for VGT & $\tau_{vgt}$ & 0.025 \\ 
25 & Time delay constant for VGT & $\tau_{dvgt}$ & 0.04 \\ \hline
26 & Specific heat capacity at constant pressure of exhaust & $c_{pe}$ & 1332 \\ 
27 & Specific heat capacity ratio of exhaust gas & $\gamma_e$ & 1.2734\\ 
28 & turbine blade radius & $R_t$ & 0.04 \\ 
29 & compressor blade radius & $R_c$ & 0.0400\\ \hline
\end{tabular}
\end{table}
\begin{table}[H]
\centering
\caption{True value of the unknown parameters.}
\label{Table:Append:Unknowns}
\begin{tabular}{c|C{1.5cm}C{1.5cm}C{2cm}C{2.5cm}} \hline
Unknown & $\eta_{sc}$ & $h_{tot}$ & $A_{egrmax}$ & $A_{vgtmax}$ \\ \hline
Value & 1.1015 & 96.2755 & $4.0\times10^{-4}$ & $8.4558\times10^{-4}$\\ \hline
\end{tabular}
\end{table}
\begin{table}[H]
\centering
\caption{Value for the coefficients of the empirical formulae.}
\label{Table:Coefficient}
\begin{tabular}{C{1.5cm}C{3cm}|C{1.5cm}C{3cm}|C{1.5cm}C{1.6cm}} \hline
Symbol & Value & Symbol & Value & Symbol & Value \\ \hline
$c_{vol1}$ & $-2.0817\times 10^{-4}$ & $c_{egr1}$ & $-1.1104\times 10^{-4}$ &  & \\
$c_{vol2}$ & -0.0034 & $c_{egr2}$ & $0.0178$ & & \\
$c_{vol3}$ & 1.1497 & $c_{egr3}$ & $0$ & & \\ \hline \hline
$c_{\omega_{\Psi1}}$ & $1.0882\times 10^{-8}$ & $c_{\omega_{\Phi1}}$ & $-1.4298\times 10^{-8}$ & $c_{{\Psi2}}$ & $0$ \\
$c_{\omega_{\Psi2}}$ & $-1.7320\times 10^{-4}$ & $c_{\omega_{\Phi2}}$ & $-0.0015$ & $c_{{\Phi2}}$ & $0$ \\
$c_{\omega_{\Psi3}}$ & $1.0286$ & $c_{\omega_{\Phi3}}$ & $29.6462$ & & \\ \hline \hline
$\pi_{copt}$ & $1.0455$ & $c_{m1}$ & $1.3563$ & $c_{vgt1}$ & $126.8719$ \\
$W_{copt}$ & $0.2753$ & $c_{m2}$ & $2.7692e+03$ & $c_{vgt2}$ & $117.1447$ \\ 
$a_1$ & $3.0919$ &  $c_{m3}$ & $0.0100$ & $c_{f1}$ &  $1.9480$\\
$a_2$ & $2.1479$ & $BSR_{opt}$ & $0.9755$ & $c_{f2}$ & $-0.7763$\\ 
$a_3$ & $-2.4823$ & $\eta_{tm,max}$ & $0.8180$ & $K_t$ & $2.8902$\\
$\eta_{cmax}$ & $0.7364$ & & & & \\
$c_\pi$ & $0.2708$ & & & & \\  \cline{1-6}
\end{tabular}
\end{table}
\setcounter{equation}{0}
\setcounter{figure}{0}
\renewcommand{\theequation}{\thesection.\arabic{equation}}
\section{Additional figures}
\label{Appendix:Additiona Figure}
In this section, we present the results for Case 1 and Case 2.
\subsection*{For Case-1: 3 unknown parameters with clean data}
\begin{figure}[H]
    \centering
    \includegraphics{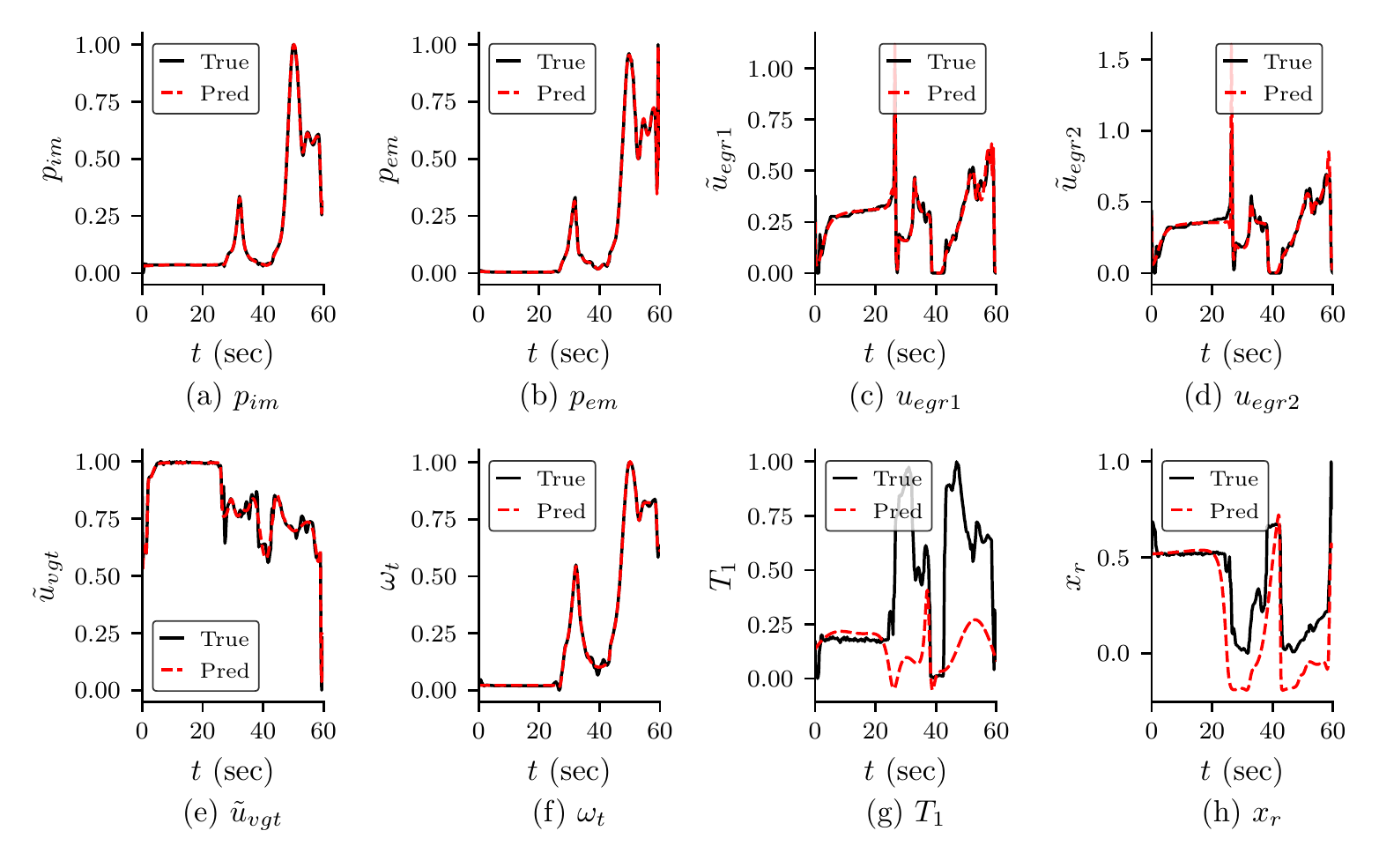}
    \caption{\textbf{Predicted states and $T_1$ and $x_r$ for Case 1:} Predicted dynamics of the state variables of the engine and $T_1$ and $x_r$ for Case 1 (PINN with self-adaptive weights for 3 unknown parameters). It can be observed that the predicted dynamics of the states are in good agreement with the true values. However, similar to 4 unknown parameters $T_1$ and $x_r$ do not match with the true value.}
    \label{Fig:Append:Case 1 States}
\end{figure}

\begin{figure}[H]
    \centering
    \includegraphics{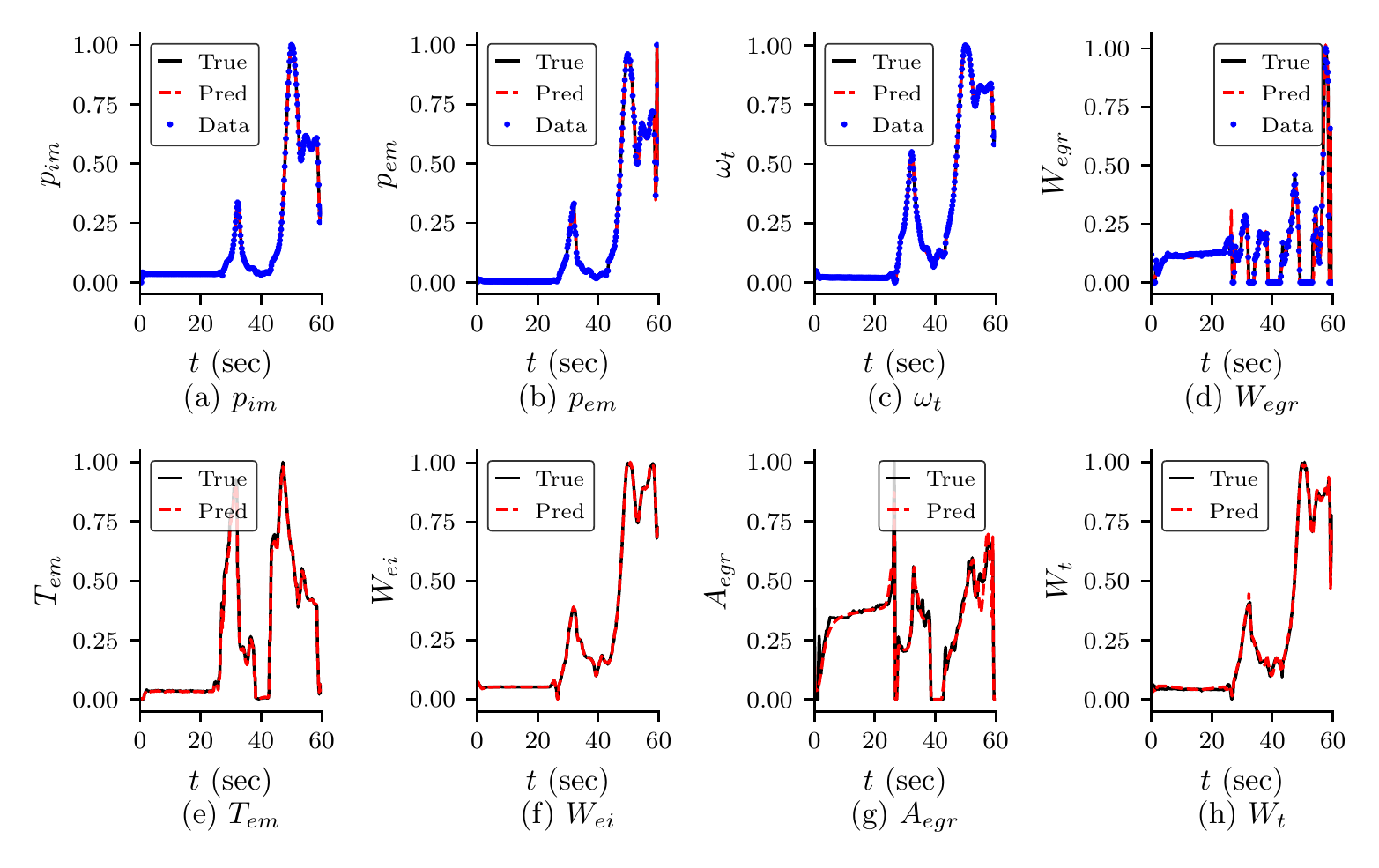}
    \caption{\textbf{Predicted dynamics of variables for Case 1:} (a)-(d) Predicted dynamics of the variables whose field measurement data are known. (e)-(h) dynamics of important variables which also depend on the unknown parameters}
    \label{Fig:Append:Case 1 additional variable}
\end{figure}

\begin{figure}[H]
    \centering
    \includegraphics{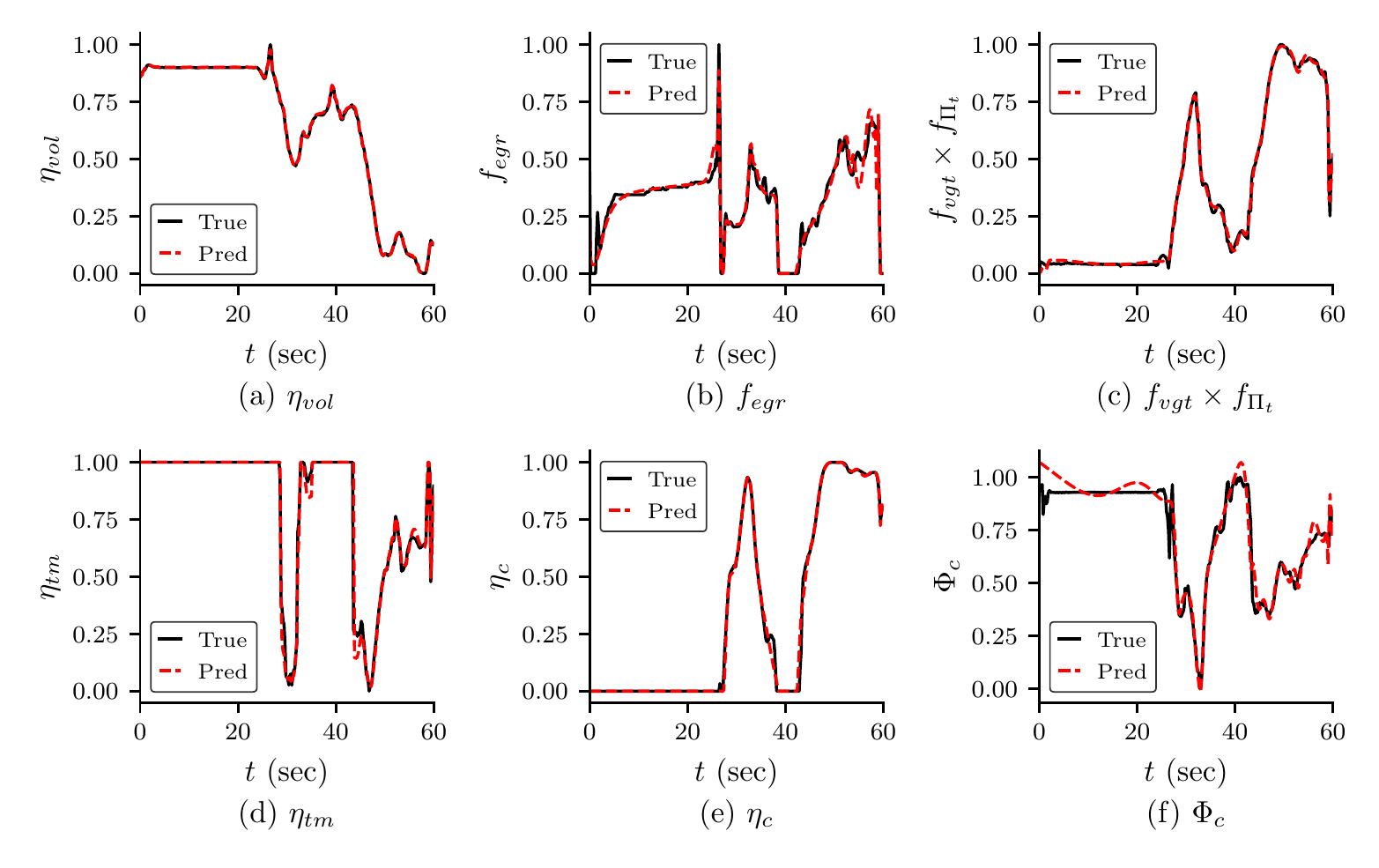}
    \caption{\textbf{Empirical formulae for Case 1:} The predicted values of empirical formulae for Case 1 (3 unknown parameters with clean data).}
    \label{Fig:Append:Case 1 empirical}
\end{figure}

\begin{figure}[H]
    \centering
    \includegraphics[width=1\textwidth]{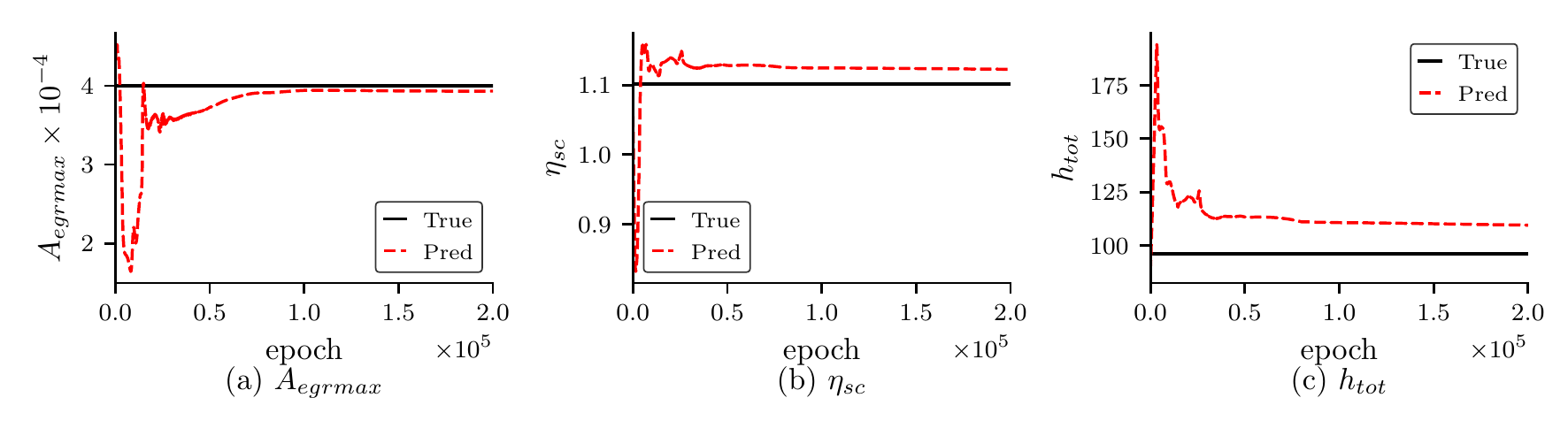}
    \caption{\textbf{Convergence of the unknown parameters for Case 1:} Convergence of the unknown parameters with epoch for Case 1 (3 unknown parameters with clean data)}
    \label{Fig:Append:Case 1 Convergence}
\end{figure}

The predicted state variables and $T_1$ and $x_r$ for Case 1 (3 unknown with clear data) are shown in Fig. \ref{Fig:Append:Case 1 States}. The predicted dynamics of the known variables are shown in Fig. \ref{Fig:Append:Case 1 additional variable}(a)-(d). In Fig. \ref{Fig:Append:Case 1 additional variable}(e)-(h), we have shown the dynamics of variables which are dependent on the unknown parameters. The predicted empirical formulae are shown in Fig. \ref{Fig:Append:Case 1 empirical}. We also studied the convergence of the unknown parameters, which are shown in Fig. \ref{Fig:Append:Case 1 Convergence}.
\subsection*{For Case-2: 3 unknown parameters with noisy data}
\begin{figure}[H]
    \centering
    \includegraphics{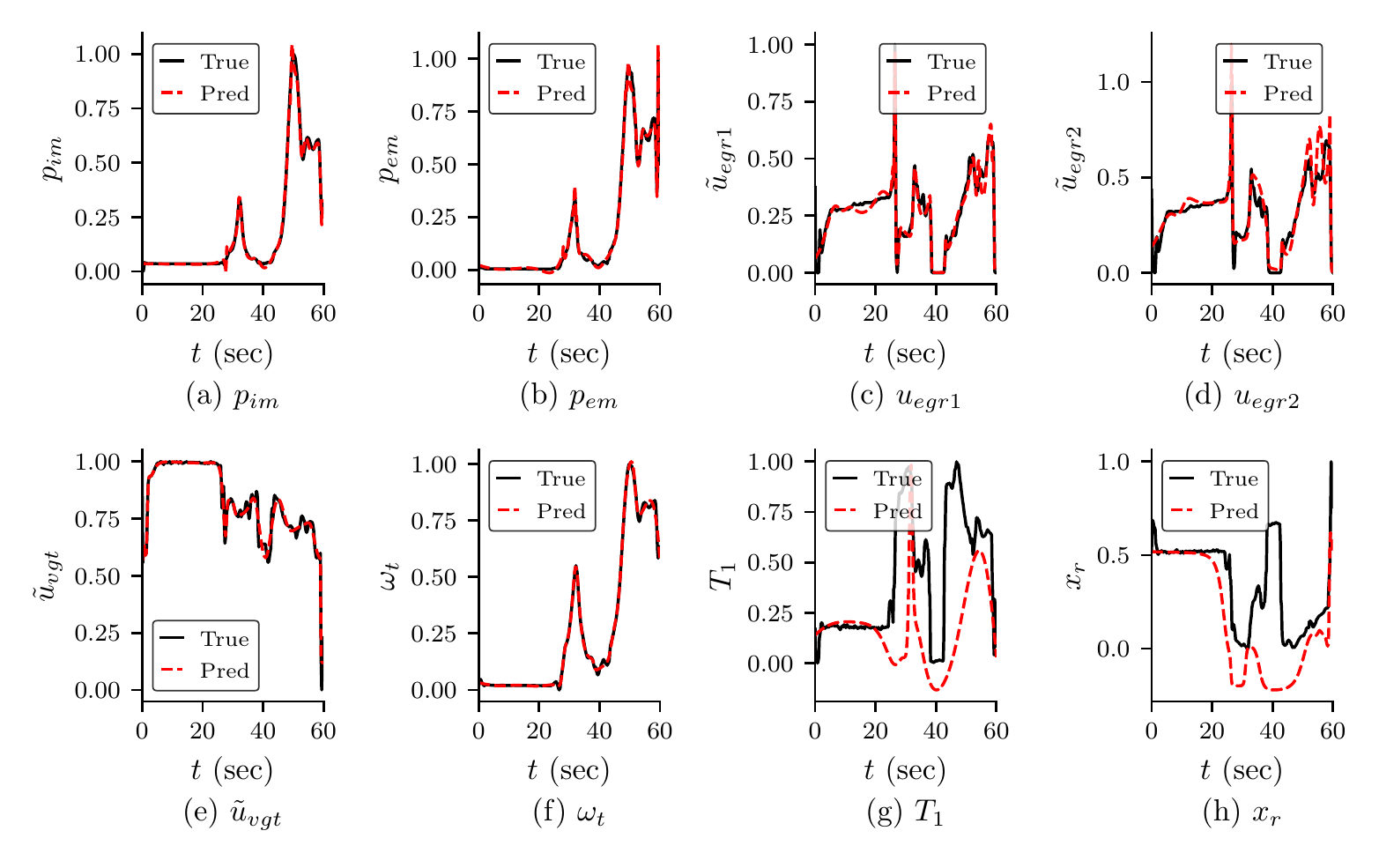}
    \caption{\textbf{Predicted states and $T_1$ and $x_r$ for Case 2:} Predicted dynamics of the state variables of the engine and $T_1$ and $x_r$ for Case 2 (PINN with self-adaptive weights for 3 unknown paramters). It can be observed that the predicted dynamics of the states are in good agreement with the true values. However, similar to 4 unknown parameters $T_1$ and $x_r$ do not match with the true value.}
    \label{Fig:Append:Case 2 States}
\end{figure}

\begin{figure}[H]
    \centering
    \includegraphics{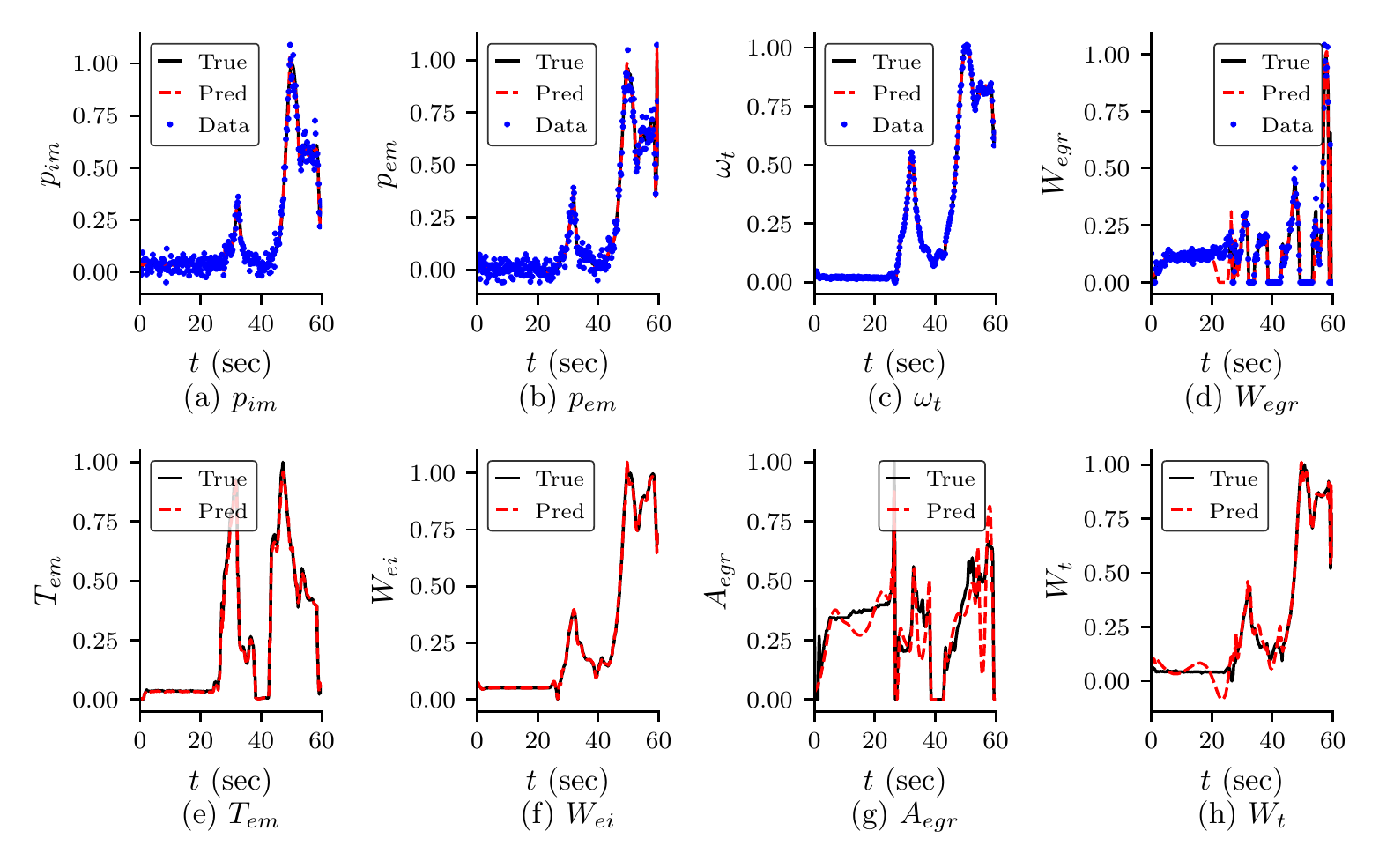}
    \caption{\textbf{Predicted dynamics of variables for Case 2:} (a)-(d) Predicted dynamics of the variables whose field measurement data are known. (e)-(h) dynamics of important variables which also depend on the unknown parameters}
    \label{Fig:Append:Case 2 additional variable}
\end{figure}

\begin{figure}[H]
    \centering
    \includegraphics{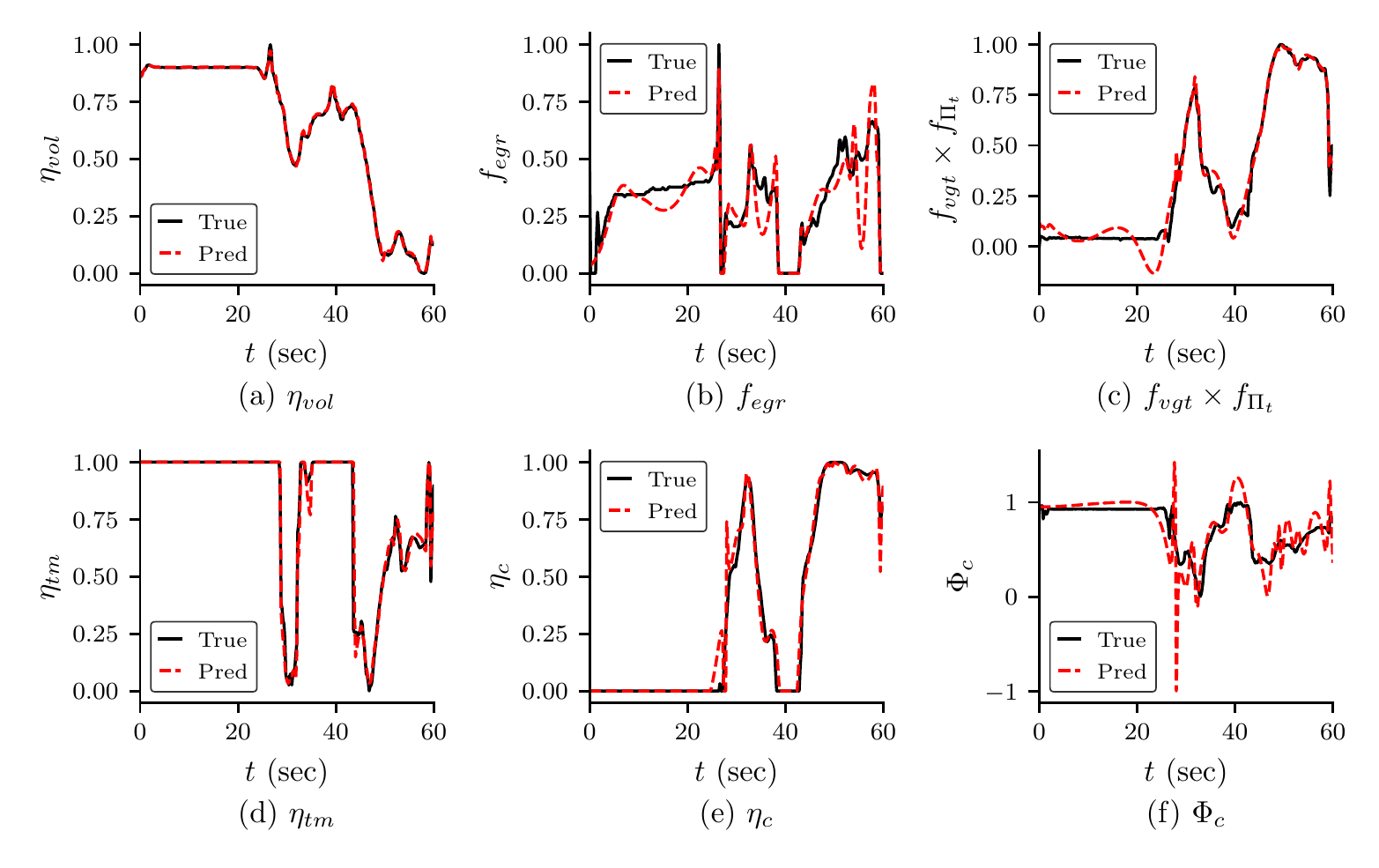}
    \caption{\textbf{Empirical formulae for Case 2:} The predicted values of empirical formulae for Case 2 (3 unknown parameters with noisy data).}
    \label{Fig:Append:Case 2 empirical}
\end{figure}

\begin{figure}[H]
    \centering
    \includegraphics[width=1\textwidth]{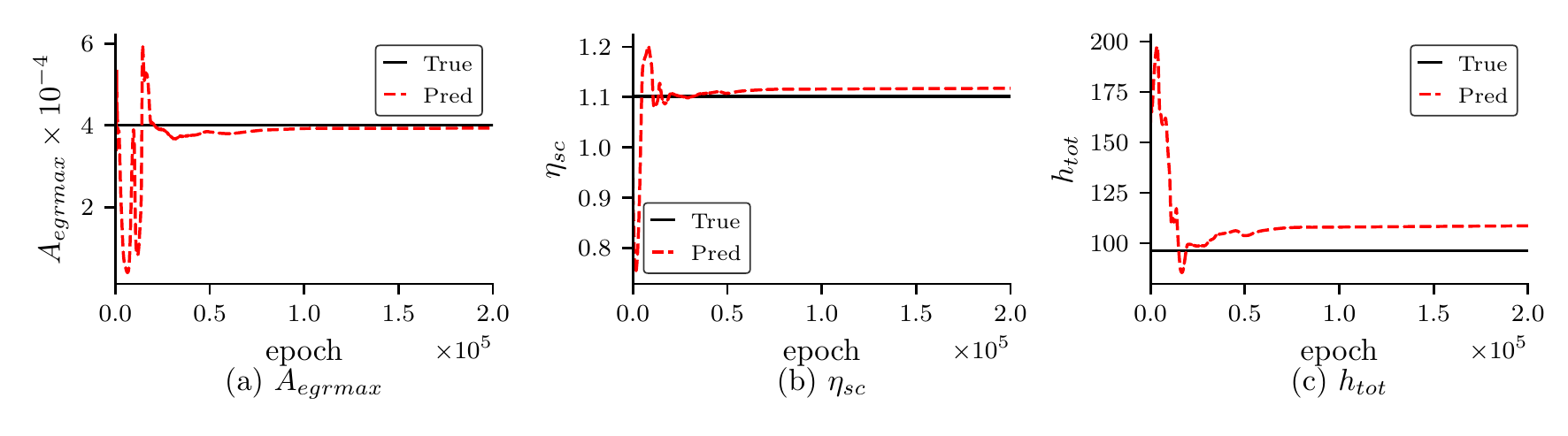}
    \caption{\textbf{Convergence of the unknown parameters for Case 2:} Convergence of the unknown parameters with epoch for Case 2 (3 unknown parameters with noisy data)}
    \label{Fig:Append:Case 2 Convergence}
\end{figure}

The predicted state variables and $T_1$ and $x_r$ for Case 2 (3 unknown with noisy data) are shown in Fig. \ref{Fig:Append:Case 2 States}. The predicted dynamics of the known variables are shown in Fig. \ref{Fig:Append:Case 2 additional variable}(a)-(d). In Fig. \ref{Fig:Append:Case 2 additional variable}(e)-(h), we have shown the dynamics of variables which are dependent on the unknown parameters. The predicted empirical formulae are shown in Fig. \ref{Fig:Append:Case 2 empirical}. We also studied the convergence of the unknown parameters, which are shown in Fig. \ref{Fig:Append:Case 2 Convergence}.
\setcounter{equation}{0}
\setcounter{figure}{0}
\setcounter{table}{0}
\renewcommand{\theequation}{\thesection.\arabic{equation}}
\section{Neural network surrogates for empirical formulae}
\label{Subsection:Traning of pretrained network}
The empirical formulae of the engine model are approximated using surrogate neural networks and are discussed in section  \ref{Subsection:Surrogate models for empirical formulae}. In section \ref{Section:Data acquisition}, we discuss the laboratory data required to train these neural networks. The laboratory data required for training of each neural network are shown in Table \ref{Table:Data lab test} (\S \ref{Section:Data acquisition}). The labelled data for training these neural networks may be calculated from static data on the entire operational range of each quantity except for turbine mechanical efficiency ($\eta_{tm}$). The labelled data for the turbine mechanical efficiency is calculated using Eq. \eqref{Eq:Append:omega_t} (\S \ref{Appendix:Turbocharger}), which is a differential equation, thus requiring dynamic data with fine $dt$. The calculations of the labelled data from the laboratory measurements are discussed in Appendix \ref{Appendix:Data for pre-trained network}. In the case of training of neural network $\mathcal{N}_3^{(P)}(\bm{x};\bm{\theta}^P_3)$ for the approximation of $F_{vgt, \Pi_t}$, $L_2$ weight regularizer is considered in the loss function with a coefficient $5\times10^{-10}$.
\par
\begin{figure}[H]
    \includegraphics{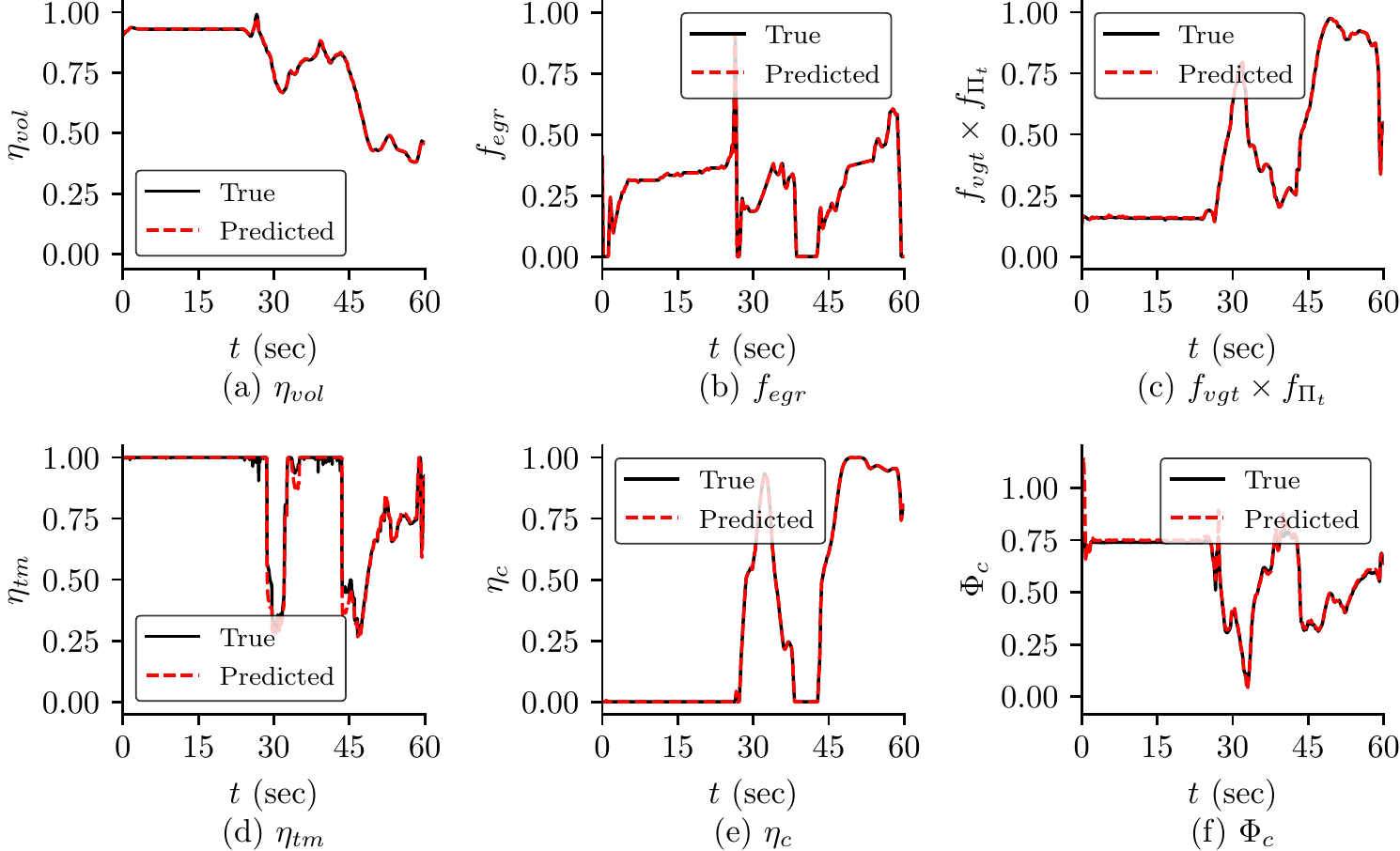}
    \caption{\textbf{Prediction of empirical formulae:} The predicted and true values of the empirical formulae for test case (Case-V). The plots are normalized within 20-minute data, and only a portion (0 to 1 minute) of the results are shown. The predicted empirical formulae are in good agreement with the true values.}
	\label{Fig:Prediction of pre-trained}
\end{figure}

 The predicted values of the empirical formulae for Case-V (Table \ref{Table:Pretrained data} in \S \ref{Section:Data acquisition}) with the true values for 1-minute duration are shown in Fig. \ref{Fig:Prediction of pre-trained}. The \% relative $L_2$ errors for training and testing data set are shown in the last two columns of Table \ref{Table:Pre train Network size}. We observe that the neural networks are able to predict the empirical quantity with very good accuracy. The testing error in the case of the surrogate neural network for $f_{egr}$ is smaller than the training error. This is because of the nature of the function and the data considered. The input-output relationship is simple, with only one input and one output. The maximum value of the testing data is smaller than the maximum value of the training data. Similarly, the minimum value of testing data is larger than the minimum value of training data. We also observed that the standard deviation of testing data is smaller than the standard deviation of the training data. Since the EGR system is independent and  $f_{egr}$ depends only on  $\tilde{u}_{egr}$, not any other variables (e.g. ambient temperature and pressure), we assume that most of the testing set of data might be within the training data set (training data set is 2 hrs while testing data set is 20 minutes). Thus, the testing error is marginally smaller than the training error. The testing error in the case of the surrogate model for $\eta_{tm}$ is smaller than that of the training error. The approximation considered in calculating the labelled data for $\eta_{tm}$ from the laboratory data, we have considered a five-point method to approximate the differentiation present in Eq. \eqref{Eq:Append:omega_t} (\S \ref{Appendix:Turbocharger}). Thus, a few noisy data are observed in both training and testing data sets. As the duration of the training data set is larger than the testing dataset, the amount of noisy data is more in the training data. Thus, the error in training is slightly higher than the testing error. These neural networks, after training, will be used in the places of the empirical formulae in the inverse problem. The trained weights and biases will be considered fixed in the inverse problem.
\begin{table}[H]
\centering
\caption{\textbf{Details of neural networks for empirical formulae:} Details of the DNNs to approximate empirical formulae. The first two columns specify the neural network (Table \ref{Table:Surrogate empirical formulae}) and their input, respectively. The third column indicates the neural network size considered. The activation function of the hidden layers is $\sigma(.) = \text{tanh(.)}$. The "Output" column specifies the empirical quantity the neural network approximated. The last two columns give the results for the test data after the completion of the training. The column "Error" specifies the relative $\%L_2$ error for the test case (Case V). Appropriate scaling of input and output are considered in the training of neural networks.}
\label{Table:Pre train Network size}
\begin{tabular}{C{2.15cm}C{2.1cm}C{2.7cm}C{1.2cm}C{2.85cm}C{0.9cm}C{0.75cm}} \hline
Neural & \multirow{1}{*}{Input} & \multirow{2}{*}{Network size} & \multirow{2}{*}{Output} & Output & \multicolumn{2}{c}{$L_2$ error (\%)} \\ \cline{6-7}
network & ($\bm{x}$) & & & restrict $^\ddag$ & Train & Test \\  \hline \\[-0.75cm]
& & & & & \\
$\mathcal{N}_1^{(P)}(\bm{x};\bm{\theta}^P_1)$ & $n_e, p_{im}$  & $[2,\;4,\;4,\;1]$ & $\eta_{vol}$ &  & 0.01 &  0.03 \\[0.35cm]
$\mathcal{N}_2^{(P)}(\bm{x};\bm{\theta}^P_2)$ & $\tilde{u}_{egr}$ & $[1,\;4,\;4,\;1]$ & $f_{egr}$ & $S(f_{egr})$ & 0.14 &  0.10  \\[0.35cm] 
$\mathcal{N}_3^{(P)}(\bm{x};\bm{\theta}^P_3)$ &  $\Pi_t, \tilde{u}_{vgt}$ & $[2,\; 8,\; 8,\; 8,\; 1]$ & $F_{vgt, \Pi_t}$ & $1.1\times S(F_{vgt, \Pi_t})$ & 0.03 & 0.52  \\[0.35cm]
$\mathcal{N}_4^{(P)}(\bm{x};\bm{\theta}^P_4)$ &  $\omega_t, \Pi_t, T_{em}$ & $[3,\;4,\;4,\;4,\;1]$ & $\eta_{tm}$ & $\text{min}(0.818, \eta_{tm})$ & 1.62 &  1.32  \\[0.35cm]
$\mathcal{N}_5^{(P)}(\bm{x};\bm{\theta}^P_5)$ &  $W_c,\Pi_c$ & $[2,\;4,\;4,\;4,\;1]$ & $\eta_c$ & $\text{max}(0.2,S(\eta_c))$ & 0.16 & 0.18  \\[0.35cm] 
$\mathcal{N}_6^{(P)}(\bm{x};\bm{\theta}^P_6)$ &  $\omega_t,\Pi_c, T_{amb}$ & $[3,10,10,10,1]$ & $\Phi_c$ & $S(\Phi_c)$ & 0.76 & 1.13 \\[0.35cm] \hline
\multicolumn{7}{l}{$^\ddag$ $S\longrightarrow$ sigmoid function}\\ \hline
\end{tabular}
\end{table}
\end{appendices}